\documentclass[sigconf]{acmart} 
\acmConference[KDD '25]{Make sure to enter the correct conference title from your rights confirmation email}{August 3--7, 2025}{Toronto, Canada}

\usepackage[normalem]{ulem}
\usepackage{array}

\usepackage[linesnumbered,lined,ruled,commentsnumbered]{algorithm2e}
\usepackage{blindtext}
\usepackage{booktabs}
\usepackage{caption}
\usepackage{multirow}
\usepackage{graphicx}
\usepackage{float}
\usepackage{subfig}
\usepackage{enumitem}
\usepackage{algpseudocode}
\usepackage{pifont}
\usepackage{diagbox}
\usepackage{amsmath}
\usepackage{amsthm}
\newtheorem{theorem}{Theorem}[section]
\newtheorem{lemma}[theorem]{Lemma}

\newcommand{\question}[1]{{\textit{#1}}}

\newcommand{\projecttitle}{\textsc{FLMarket}\xspace}

\usepackage{tikz}

\newcommand{\myparagraph}[1]{\smallskip \noindent{\bf {#1}.}}

\newcommand{\out}[1] {}

\newcounter{codeLineCntr}

\reversemarginpar
\newif\ifnotes
\notestrue

\newcommand{\punt}[1]{}

\renewcommand{\eqref}[1]{Equation~(\ref{#1})}

\newcommand{\proc}[1]{\ifmmode\mbox{\textsc{#1}}\else\textsc{#1}\fi}

  \newcommand{\func}[1]{\ifmmode\mathrm{#1}\else\textrm{#1}fi} %

\newcounter{remark}[section]

\acmConference[KDD '25]{In Proceedings of the 31th ACM SIGKDD Conference on Knowledge Discovery and Data Mining}{August 03--07,2025}{Toronto,Canada}
\acmISBN{ }

\begin{document}

\title[\projecttitle]{\projecttitle: Enabling Privacy-preserved Pre-training Data Pricing for Federated Learning}


\author{Zhenyu Wen}
\email{zhenyuwen@zjut.edu.cn}
\affiliation{%
    \institution{Zhejiang University of Technology}
    \city{Hangzhou}
    \country{China}
}

\author{Wanglei Feng}
\email{wlfeng97@gmail.com}
\affiliation{%
    \institution{Zhejiang University of Technology}
    \city{Hangzhou}
    \country{China}
}

\author{Di Wu}
\authornote{Corresponding author.}
\email{dw217@st-andrews.ac.uk}
\affiliation{%
    \institution{University of St Andrews}
    \city{St Andrews}
    \country{UK}
}

\author{Haozhen Hu}
\email{huhaozhen5125@163.com}
\affiliation{%
    \institution{Zhejiang University of Technology}
    \city{Hangzhou}
    \country{China}
}

\author{Chang Xu}
\email{xuchang19980309@gmail.com}
\affiliation{%
    \institution{Zhejiang University of Technology}
    \city{Hangzhou}
    \country{China}
}

\author{Bin Qian}
\email{bin.qian@zju.edu.cn}
\affiliation{%
    \institution{Zhejiang University}
    \city{Hangzhou}
    \country{China}
}

\author{Zhen Hong}
\email{zhong1983@zjut.edu.cn}
\affiliation{%
    \institution{Zhejiang University of Technology}
    \city{Hangzhou}
    \country{China}
}

\author{Cong Wang}
\authornotemark[1]
\email{cwang85@zju.edu.cn}  
\affiliation{%
    \institution{Zhejiang University}
    \city{Hangzhou}
    \country{China}
}

\author{Shouling Ji}
\email{sji@zju.edu.cn}
\affiliation{%
    \institution{Zhejiang University}
    \city{Hangzhou}
    \country{China}
}

\begin{abstract}
Federated Learning (FL), as a mainstream privacy-preserving machine learning paradigm, offers promising solutions for privacy-critical domains such as healthcare and finance. Although extensive efforts have been dedicated from both academia and industry to improve the vanilla FL, little work focuses on the data pricing mechanism. In contrast to the straightforward in/post-training pricing techniques, we study a more difficult problem of pre-training pricing without direct information from the learning process. We propose \projecttitle that integrates a two-stage, auction-based pricing mechanism with a security protocol to address the utility-privacy conflict. Through comprehensive experiments, we show that the client selection according to \projecttitle can achieve more than 10\% higher accuracy in subsequent FL training compared to state-of-the-art methods. In addition, it outperforms the in-training baseline with more than 2\% accuracy increase and 3$\times$ run-time speedup. 
\end{abstract}




\maketitle

\section{Introduction}\label{sec:introduction}
Federated Learning (FL) provides a privacy-preserving paradigm without exposing private data for learning machine learning (ML) models~\cite{deng2021auction,liu2020fedcoin,nagalapatti2021game}. It has found an increasing number of applications such as medical information systems~\cite{dayan2021federated}, financial data analysis~\cite{basu-etal-2021-privacy}, and cross-border corporate data integration~\cite{wan2023closed}. Research and industry efforts on FL primarily focus on improving the accuracy~\cite{li2020federated,wang2020federated}, computation~\cite{lai2021oort,nishio2019client} and communication performance~\cite{perazzone2022communication,ribero2020communication}, as well as enhancing security and privacy~\cite{wang2022protect,bonawitz2017practical}. However, little attention has been paid to incentivizing the participants to join FL, which is crucial because valuable data is not readily available for FL tasks without proper incentives.

\myparagraph{Pre-Training Pricing for FL Data Market}
In traditional centralized ML tasks, servers (such as companies) typically purchase the training data they need from \textit{a data market}. These data are often collected, cleaned, and pre-processed at considerable cost by third parties~\cite{schomm2013marketplaces,liu2014online}. A typical FL data market has three entities: data sellers (clients) who generate data and participate in FL tasks~\cite{spiekermann2015challenges}, model buyers who purchase the model and FL data market (server) that coordinates between sellers and buyers~\cite{zheng2022fl}.

Existing incentive mechanisms mainly target the in/post-training pricing based on the accuracy improvement of FL training~\cite{zhan2020learning,tang2021incentive,zhan2021survey}. Unfortunately, the data providers (FL clients) cannot anticipate the reward before contributing their data and resources. Meanwhile, data buyers cannot evaluate the data quality before the actual training takes place. These are crucial because (1) anticipated rewards before training can greatly encourage the participation of clients, increasing client enrollment and the diversity of training data~\cite{zeng2022incentive,zhan2021survey}. (2) Clients who know the early-estimated pricing before training are more likely to continue, which guarantees the stability of FL training~\cite{lai2021oort}. For instance, based on our empirical survey~\footnote{The complete survey is shown in Appendix \ref{appendix:survey}}, 84.7\% of respondents believe that pre-training incentives increase their willingness to participate in FL. Additionally, 82.7\% of respondents prefer FL training with pre-training rewards over post-training rewards. Therefore, a framework for \textit{pre-training evaluation} and \textit{pricing} is indispensable for a sustainable FL data market.

\myparagraph{Use Case}
A medical research institution aims to develop a model for disease diagnosis and there are several healthcare providers with patient data. The medical research institution first publishes the task in the FL data market with the total payments. Then the task is forwarded to the corresponding healthcare providers to ask for their willingness to join. All healthcare providers who intend to participate in the training task negotiate a satisfactory reward for contributing their data. Hence, the data market collaborates with them to negotiate and establish reasonable pricing as an anticipated reward for subsequent FL training. To this end, the FL marketplace must have the ability to determine the price of each participating client before performing the actual FL training.

\myparagraph{Challenges}
Compared to the conventional in/post-training pricing~\cite{liu2020fedcoin,wang2019measure,song2019profit,zeng2022incentive,zhan2021survey}, implementing pre-training pricing is inherently challenging:

\textit{Challenge 1: Pre-training pricing require to value clients' data without direct model feedback.} The first challenge comes from limited information before training, i.e., we cannot use aggregated model accuracy as direct feedback for pricing and client selection. Instead, we need to peek into the statistics such as volume and categorical distributions, and analyze their correlations with model accuracy in a federated setting~\cite{deng2021auction}. 

\textit{Challenge 2: Pre-training pricing requires a consensus agreement between client and server.}
An optimal pricing system should effectively balance the needs of both the client and the server. From the client's perspective, pricing should not only reflect the intrinsic value of their data but also offer incentives that encourage them to share their data. On the server side, the total compensation should remain within budgetary limits, and the price set for each client should meet or exceed their expected value.

\textit{Challenge 3: Pre-training pricing requires to solve the privacy-utility conflict}. Although data volume and class distribution bring more insights for pre-training pricing, they also ask the clients to share distributional information regarding their private data. E.g., knowing the distribution of human activities would easily reveal individual habits~\cite{qian2020orchestrating,deshpande2004item}, hence deviating from the original intention of FL to preserve privacy.

\myparagraph{Our Solution}
To tackle the above challenges, we present \projecttitle, an fairness, incentive, privacy-preserving client pricing framework for the FL data market. The price is determined through an auction that consists of a two-stage pricing mechanism, i.e., the initial price based on the statistical information and a contribution-proportional allocation strategy. 
The first-phase pricing determine the price of each client based on its statistical information and computed by a data value score function. 
To avoid client's sensitive information leakage, we design a privacy-preserving protocol to enable secured sharing of private distributions with the server. 
In addition, the second-phase pricing determine the actual payment of the selected clients through \textit{Budget-constrained Pricing Mechanism} which is a consensus price between clients and server in terms of fairness and incentive.

\myparagraph{Contributions}
This paper makes the following contributions:
\begin{enumerate}
    \item 
     To the best of our knowledge, \projecttitle is the first framework that addresses the challenges for pre-training pricing in FL data markets, which would significantly motivate both data providers and buyers with a monetary incentive (see Table~\ref{tab:comparsion} in the appendix for a comprehensive list of FL data sharing frameworks).
    \item We propose an effective two-stage data pricing mechanism, in which the pricing of data explicitly reflects its value, the bidding costs of data providers, and the budget constraints of data buyers. We also design a privacy-preserving protocol to secure private information that could be potentially leaked during the bidding process.
    \item We have conducted a series of experiments to evaluate the performance of \projecttitle by selecting clients on three different datasets and in different levels of unbalanced data distributions. Our results show that \projecttitle achieves more than 10\% higher accuracy compared to state-of-the-art pre-training client selection baselines. In addition, it outperforms the in-training baseline with more than 2\% accuracy increase and 3$\times$ runtime speedup.
\end{enumerate}


\section{Related Work}
\label{sec:related}

This section outlines the related works on establishing an FL data marketplace. We focus on two crucial elements for constructing such a marketplace: \emph{Client Pricing } and \emph{Incentive Mechanisms}.

\myparagraph{Client Pricing in FL} A fundamental challenge in pricing clients in FL is the accurate assessment of each client's contribution. There are two main approaches to evaluating clients: model-based and data-based.
For model-based approaches, FedCoin~\cite{liu2020fedcoin} evaluates each client's contribution using the Shapley value. However, computing the Shapley value is time-consuming because it requires calculating the value for all possible client combinations. Other model-based approaches assess each client based on their gradients \cite{gao2021fifl,zhang2021incentive,xu2021gradient} or test performance \cite{li2024performance,deng2022improving,hu2022incentive}. However, these methods introduce significant computational overhead and require in-training feedback.
The data-based approach considers the quality of data to evaluate clients. For instance, AUCTION~\cite{deng2021auction} assesses client contributions based on the data's mislabel rate and size, while Ren~\cite{Ren2024noniidauction} compares the data class distribution with a reference distribution to evaluate client contributions. However, accurately assessing clients prior to training remains an area that requires further exploration.

\myparagraph{Incentive Mechanisms for FL}
Several studies have explored the development of incentive mechanisms for FL~\cite{tang2021incentive,zhan2021survey,mao2024game}, typically assuming a consensus on pricing between servers and clients. However, in real-world applications, information asymmetry between these parties often complicates the establishment of a standardized pricing model. To address this, auction-based mechanisms have been proposed, allowing clients and servers to negotiate prices.
For instance, SARDA~\cite{sun2024socially} introduces a social-aware iterative double auction mechanism to incentivize participation. Similarly, Fair~\cite{deng2021fair} employs a reverse auction model to attract high-quality clients while maintaining a manageable budget for the server. 
Nevertheless, these mechanisms tend to rely on metrics such as model quantity or resource usage, often neglecting critical pre-training information related to user data. This pre-training information necessitates a more rigorous evaluation before training and requires the implementation of a security design that is both effective and privacy-preserving.

\section{Data marketplace pricing Framework for federated learning}
\label{sec:ipp}

In this section, we provide an overview of the framework with the designs of the two-phase pricing mechanism. Table~\ref{tab:notation} in the appendix summarizes the notions of this paper.

\begin{figure}[t]
 \vspace{0mm}
     \centering
    \includegraphics[width=0.42\textwidth]{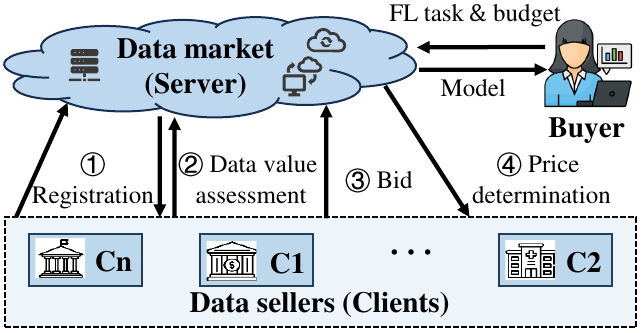}
     \vspace{-3mm}
     \caption{\projecttitle high-level system overview}
     \label{fig:framework}
     \vspace{-6mm}
 \end{figure}
 
\subsection{High-Level System Overview}
\label{subsec:overview}
Our framework consists of three parties: \textit{Data market} (Server), \textit{Data sellers} (Clients) and \textit{Data buyers}. A buyer publishes an FL training task with a budget and the data market groups a set of clients to perform the training task. Finally, the data market returns a trained model to the buyer. Figure~\ref{fig:framework} presents the overview of the \projecttitle pricing framework: 
\textcircled{1} seller first registers the FL task; \textcircled{2} the data market evaluates the data value of each client and returns an assessed price; \textcircled{3} clients make bids (quote) for contributing their data; \textcircled{4} the data market finalizes the price for the clients.

The pricing process consists of three steps: 1) evaluate data for each client by aggregating and getting global data distribution with a privacy-aware secret-sharing (PASS) protocol (see \S \ref{sec:system}); 2) send score function to the clients for computation; 3) receive the scores back at the server. In principle, the clients with unseen classes and sufficient data volume have higher pricing and are regarded as high-quality clients.

\myparagraph{Assumptions}
We target cross-silo FL scenarios across organizations such as healthcare~\cite{dayan2021federated} and finances~\cite{basu-etal-2021-privacy}. We assume that the participants would follow the protocols and refrain from modifying the data or sharing incorrect information with the server~\cite{zhang2021leakage,bonawitz2017practical} since such malicious activities would be quickly detected. On the other hand, the aggregation of individual class distribution would lead to privacy leakage as they often reveal sensitive information such as attribute percentage~\cite{zhang2021leakage}, personal habits~\cite{deshpande2004item}, sales record~\cite{hartmann2023distribution} and financial status~\cite{chen2023protecting}. 
We aim to design an integrated pricing and privacy-preserving framework to resolve the tension between data privacy and data pricing.

\subsection{Single-Client Evaluation}
\label{subsec:first-round}

We evaluate individual client data based on their contribution to the FL training, which is assessed by a score function. 

\begin{figure}[t]
\centering
    \subfloat[Data quantity]{
    \includegraphics[height=.18\textwidth]{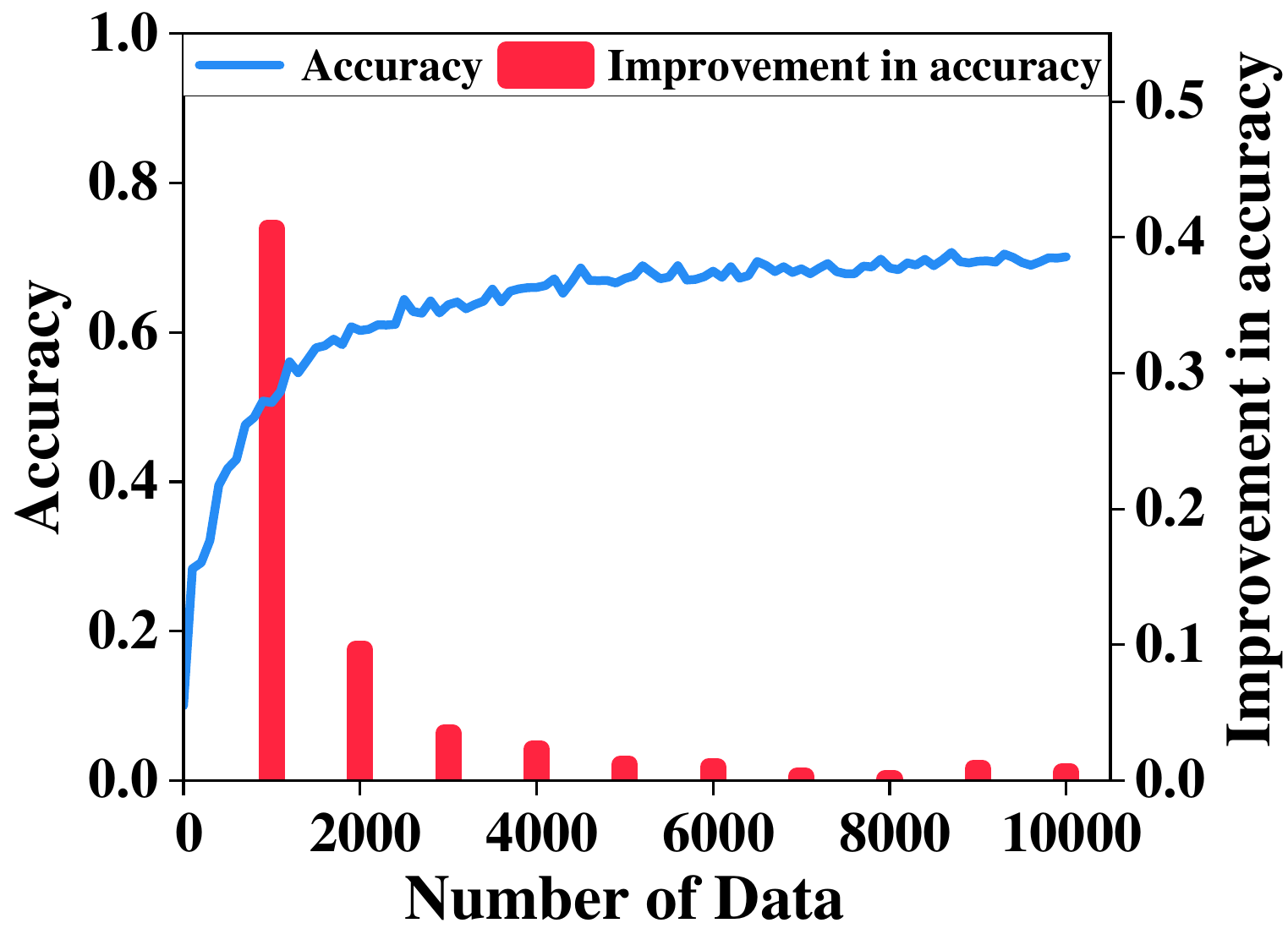}
    \label{fig:Amountandacc1}}
    \subfloat[Data balance]{
    \includegraphics[height=.18\textwidth]{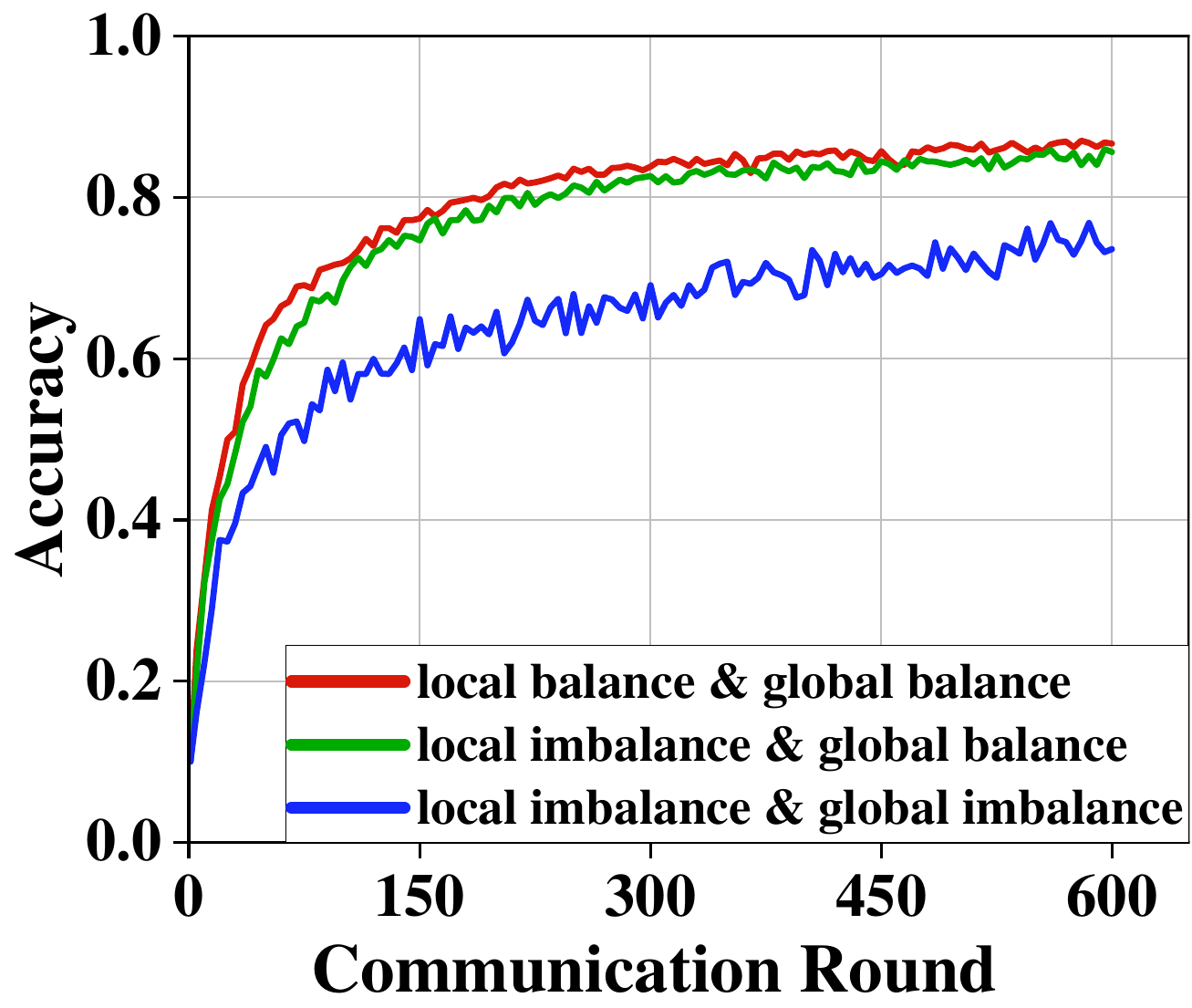}
    \label{fig:motivation_data1}}
    \vspace{-3mm}
     \caption{The impact of data quantity and distribution for FL training: a) 
     An incremental increase of 1,000 data points has shown a diminishing rate of improvement in global accuracy; b) Imbalanced categories in both local and global data result in the worst performance. }
    \label{fig:dsads1}
    \vspace{-4mm}
\end{figure}

\myparagraph{Motivational Example}
To evaluate data quality before training, a common technique is based on statistical information~\cite{deng2021auction,saha2022data,li2021sample}. For FL tasks, we consider two major factors of data quantity and class distribution by evaluating their impact on testing accuracy in Figure \ref{fig:dsads1}.

\noindent \textbf{Observation 1:} \textit{The contribution from client's data increases regarding the data size but with diminishing marginal values.}
Figure~\ref{fig:Amountandacc1} shows that the test accuracy climbs up with the increase of training data. However, $\Delta Accuracy$ drops rapidly, indicating that the contribution from data amount to test accuracy is marginal as the data amount increases.

\noindent \textbf{Observation 2:} \textit{Clients with data in scarce categories can significantly improve the overall FL accuracy.}
Figure~\ref{fig:motivation_data1} shows that local and global data imbalance significantly deteriorates the test accuracy. If the global data is balanced by bringing more clients with scarce categories (the green curve in Figure~\ref{fig:motivation_data1}), the accuracy degradation for FL training is much smaller. This aligns with the findings in~\cite{wang2021tpds} that unseen categories help increase gradient diversity and improve model generalization.

\myparagraph{Score Function}
To calculate the score for each client, the server needs to obtain the data volume and categorical distribution from the clients in advance (privacy risks are addressed in Section~\ref{sec:system}). For class $c$ of $C$ categories, the $c$-class volume is represented as $n_s^c$. The total amount of data $N_s$ is the summation from all the classes: $N_s = \sum_{c=1}^C n_s^c$. Each client $e$ has the amount of $N_e$ data. For each class $c$ from the same set of $C$ categories, the volume of $c$ at client $e$ is denoted as $n_e^c$. As a result, we can define $N_e$ as the sum of volumes across all classes: $N_e = \sum_{c=1}^C n_e^c$. The score function on client $e$ is,
\begin{equation}
\small
\label{eq:evaluation_function}
    u_e =\sum_{c=1}^{C} \theta_{c} \cdot \phi(n_{e}^{c}), 
\end{equation}
where $\theta_{c}$ represents the unique coefficient for this category and $\phi(\cdot)$ models the relation between input data and model accuracy. The score function can be interpreted as the sum of the $\phi(\cdot)$ function, with the data volume of each category as the input, weighted by $\theta_{c}$ for the different categories.

$\phi(\cdot)$ is formalized as equation~\ref{eq:quantity_coe}.
\begin{equation}
    \small
    \begin{split}
        \phi(x) =\sum_{t=1}^x f({\rho(t)}) \qquad
        s.t. \quad \rho(t) = \min (t/\alpha, 1),
    \label{eq:quantity_coe}
    \end{split}
\end{equation}
where $x$ takes the value of $n_e^c$ for each client and $\rho(\cdot)$ is the normalising function, constrained within the range of $(0, 1]$. $\alpha$ is a threshold parameter with $\alpha = N_s / (E \cdot C)$, which is the average data volume for each category across all the clients. If $n_e^c > \alpha$, we consider that an increase in data volume $n_e^c$ does not provide additional benefits, which aligns with \textit{Observation 1} in Section \ref{subsec:overview}. 

To model the relationship between data volume and accuracy, we draw inspiration from~\cite{ben2010theory} and empirical observations~\cite{kaplan2020scaling}, which suggest an $O(\log(x))$ contribution of data volume to accuracy. Consequently, we use the negative natural logarithm $-\ln (\cdot)$ as the function $f(\cdot)$ to map data volume to accuracy. The rationale behind the choice of $f(\cdot)$ is detailed in Appendix \ref{sec:sen_analysis}.

For the coefficient $\theta_c$, it can be calculated as:
\begin{equation}
\small
\theta_c =  1 - n_s^c / N_s .
\end{equation}
It reflects that if a category is relatively scarce in the global distribution, its weighted coefficient would be higher, which satisfies the demands from \textit{Observation 2}.

\subsection{Reaching a Consensus Price between Client and Server}
\label{subsec:second-round}

As a data market, we should satisfy both data sellers and buyer. The score function computes a value for each client which meets server's requirements. To satisfy clients requirements while ensuring the payments are within the given budget, we propose a second-phase pricing mechanism as follows.

\myparagraph{Bidding Mechanism}
We incorporate the design of the classical bidding mechanism~\cite{mcafee1987auctions,lin2021multi,zheng2016budget}, where the server $S$ orchestrates the bidding procedures with the budget $R$. A client pool $\mathbb{E}$ consists of $E$ clients. The bidding procedures are designed to choose the most valuable clients within the budget constraint of $R$. We break down the bidding mechanism into two parts: winner selection and payment determination. Specifically, at the beginning of the bidding, client $e$ decides the bid $b_e$ according to the evaluation score and the potential cost of the task. The set of bids of $E$ clients is $\mathbb{B} = \{b_e\}_{e \in \mathbb{E}}$. The server selects the winning clients to join the task with the budget constraints $R$. And then decide the final payment $p_e$ for each winner $e$ with budget constraints.

\myparagraph{Winner Selection}
The procedure of the winner selection part is detailed in Algorithm~\ref{algorithm1} line~\ref{code:winner_begin}-\ref{code:winner_end}. First, the server sorts clients by their score per bid and gets the list $\mathbb{V}$. Then, the server selects the winners in the order of $\mathbb{V}$. In particular, only clients with bids $b_e$ smaller than the budget constraint condition $\frac{R}{2}\cdot \frac{u_e}{U(\mathbb{S}_k\cup \{e\})}$, $e$ are eligible as a winner, a criterion also adopted in ~\cite{zheng2016budget}. If a client's bid exceeds this constraint, the winner selection process terminates, and subsequent clients are not considered as winners.

\myparagraph{Payment Determination}
After the winners are selected, the server determines the payments for each of them. The basic idea of the payment determination can be described as follows. For each winner $e$, we consider a new list $\mathbb{V}^{-e}$ which eliminates the client $e$ from $\mathbb{V}$. 
Select winners from the list $\mathbb{V}^{-e}$ similar to the winner selection part. We assume that client $e$ replaces client $j$ as the winner in list $\mathbb{V}^{-e}$ and calculate the maximum bid $p_{e(j)}^{\prime}$ at which $e$ can win the auction in position $j$. However, each $j$ corresponds to a different $p_{e(j)}^{\prime}$. Hence, we take the maximum value of $p_{e(j)}^{\prime}$ as the final payment for $e$.

The procedure of the payment determination part is detailed in Algorithm~\ref{algorithm1} line~\ref{code:payment_begin}-\ref{code:payment_end}. Similar to the winner selection, we select the client in order of list $\mathbb{V}^{-e}$. Then compute the maximum bid $p_{e(j)}^{\prime}$ that client $e$ can provide to win the auction in list $\mathbb{V}^{-e}$. The bid $b_{e(j)}$ should satisfy two conditions.
\begin{itemize}
    \item Client $e$ should have more larger score per bid than $j$, i.e.,
    \begin{equation}\label{eq:b_e_condidtion1}
        \frac{u_e}{b_{e(j)}}\geq \frac{u_j}{b_j} \Rightarrow
        b_{e(j)}\leq \frac{u_e\cdot b_j}{u_j}=\lambda_{e(j)}.
    \end{equation}
    \item The bid of client $e$ should satisfy the budget constraint condition, i.e.,
    \begin{equation}\label{eq:b_e_condition2}
        b_{e(j)}\leq \frac{R}{2}\cdot \frac{u_e}{U(\mathbb{S}_{j-1}^{\prime}\cup \{e\})}=\beta_{e(j)}.
    \end{equation}
\end{itemize}
We set $\hat{k}$ as the smallest index which satisfies the budget constraint in $\mathbb{V}^{-e}$, i.e., 
       $ b_{\hat{k}+1}> \frac{R}{2}\cdot \frac{u_{\hat{k}+1}}{U(\mathbb{S}_{\hat{k}+1}^{\prime})}$.
Therefore, the maximum index in $\mathbb{V}^{-e}$ that $e$ can replace as the winner is $\hat{k}+1$.
Since the bid should satisfy both of the above two conditions, we get the maximum of the bid is $p_{e(j)}^{\prime}=\min\{\lambda_{e(j)},\beta_{e(j)}\}$. In Inequality (\ref{eq:b_e_condidtion1}), the value of $\lambda_{e(j)}$ monotonically decreases with the index $j$. In Inequality (\ref{eq:b_e_condition2}), the value of $\beta_{e(j)}$ is dynamically changing with $j$. Therefore, we determine the maximum of $p_{e(j)}^{\prime}$ in different $j$ for $j\in[1,\hat{k}+1]$ as the final payment. i.e., 
    $p_e=\max_{1\leq j\leq \hat{k}+1}\{p_{e(j)}^{\prime}\}$.

\begin{algorithm}[t]
    
    \caption{Budget-constrained Pricing Mechanism}
    \label{algorithm1}
    \textbf{Input:} 
    Candidate clients $\mathbb{E}$, Budget $R$, Bids $\mathbb{B} = \{b_e\}_{e \in \mathbb{E}}$ and scores $\mathbb{U} = \{u_e\}_{e \in \mathbb{E}}$\\
    \textbf{Output:} Selected clients $\mathbb{S}_k$ and Payment $\mathbb{P}$\\
    \textbf{Winner Selection:}\label{code:winner_begin}\\ 
    Sort according score per bid get $\mathbb{V}$: $\frac{u_1}{b_1}\geq...\geq \frac{u_e}{b_e} \geq...\geq \frac{u_E}{b_E}$\\
    $\mathbb{S}_k\leftarrow \emptyset$;
    $U(\mathbb{S}_k)=\sum_{e\in \mathbb{S}_k}u_e$\\
    \For{$\frac{u_e}{b_e}$ in $\mathbb{V}$}{
        \If{$b_e > \frac{R}{2}\cdot \frac{u_e}{U(\mathbb{S}_k\cup \{e\})}$\label{code:budget_constraint}}{
            break;
        }
        $\mathbb{S}_k\leftarrow \mathbb{S}_k \bigcup \{e\}$; \\ \label{code:add_winner}
    }\label{code:winner_end}
    \textbf{Payment Determination:}\label{code:payment_begin}\\ 
    \For{$e$ in $\mathbb{S}_k$}{
        $j\leftarrow 1$;
        $\mathbb{S}_{j-1}^{\prime}\leftarrow \emptyset$;
        $p_e\leftarrow 0$;
        $\mathbb{V}^{-e}$: $\frac{u_1}{b_1}\geq...\geq \frac{u_{E-1}}{b_{E-1}}$\\
        \For{$\frac{u_j}{b_j}$ in $\mathbb{V}^{-e}$}{
            \If{$b_j>\frac{R}{2}\cdot \frac{u_j}{U(\mathbb{S}_{j-1}^{\prime}\cup \{j\})}$}
            {
                break;
            }
            $\lambda_{e(j)}\leftarrow\frac{u_e\cdot b_j}{u_j}$; 
            $\beta_{e(j)}\leftarrow \frac{R}{2}\cdot \frac{u_e}{U(\mathbb{S}_{j-1}^{\prime}\cup \{e\})}$;\\
            $p_{e(j)}^{\prime}=\min\{\lambda_{e(j)},\beta_{e(j)}\}$; 
            $p_e=\max\{p_e,p_{e(j)}^{\prime}\}$;\\
            $\mathbb{S}_{j}^{\prime}\leftarrow\mathbb{S}_{j-1}^{\prime}\bigcup\{j\}$; $j\leftarrow j+1$\\
        }
        $\lambda_{e(j)}\leftarrow\frac{u_e\cdot b_j}{u_j}$; $\beta_{e(j)}\leftarrow \frac{R}{2}\cdot \frac{u_e}{U(\mathbb{S}_{j-1}^{\prime}\cup \{e\})}$;\\
        $p_e\leftarrow\max\{p_e,\min\{\lambda_{e(j)},\beta_{e(j)}\}\}$;\label{code:payment_end}\\
    
    }
    $\mathbb{P} = \{p_e\}_{e\in \mathbb{S}_k}$\\
\Return{$\mathbb{S}_k$ and $\mathbb{P}$}

\end{algorithm}

\myparagraph{A Walk-Through Example}
Consider an example with a client pool of $E = 4$ clients and a budget constraint $R = 140$.
The scores and bids of all clients in $\mathbb{E}$ is $\mathbb{U}=\{5,6,10,20\}$ and $\mathbb{B}=\{10,13,80,45\}$. 
Then calculate each score per bid and sort clients get list $\mathbb{V}$:
$$\{\frac{u_1}{b_1}:\frac{5}{10}=0.5; \frac{u_2}{b_2}:\frac{6}{13}=0.46; \frac{u_3}{b_3}:\frac{20}{45}=0.44; \frac{u_4}{b_4}:\frac{10}{80}=0.125\}$$

Sequential client selection according to list $\mathbb{V}$, we get
\begin{align*}\nonumber
    b_1<\frac{R}{2}\frac{u_1}{U(\emptyset \cup \{1\})}=70;\quad
    &b_2<\frac{R}{2}\frac{u_2}{U(\{1\} \cup \{2\})}=38.2;\\
    b_3<\frac{R}{2}\frac{u_3}{U(\{1,2\} \cup \{3\})}=45.2; \quad
    &b_4>\frac{R}{2} \frac{u_4}{U(\{1,2,3\} \cup \{4\})}=17.1.
\end{align*}
Since $b_4>\frac{R}{2}\cdot \frac{u_4}{U(\{1,2,3\} \cup \{4\})}$, the winner set $\mathbb{S}_k=\{1,2,3\}$. 
Then the server determines the payments for them. For client $1$, $\mathbb{V}^{-1}:\{\frac{u_2}{b_2};\frac{u_3}{b_3};\frac{u_4}{b_4}\}$, then we select clients from $\mathbb{V}^{-1}$ one by one until violate budget constraints. Client \{2,3\} satisfy the budget constraints in $\mathbb{V}^{-1}$.
When select to client $2$, 
$$\lambda_{1(2)}=\frac{u_1\cdot b_2}{u_2}=10.9; \beta_{1(2)}=\frac{140}{2}\times \frac{u_1}{0+u_1}=70; p_{1(2)}^{\prime}=10.9.$$
Similarly, $p_{1(3)}^{\prime}=\min\{11.4,31.8\}=11.4$, $p_{1(4)}^{\prime}=\min\{40,11.3\}=11.3$. Then the final payment of client $1$ is $p_1=p_{1(3)}^{\prime}=11.4$. Similar to the client $1$, we calculate each winner payment $p_2=p_{2(3)}^{\prime}=13.6$, $p_3=p_{3(4)}^{\prime}=45.2$. And the total payment is $70.2<R$.

\myparagraph{Property of the Mechanism}
A reasonable bidding mechanism needs to satisfy \textit{Truthful}, \textit{Individual Rationality}, and \textit{Budget Constraints}. Next, we prove our mechanism satisfies these properties.

\begin{theorem}\label{th:truthful}
    A bidding mechanism is truthful if and only if~\cite{singer2010budget}:
    \begin{enumerate}
        \item The selection algorithm is monotone, i.e. if $e$ wins the bidding by $b_e$, it would also win by bidding $b_e^{\prime} < b_e$;
        \item Each winner is paid at the critical value: $e$ would not win the bidding if $b_e^{\prime} > p_e$.
    \end{enumerate}
\end{theorem}

\begin{lemma}\label{lemma:monotone}
    The algorithm for winner selection is monotone.
\end{lemma}
\begin{proof}
    The proof is given in the Appendix~\ref{appendix:proof_of_monotone}
\end{proof}
    

\begin{lemma}\label{lemma:critical_price}
    The payment $p_e\in \mathbb{P}$ is the critical price of auction winner $e\in\mathbb{S}_k$.
\end{lemma}

\begin{proof}
    The proof is given in the Appendix~\ref{appendix:proof_of_critical_price}
\end{proof}

\begin{theorem}
    The bidding mechanism is Truthful.
\end{theorem}
\begin{proof}
    According to the Lemmas~\ref{lemma:monotone},~\ref{lemma:critical_price} and Theorem~\ref{th:truthful}, our bidding mechanism satisfies the monotone and the final payment is the critical price. Thus, the bidding mechanism is \textit{Truthful}.
\end{proof}

\begin{theorem}
    The auction satisfies Individual Rationality.
\end{theorem}
    
\begin{proof}
    If the payment for client $e$ is larger than its bid $b_e$, the auction is \textit{Individual Rationality}. let's compare the bid $b_e$ with $p_{e(\gamma)}^{\prime}$, where the index $\gamma$ in $\mathbb{V}^{-e}$ is same with $e$ in $\mathbb{V}$. Therefore, the winners before $e$ in $\mathbb{V}$ is same with the winners before $\gamma$ in $\mathbb{V}^{-e}$.
    We know the payment for client $e$ is the maximum over all possible $p_{e(j)}^{\prime}$ for $j\in[1,\hat{k}+1]$, thus $p_{e(\gamma)}^{\prime}\leq p_e$. According to the winner selection part, we know $b_e$ satisfied the budget constraint, i.e.,
    \begin{equation}\label{eq:b_e_beta}
         b_e\leq \frac{R}{2}\cdot\frac{u_e}{U(\mathbb{S}_{e-1}\cup \{e\})}=\frac{R}{2}\cdot\frac{u_e}{U(\mathbb{S}_{\gamma-1}^{\prime}\cup \{e\})}=\beta_{e(\gamma)}.  
    \end{equation}
    In list $\mathbb{V}$, $\gamma$ is behind $e$ then we get
    \begin{equation}\label{eq:b_e_lambda}
        \frac{u_e}{b_e}\geq \frac{u_{\gamma}}{b_{\gamma}}\Rightarrow b_e\leq \frac{u_e\cdot b_{\gamma}}{u_{\gamma}}=\lambda_{e(\gamma)}.
    \end{equation}
    Recall that $p_{e(\gamma)}^{\prime}\leq p_e$ and according to inequalities (\ref{eq:b_e_beta},\ref{eq:b_e_lambda}), we get $b_e\leq \min\{\lambda_{e(\gamma)},\beta_{e(\gamma)}\}=p_{e(\gamma)}^{\prime}\leq p_e$.
    Therefore, the payment for the winner $e$ is always larger than its bid $b_e$ and the auction is \textit{Individual Rationality}.
\end{proof}

\begin{lemma}\label{lemma:extra_lemma}
    For clients set  $\mathbb{S}_1\subset\mathbb{S}_2\subseteq\mathbb{S}$, if $\hat{e} = \arg\max_{e\in\mathbb{S}_2\setminus\mathbb{S}_1}\frac{u_e}{b_e}$ then the following inequality is valid.
       \begin{equation}\label{eq:theorem_subset}
            \frac{U(\mathbb{S}_2)-U(\mathbb{S}_1)}{\sum_{i\in\mathbb{S}_2}b_i-\sum_{j\in\mathbb{S}_1}b_j}< \frac{u_{\hat{e}}}{b_{\hat{e}}} 
       \end{equation}
\end{lemma}
\begin{proof}
    The proof is given in the Appendix~\ref{appendix:proof_of_lemma}.
\end{proof}

\begin{theorem}
    \label{lemma_budget_cons}
    The mechanism is within the budget constraint.
\end{theorem}

\begin{proof}
   We try to show that the auction satisfies the budget constraint by proving that the upper bound of the payment $p_e$ is $\frac{u_e}{U(\mathbb{S}_k)}R$. We prove it by contradiction, assume $p_e>\frac{u_e}{U(\mathbb{S}_k)}R$. According to the payment determination mechanism above, we know the payment $p_e$ satisfies the following conditions:
   \begin{equation}\label{eq:payment_condition}
       p_e\leq \frac{u_e\cdot b_r}{u_r} \qquad p_e\leq \frac{R}{2}\frac{u_e}{U(\mathbb{S}_{r-1}^{\prime}\cup \{e\})} 
   \end{equation}
   Since $e$ is the $e$-th winner, thus for $j\in [1,e-1]$, $b_e>\lambda_{e(j)}$. We know that $p_{e(j)}^{\prime}\leq \lambda_{e(j)}$, then we get $b_e>\lambda_{e(j)}\geq p_{e(j)}^{\prime}$. In Theorem \ref{lemma:critical_price} we prove the bid of $e$ is no larger than the final payment, i.e., $b_e\leq p_e$. Therefore, we get the inequality:
   \begin{equation}\label{eq:r_is_not_in}
       p_{e(j)}^{\prime}<b_e\leq p_e=p_{e(r)}^{\prime},\quad j\in [1,e-1]
   \end{equation}
   From the Inequality (\ref{eq:r_is_not_in}) we know $r$ is not in list $[1,e-1]$, so $r$ is behind $e$ and we have $\mathbb{S}_{e-1} \subseteq \mathbb{S}_{r-1}^{\prime}$.
   Let's consider the following two scenarios:
   \begin{itemize}[leftmargin=10pt]
       \item $\mathbb{S}_{r-1}^{\prime}\bigcup\{e\}=\mathbb{S}_{r-1}^{\prime}\bigcup\mathbb{S}_{k}$.
        Since $\mathbb{S}_{e-1} \subseteq \mathbb{S}_{r-1}^{\prime}$, the Inequality (\ref{eq:payment_condition}) can be rewritten as:
        \begin{equation}
        \label{eq:budget_s1_last}
            \frac{u_e}{p_e}\geq \frac{ 2U(\mathbb{S}_{r-1}^{\prime}\bigcup\{e\})}{R}=\frac{2U(\mathbb{S}_{r-1}^{\prime}\bigcup \mathbb{S}_k)}{R}\geq \frac{2U(\mathbb{S}_k)}{R}
        \end{equation}
        Therefore, according to Inequality (\ref{eq:budget_s1_last}) we get $p_e\leq \frac{u_e}{U(\mathbb{S}_k)}\cdot R$ which contradicts the assumption.
        Thus the assumption is not valid.
        
       \item $\mathbb{S}_{r-1}^{\prime}\bigcup\{e\}\subset\mathbb{S}_{r-1}^{\prime}\bigcup\mathbb{S}_{k}$. 
       Set $\mathbb{S}_1=\mathbb{S}_{r-1}^{\prime}\bigcup\{e\}, \mathbb{S}_2=\mathbb{S}_{r-1}^{\prime}\bigcup\mathbb{S}_{k}$ and $\mathbb{S}_1\subset\mathbb{S}_2$. Assume $\hat{r}=\arg\max_{t\in\mathbb{S}_2\setminus\mathbb{S}_1}\frac{u_t}{b_t}$, according to the Inequalities (\ref{eq:theorem_subset},\ref{eq:payment_condition}) and Lemma \ref{lemma:extra_lemma} we get:
      \begin{equation}
           \frac{U(\mathbb{S}_2)-U(\mathbb{S}_1)}{\sum_{i\in\mathbb{S}_2}b_i-\sum_{j\in\mathbb{S}_1}b_j}< \frac{u_{\hat{r}}}{b_{\hat{r}}}\leq \frac{u_e}{p_e}
      \end{equation}
      Since we previously assume $p_e>\frac{u_e}{U(\mathbb{S}_k)}\cdot R$, thus $\frac{u_e}{p_e}<\frac{U(\mathbb{S}_k)}{R}$.
      We know that $b_e\leq p_e\leq \frac{R}{2}\frac{u_e}{U(\mathbb{S}_{r-1}^{\prime}\cup \{e\})}$ (Inequality \ref{eq:payment_condition}). Then:
      \begin{equation}\label{eq:all_client_smaller_than Sk}
          \frac{u_e}{b_e}\geq \frac{ 2U(\mathbb{S}_{r-1}^{\prime} \cup \{e\})}{R}
          \Rightarrow \frac{u_k}{b_k}\geq \frac{ 2U(\mathbb{S}_k)}{R}
      \end{equation}
      According to the Inequality (\ref{eq:all_client_smaller_than Sk}), we can get inequality as follows.
      \begin{equation}
          \frac{u_1}{b_1}\geq\frac{u_2}{b_2}\geq \dots \geq  \frac{u_k}{b_k} \geq \frac{ 2U(\mathbb{S}_k)}{R}
      \end{equation}
      Then we get $b_e\leq \frac{R}{2} \cdot \frac{u_e}{U(\mathbb{S}_k)}$, then $\sum_{e\in \mathbb{S}_k}b_e\leq \frac{R}{2} \cdot \frac{\sum_{e\in \mathbb{S}_k}u_e}{U(\mathbb{S}_k)}=\frac{R}{2}$. 
      Therefore, we can get :
      \begin{equation}
        \sum_{i\in\mathbb{S}_2}b_i-\sum_{j\in\mathbb{S}_1}b_j=\sum_{e\in\mathbb{S}_2\setminus\mathbb{S}_1}b_e\leq \sum_{e\in\mathbb{S}_k}b_e \leq \frac{R}{2}    
      \end{equation}
     Recall that $\mathbb{S}_2=\mathbb{S}_{r-1}^{\prime}\bigcup\mathbb{S}_{k}$, thus $\mathbb{S}_k\subseteq\mathbb{S}_2$. Then we get the Inequality (\ref{eq:long_inequality}).
     \begin{equation}\label{eq:long_inequality}
     \begin{aligned}
         \frac{2 (U(\mathbb{S}_k)-U(\mathbb{S}_1))}{R}
         &\leq \frac{2 (U(\mathbb{S}_2)-U(\mathbb{S}_1))}{R}\\
         &\leq \frac{U(\mathbb{S}_2)-U(\mathbb{S}_1)}{\sum_{e\in \mathbb{S}_2\setminus\mathbb{S}_1}b_e}\leq \frac{u_e}{p_e}
         <\frac{U(\mathbb{S}_k)}{R}
     \end{aligned}
     \end{equation}
    Then we can deduce from the above inequality that $2(U(\mathbb{S}_k)-U(\mathbb{S}_1))<u(\mathbb{S}_k)$. Thus
    \begin{equation}
    \begin{aligned}
     &U(\mathbb{S}_k)< 2U(\mathbb{S}_1)=2 U(\mathbb{S}_{r-1}^{\prime} \cup \{e\})\\
     \Rightarrow &\frac{u_e}{p_e}\geq \frac{ 2U(\mathbb{S}_{r-1}^{\prime}\cup \{e\})}{R}\geq \frac{U(\mathbb{S}_k)}{R}
    \end{aligned}
    \end{equation}
    From the inequality above, we can conclude that $p_e\leq \frac{u_e}{U(\mathbb{S}_k)}\cdot R$ which contradicts the assumption. 
   \end{itemize}
   
   Therefore, the assumption is invalid and the upper bound of the payment is $\frac{u_e}{U(\mathbb{S}_k)}\cdot R$. So, $\sum_{e\in\mathbb{S}_k}p_e\leq \frac{\sum_{e\in\mathbb{S}_k}u_e}{U(\mathbb{S}_k)}\cdot R=R$, satisfying budget constraints.
\end{proof}

\section{Privacy-aware Secret-sharing Mechanism}
\label{sec:system}

In this section, we first discuss the design idea and primitives for the PASS protocol in \S~\ref{sec:preliminary}. Then, in \S~\ref{subsec:protocol_security} we introduce the implementation of PASS for obtaining the global data distribution. Finally, we analyse the security of the proposed protocol.

\vspace{-2mm}
\subsection{Design of PASS}
\label{sec:preliminary}

\myparagraph{Main Idea} 
To protect the sensitive information of clients, we leverage the Diffie-Hellman key agreement~\cite{hellman1976new} and distribution aggregation method to generate a pair of \emph{positive and negative} random noises to obfuscate the local data distribution before sharing them with the server.
Thereafter, we sum up these obfuscated local distributions and the added noises can be cancelled out, reminding the global distribution.
To be precise, we construct a random seed based on key pairs $(SK_e,PK_v)$ in client $e$ and $(SK_v,PK_e)$ in client $v$, which the public keys $PK_e,PK_v$ are distributed by client $e$ and $v$ respectively. Using the above key pairs, we can generate the same random seed $s_{e,v}$, which can be used to generate a \emph{noise} by a pseudorandom generator (PRG)~\cite{yao1982theory} for both client $e$ and $v$, i.e., $PRG(s_{e,v})$. The $PRG(s_{e,v})$ that has the same dimensions as the host clients' local distribution (i.e., $|\mathbb{N}_e|==|PRG(s_{e,v})|$ and $|\mathbb{N}_v|==|PRG(s_{e,v})|$) will be added to $\mathbb{N}_e$ and $\mathbb{N}_v$. Therefore, we develop a distribution aggregation method that constructs a pair of \emph{positive and negative} of noises, which can be cancelled after aggregation on the server side.

\myparagraph{Key Agreement}
\label{sec:preliminary sharing}
In this paper, we use Diffie-Hellman key agreement to achieve pairwise clients agreeing on a random seed that not be disclosed by the server or clients. The key agreement consists of a tuple of algorithms ($KA.param,KA.gen,KA.agree$). $KA.param (r)\\ \rightarrow R $ generate a public parameter $R$ based on the security parameter $r$. 
$KA.gen(R)\rightarrow (SK,PK)$ uses the public parameter $R$ to produce a private-public key pair, and $KA.agree(SK_e,PK_v)\rightarrow s_{e,v}$ can generate the identical private shared key $s_{e,v}$ using the private key of $e$ and public key of $v$ which are generated from the same public parameter $R$. As a result, we can have two key pairs that generate the same random seed, i.e.,  $KA.agree(SK_e,PK_v)=KA.agree(SK_v,PK_e)=s_{e,v}$.

\myparagraph{Distribution Aggregation} We assume that each client $e\in \mathbb{E}$ possesses a $C$-dimensional private vector $\mathbb{N}_e=\{n_e^c\}_{c\in\{1,...,C\}}$ indicating their local data distribution. 
The proposed secret share method aims to enable the server $S$ to securely compute global data distribution $\mathbb{N}_s=\sum_{e\in \mathbb{E}}\mathbb{N}_e$ without accessing the private local distributions $\mathbb{N}_e$.

We first assign each client $e$ a unique identifier from \{$1$ to $E$\}, pairing all clients in a pairwise manner, denoted as $(e,v)$. Then, we use the PRG to generate the identical random vectors $PRG(s_{e,v})$ based on the random seed $s_{e,v}$. Furthermore, we introduce the $\epsilon_{e,v}$ to determine whether to add or subtract random vectors $PRG(s_{e,v})$. When $e$ is less than $v$ (i.e., $e<v$), the $\epsilon_{e,v}$ equals to 1. On the contrary, if $e$ is greater than $v$ (i.e., $e>v$), the $\epsilon_{e,v}$ equals to -1.
Thereafter, we use Equation~(\ref{eq:distribution_y}) to compute an alternative distribution of $\mathbb{N}_e$ (i.e., $\mathbb{Y}_e$) by adding all $\epsilon_{e,v}\cdot PRG(s_{e,v})$ generated by client $e$ paired with other clients $v\in \mathbb{E}$ to avoid $\mathbb{N}_e$ being disclosed.

\begin{equation}\label{eq:distribution_y}
\small
    \mathbb{Y}_e=\mathbb{N}_e+\sum_{v\in \mathbb{E}} \epsilon_{e,v} \cdot PRG(s_{e,v}); \; \epsilon_{e,v} =\left\{
\begin{aligned}
1  & \hspace{5mm} e<v \\
-1  & \hspace{5mm} e>v
\end{aligned}
\right.
\end{equation}

Once the server $S$  obtains all $\mathbb{Y}_e$, the global data distribution $\mathbb{N}_s$ can be computed via Equation~(\ref{eq:global}), where the added $PRG(s_{e,v})$ are cancelled each other out, reminding the global data distribution $\mathbb{N}_s$.

\begin{equation}
\begin{aligned}
\small
      \mathbb{N}_s=\sum_{e\in \mathbb{E}}\mathbb{Y}_e
      =\sum_{e\in \mathbb{E}}[\mathbb{N}_e+\sum_{v\in \mathbb{E}} \epsilon_{e,v} \cdot PRG(s_{e,v})]
      =\sum_{e\in \mathbb{E}}\mathbb{N}_e  
\end{aligned}
\label{eq:global}
\end{equation}


\subsection{The PASS Protocol}
\label{subsec:protocol_security}

Table \ref{tab:protocol} in appendix~\ref{pass_protocol} shows the protocol of the $PASS$ to aggregate data distribution from clients. 
In step 1, each client uses the given security parameters $r$ to generate the public parameters $R$.
Then, client $e$ uses $R$ to generate key pairs $(PK_e, SK_e)$.
After that, client $e$ sends its public key $PK_e$ to the server.
The server received $PK_{e,e\in \mathbb{E}}$ from each client, and the server broadcast $(PK_e,e)$ to each client when received all $PK_e$. 

In step 2, client $e$ received all $PK_v$ and its client identifier $v$ from the server. Then, it uses $PK_e$ and the private key $SK_e$ to generate random seed $s_{e,v}$ through the $KA.agree$ algorithm. Based on $s_{e,v}$, a $C$-dimensional random vector $\mathbb{RV}_{e,v}\leftarrow \epsilon_{e,v} \cdot PRG(s_{e.v})$ is generated via PRG, where $\epsilon_{e,v}=1$ if $e<v$ and $\epsilon_{e,v}=-1$ if $e>v$. Then, the client $e$ adds the local distribution $\mathbb{N}_e$ with all $\mathbb{RV}_{e,v} $ to obtain the pseudo local data distribution $\mathbb{Y}_e$ and then sends it to the server.
The server sums all $\mathbb{Y}_e$ from all client $e \in \mathbb{E}$ to get the global distribution $\mathbb{N}_s$ when receiving all $\mathbb{Y}_e$. 
Finally, the server broadcasts $\mathbb{N}_s$ to all clients for data evaluation. A running example of $PASS$ is included in appendix~\ref{pass_protocol}.

\myparagraph{Performance and Security Analysis} 
The communication complexity of each client and server is $O(E)$ and $O(E^2)$, respectively. This arises from each client's requirement to receive public keys from the other $E-1$ clients. For communication costs, the actual communication cost is negligible compared to FL training since the size of the keys is much smaller than the weights of the model. Additionally, the PASS can protect the clients' privacy in the semi-honest client environment. 
The detailed analysis of the communication cost and security is shown in Appendix~\ref{communication_analyze} and ~\ref{sec:analyse of privacy}.

\section{Evaluation}
\label{sec:evaluation}
In this section, we evaluate the impact of the client selection on the subsequent FL training accuracy by comparing \projecttitle to four state-of-the-art pre-training client selection methods. In addition, we compared \projecttitle with an in-training client selection algorithm, which can effectively use the training feedback. Experimental results show that, \projecttitle outperforms other state-of-the-art pre-training client selection methods in most cases. Even when compared to the in-training client selection algorithm, \projecttitle still achieves competitive accuracy with less runtime overhead.

\subsection{Experimental Setup}
\label{sec:expsetup}

\myparagraph{Applications, Datasets and Models}
We evaluate \projecttitle in two applications: Image Classification (IC) and Human Activity Recognition (HAR). For IC, we utilize the CIFAR-10~\cite{krizhevsky2009learning} and CINIC-10~\cite{darlow2018cinic} datasets with a ResNet-56~\cite{he2016deep} model. For HAR, we employ the DEAP dataset~\cite{koelstra2011deap} using peripheral physiological signals and train a three-layer CNN model on it.

\myparagraph{FL Training Set Up}
We evaluated \projecttitle under different numbers of clients, including both 20 and 100 clients. Each client's data distribution is generated through a two-step process. Firstly, we adjust the global data distribution by randomly pruning data from different classes to achieve an unbalanced global distribution. In the second step, based on the different global data distributions, we allocate data for each class to each client following a Dirichlet distribution ($\alpha=0.5$) for simulating Non-i.i.d. scenarios~\cite{hsu2019measuring,li2022federated,lin2020ensemble}.
Appendix~\ref{sec_dataset} shows the example of the generated dataset on CIFAR-10. In summary, our evaluation encompasses three datasets, each tested under three selection ratios and six distributions (seven distributions for CINIC-10), resulting in a total of 57 test cases. We trained all FL tasks for 600 rounds, with an initial learning rate of 0.001 for CINIC-10 and 0.003 for CIFAR-10 and DEAP.

\myparagraph{Baselines}
\projecttitle focuses on pre-training data evaluation in a more challenging setting, where no direct feedback is available from the training process. To align with this focus, we selected four primary baseline methods: random selection (RS), quantity-based selection (QBS)~\cite{zeng2020fmore}, DICE~\cite{saha2022data}, and diversity-driven selection (DDS)~\cite{li2021sample}, all of which are based on pre-training data evaluation. It is worth noting that, while more recent works on model-based evaluation methods exist~\cite{li2023martfl,sun2022profit,sun2024socially}, these methods require training feedback and can only be applied during the FL training phase. \projecttitle fundamentally differs from these methods but can be integrated with them since they target different phases. Therefore, we included an in-training client selection method for reference. A brief description of each baseline method is provided in Appendix~\ref{baseline}.

\begin{figure}[t]
\centering
    \subfloat[5 clients selected]{
    \includegraphics[width=.15\textwidth]{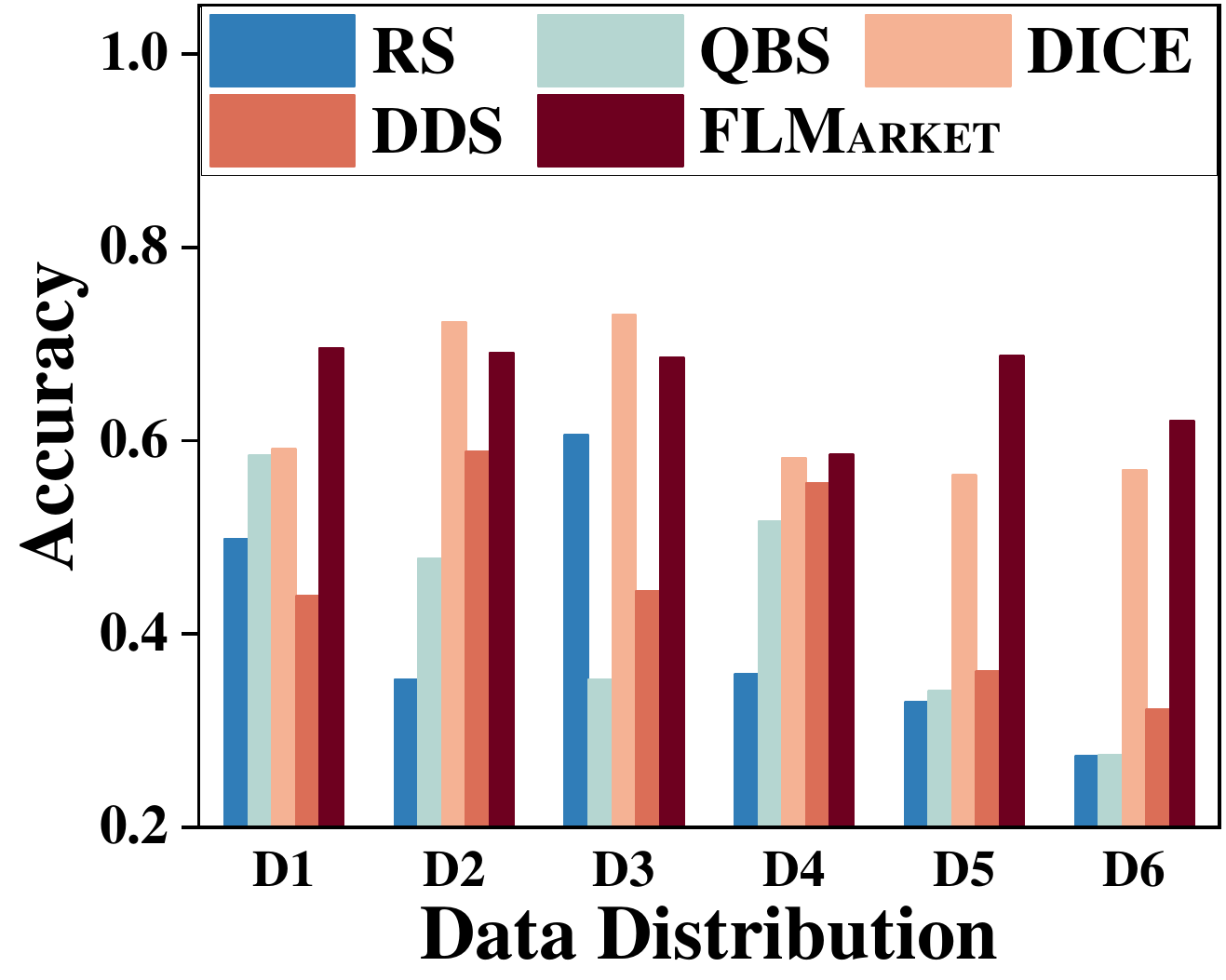}
        \label{fig:cf20x5}}
    \subfloat[10 clients selected]{
    \includegraphics[width=.15\textwidth]{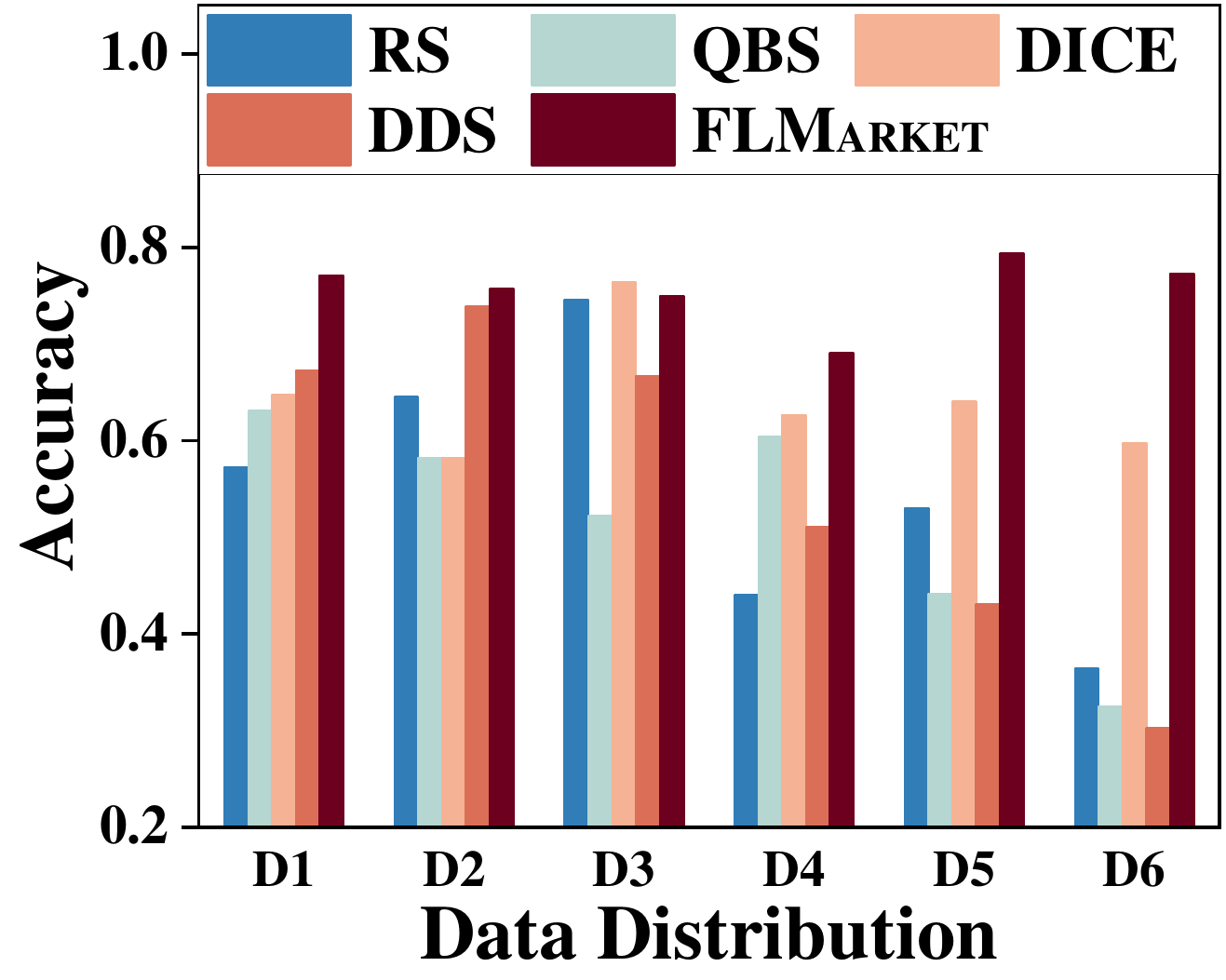}
        \label{fig:cf20x10}}
    \subfloat[15 clients selected]{
    \includegraphics[width=.15\textwidth]{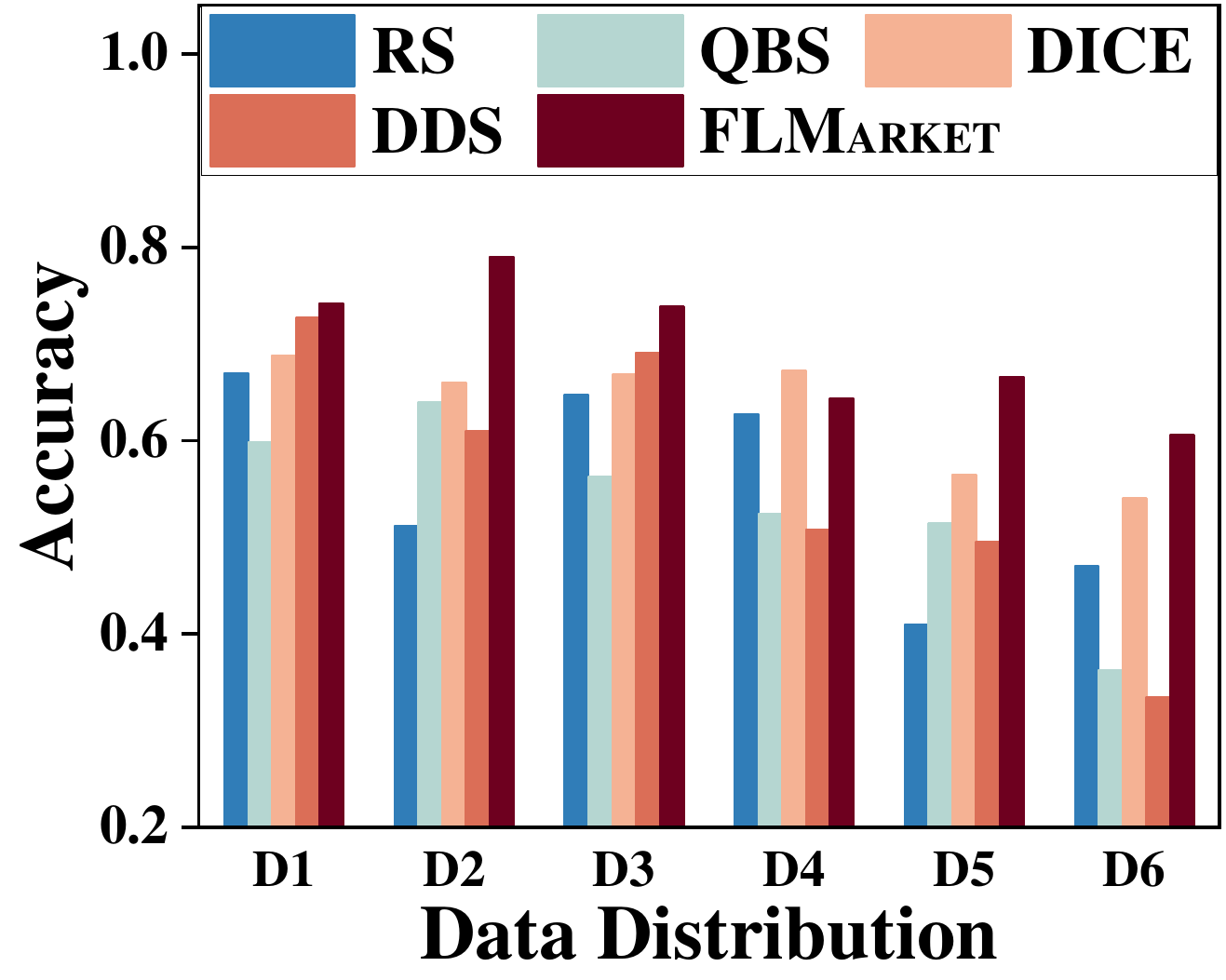}
        \label{fig:cf20x15}}
    \vspace{-3mm}
    \caption{CIFAR-10: $n$ clients selected from 20 clients}
    \label{fig:cifar}
    \vspace{-6mm}
\end{figure}

\begin{figure}[t]
\centering
    \subfloat[5 clients selected]{
      \includegraphics[width=.15\textwidth]{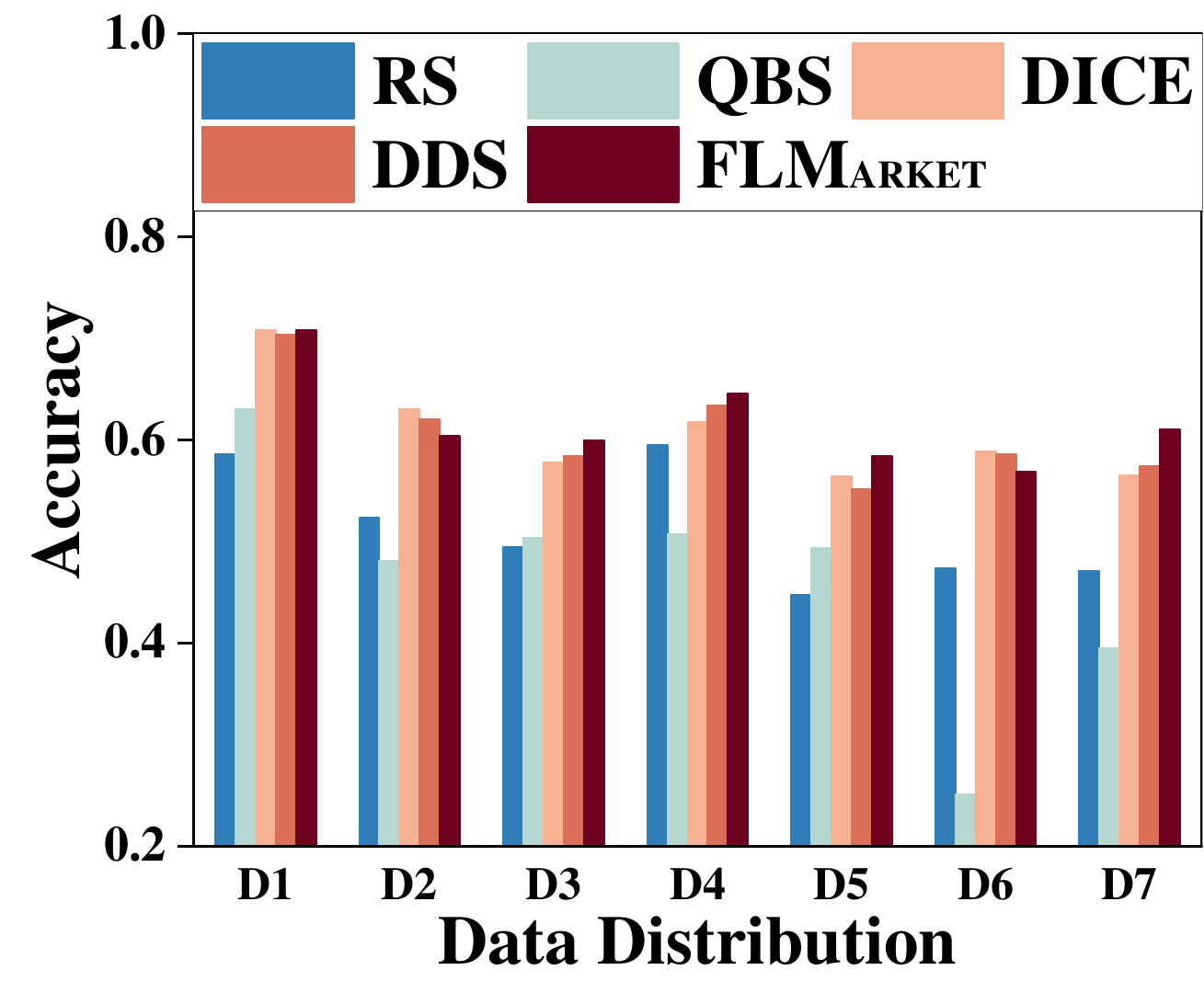}
    \label{fig:cn20x5}}
    \subfloat[10 clients selected]{
    \includegraphics[width=.15\textwidth]{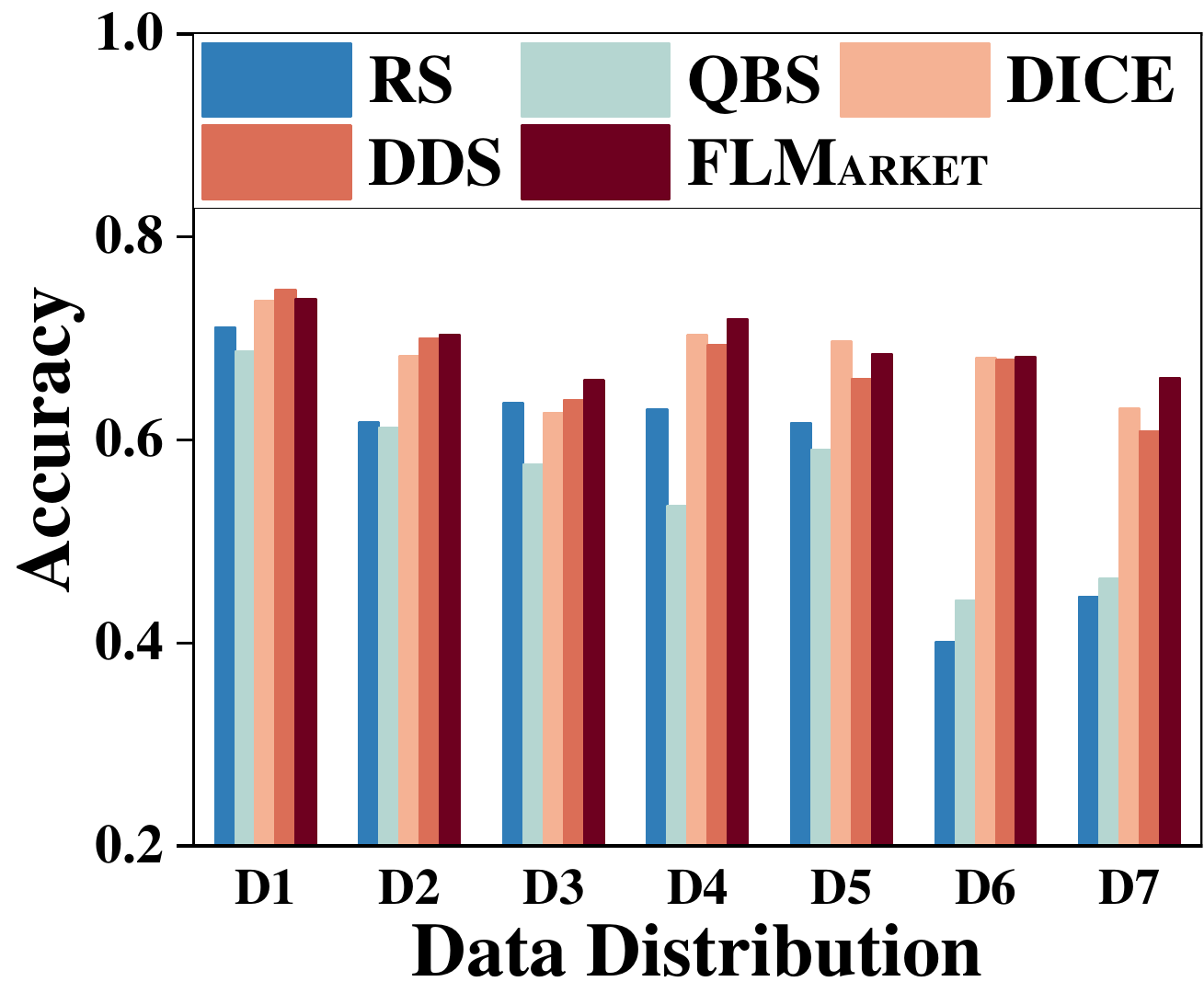}
    \label{fig:cn20x10}}
    \subfloat[15 clients selected]{
    \includegraphics[width=.15\textwidth]{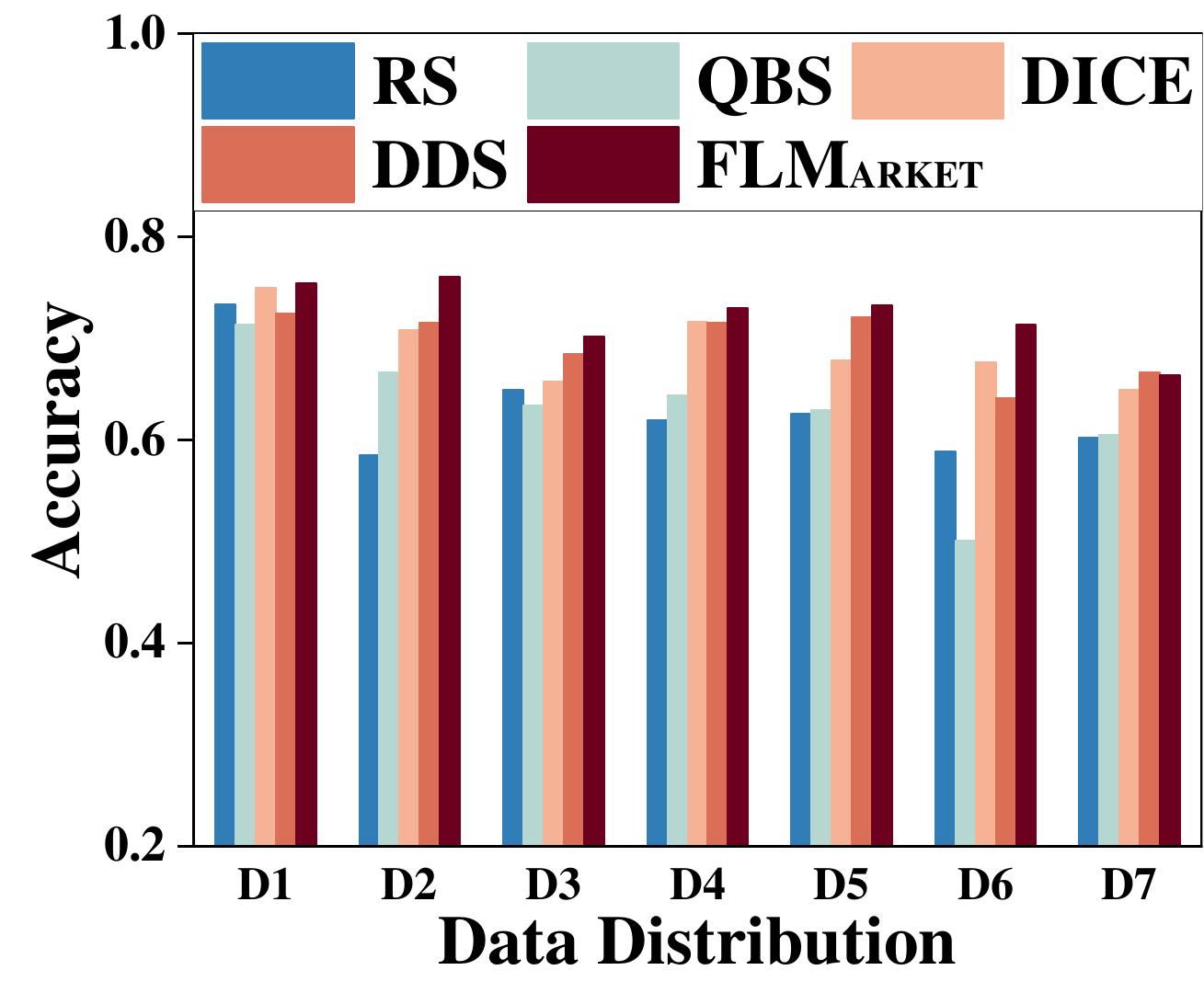}
    \label{fig:cn20x15}}
    \vspace{-3mm}
    \caption{CINIC-10: $n$ clients selected from 20 clients}
    \label{fig:cinic}
    \vspace{-6mm}
\end{figure} 

\begin{figure}[t]
\centering
    \subfloat[5 clients selected]{
    \includegraphics[width=.15\textwidth]{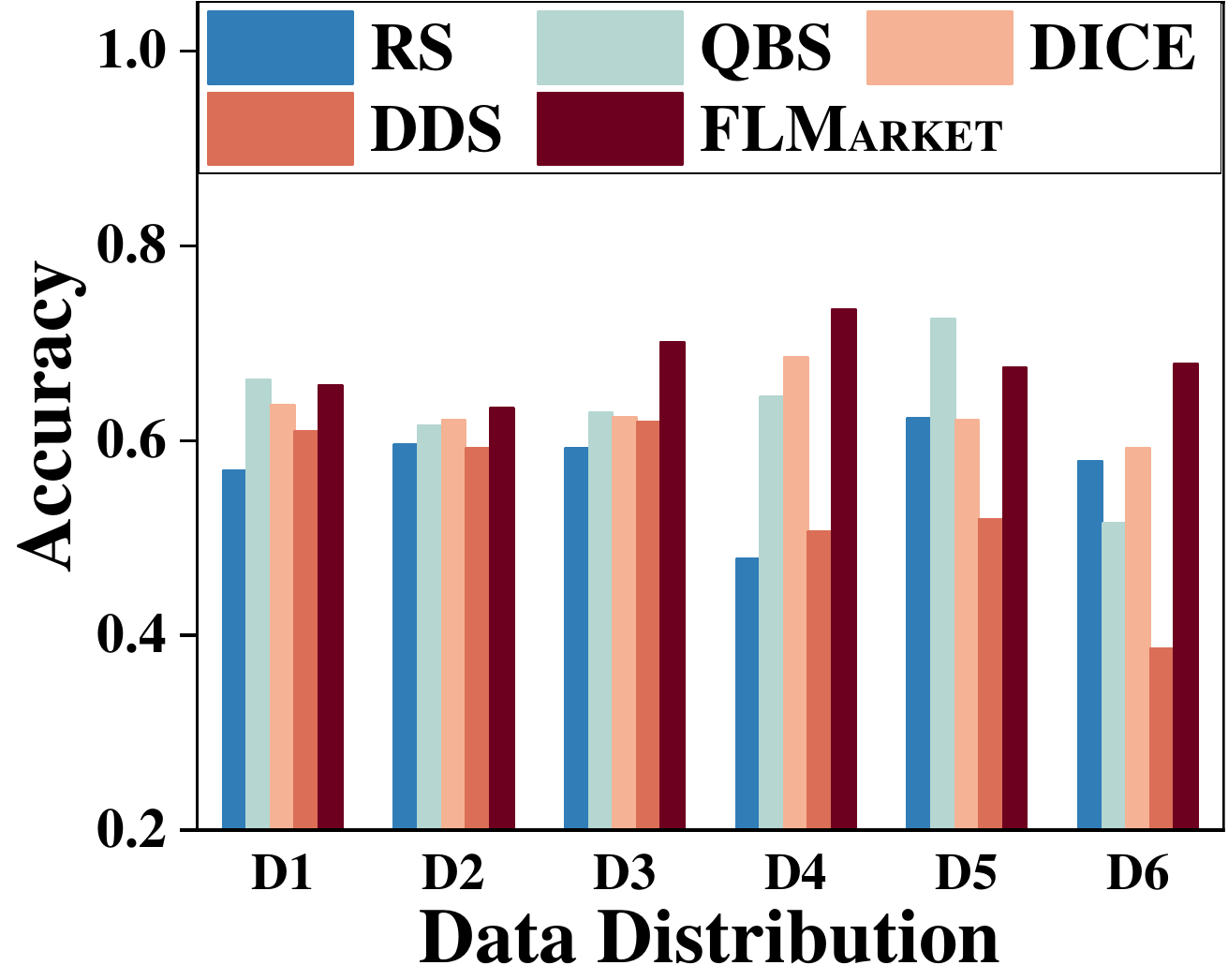}
    \label{fig:dsa20x5}}
    \subfloat[10 clients selected]{
    \includegraphics[width=.15\textwidth]{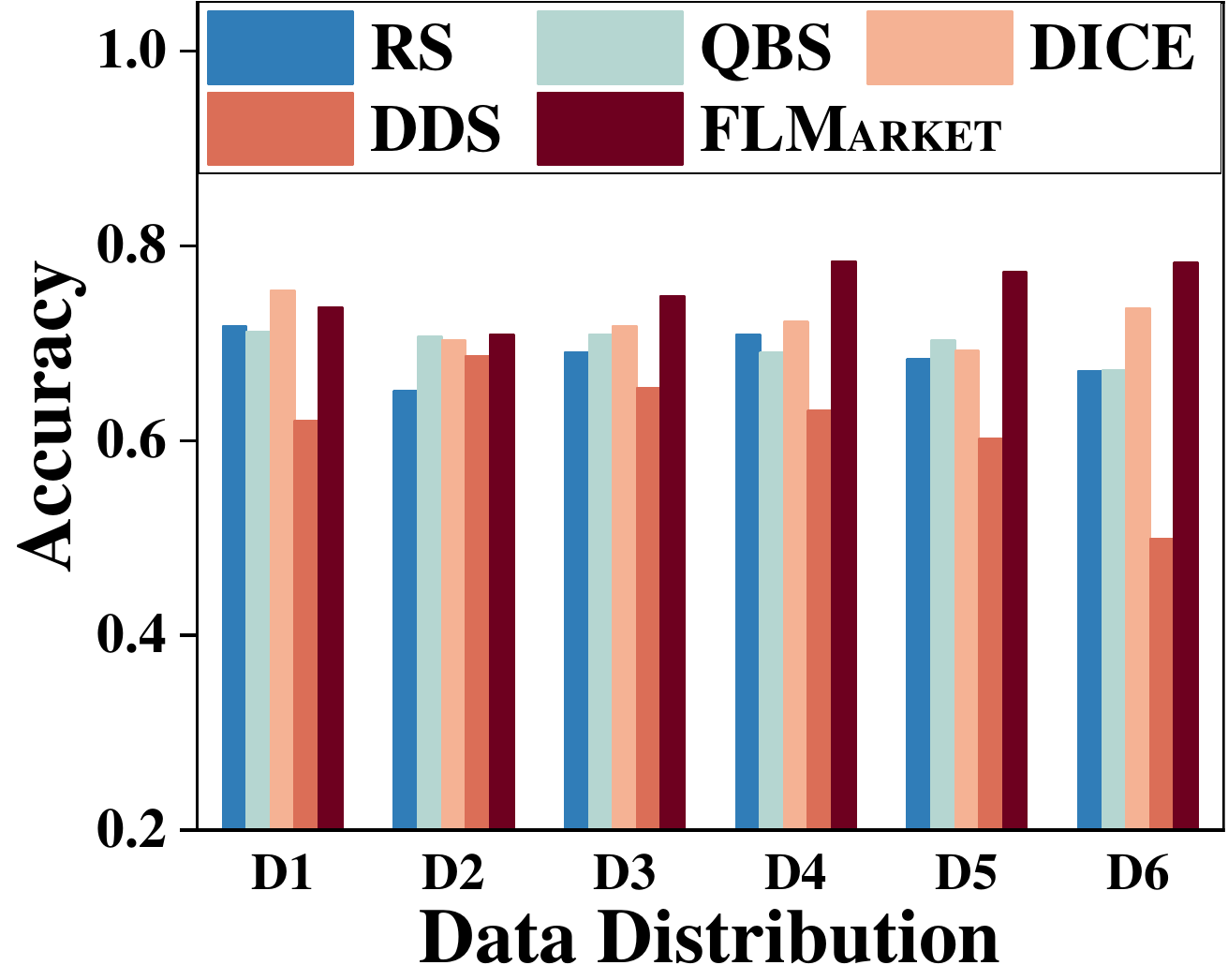}
    \label{fig:dsa20x10}}
    \subfloat[15 clients selected]{
    \includegraphics[width=.15\textwidth]{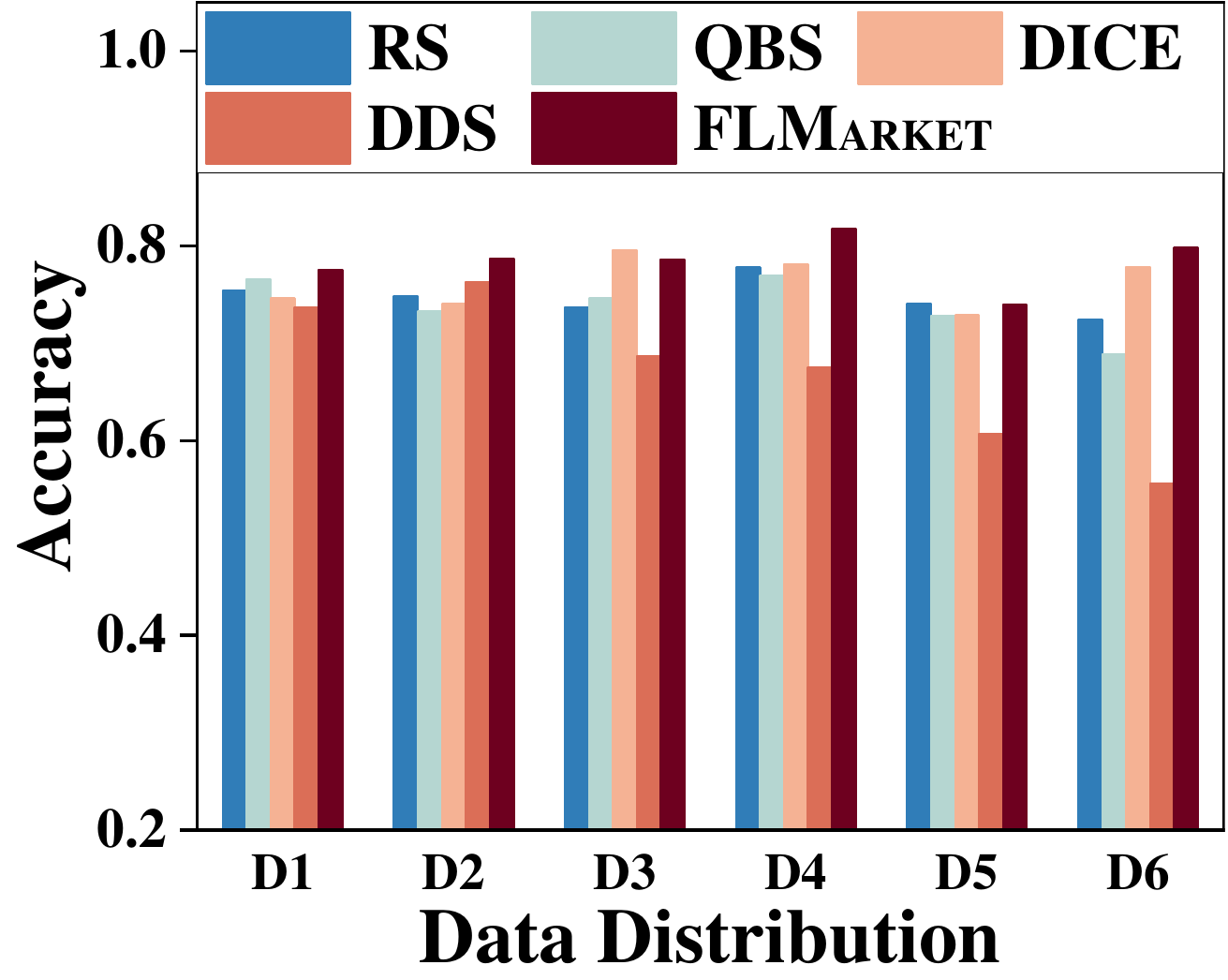}
    \label{fig:dsa20x15}}
    \vspace{-3mm}
    \caption{DEAP: $n$ clients selected from 20 clients}
    \label{fig:deap}
    \vspace{-4mm}
\end{figure}

\subsection{Comparison with Pre-training Selection}\label{sec:CSE}
We evaluate the performance of \projecttitle on selecting clients by comparing the proposed algorithm with four pre-training client selection baselines. Figures~\ref{fig:cifar},~\ref{fig:cinic} and ~\ref{fig:deap} show the final test accuracy of each baseline and \projecttitle.
It is observed that \projecttitle outperforms other baselines (RS, QBS, DICE, DDS) in a wide range of data distributions across three datasets, particularly when the distribution is highly unbalanced (e.g., D4, D5, D6).

For CIFAR-10, \projecttitle achieves up to 40.78 $\%$,  44.78 $\%$,  17.52 $\%$ and 47.05 $\%$ (all on D6 in 10 out of 20 selection) higher accuracy when compared to RS, QBS, DICE and DDS, respectively. In terms of CINIC-10, compared to four baselines, \projecttitle has up to 28.10 $\%$ (on D6 in 10 out of 20 selection),  31.74 $\%$ (on D6 in 5 out of 20 selection),  5.44 $\%$ (on D5 in 15 out of 20 selection) and  7.27 $\%$ (on D6 in 15 out of 20 selection) higher accuracy, respectively. Regarding DEAP, Figure~\ref{fig:deap} shows that \projecttitle achieves up to 25.67 $\%$ (on D4 in 5 out of 20 selection),  16.41 $\%$ (on D6 in 5 out of 20 selection),  8.63 $\%$ (on D6 in 5 out of 20 selection) and  29.21 $\%$ (on D6 in 5 out of 20 selection) higher accuracy than other baselines. As the data distribution becomes increasingly unbalanced (e.g., D4, D5 and D6), \projecttitle exhibits significantly better accuracy than other baselines. This demonstrates \projecttitle's ability to mitigate the impact of global imbalance. 

Among all the baselines, DICE exhibits the best performance. Nonetheless, \projecttitle consistently outperforms DICE across three datasets and three different selection scenarios, with average accuracy improvements of 7.08 $\%$, 1.81 $\%$ and 3.53 $\%$ on CIFAR-10, CINIC-10 and DEAP respectively.
Across all 57 tested distributions, \projecttitle achieves the highest accuracy in 44 distributions (77\%). For the most imbalanced distributions (D4, D5, and D6), \projecttitle achieves the highest accuracy in 22 out of 27 test distributions (81\%), highlighting its strength in handling imbalanced distributions.
When the number of clients is scaled up to 100, \projecttitle can still achieve better performance compared to the baselines (the experimental results are detailed in Appendix~\ref{sec:large_clients}).

\begin{figure}[t]
\centering
    \subfloat[Accuracy convergence]{
    \includegraphics[width=0.23\textwidth]{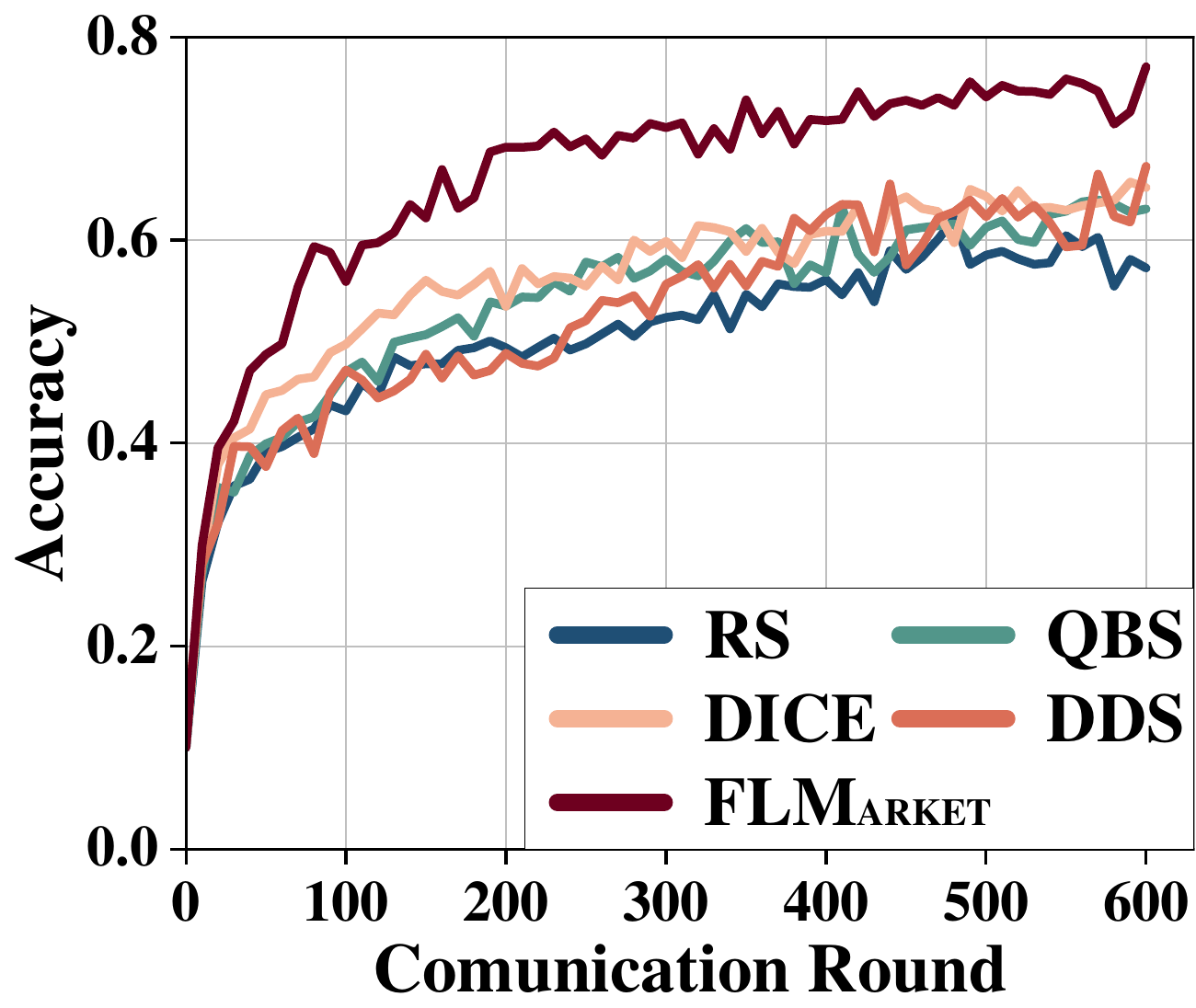}
    \label{fig:my_acc}
    }
    \subfloat[Loss convergence]{
     \includegraphics[width=0.23\textwidth]{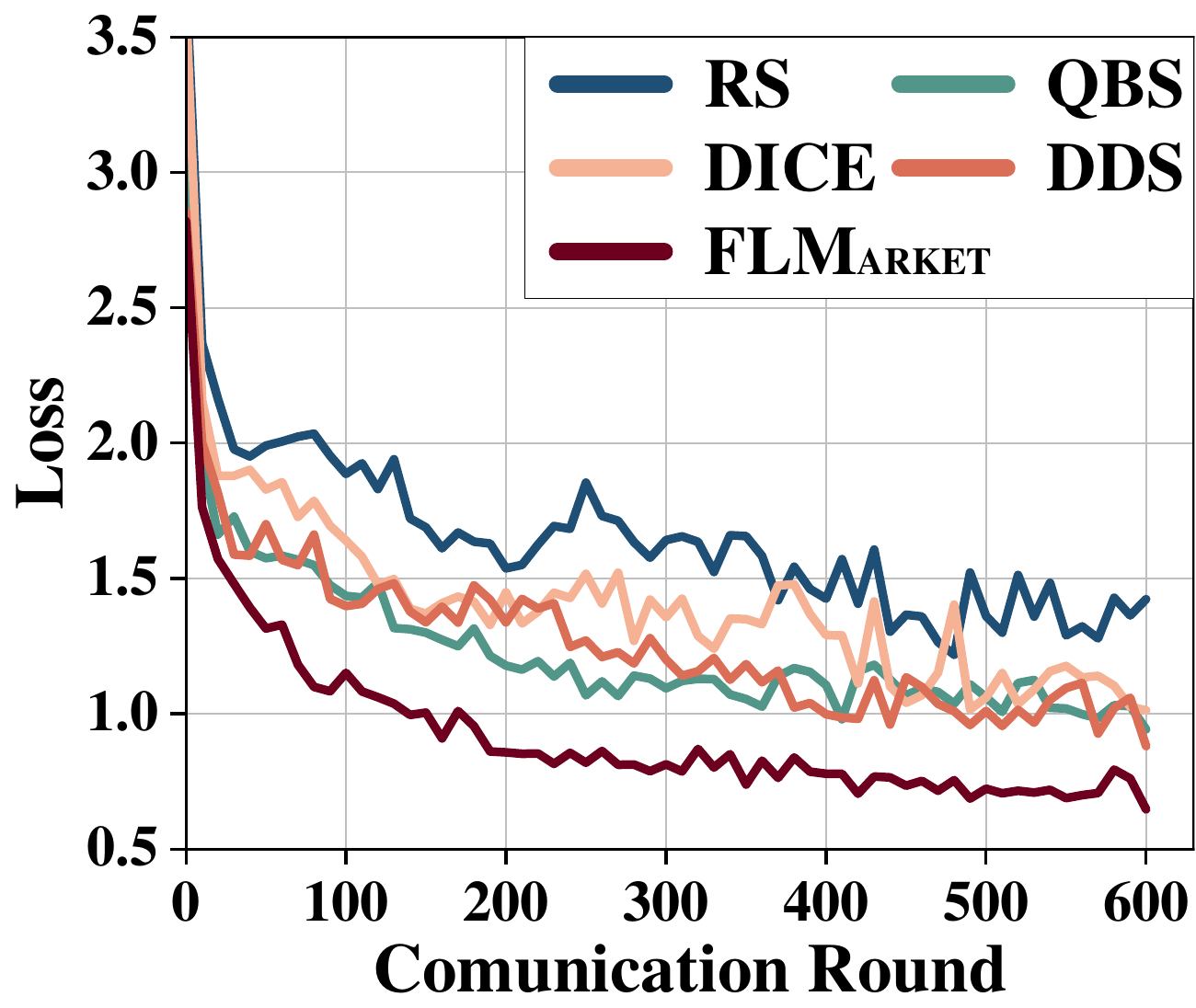}
    \label{fig:my_loss}
    }
    \caption{The accuracy and loss curves for \projecttitle and other four baselines under the D1 distribution with a selection of 10 out of 20 clients on CIFAR-10 dataset.
    }
    \label{fig:my_convergence}
    \vspace{-3mm}
\end{figure}

\myparagraph{Convergence speed}\label{sec:convergence}
We compare the convergence rate of \projecttitle with all the baseline methods. Figure~\ref{fig:my_convergence} demonstrates that \projecttitle exhibits significantly faster convergence and attains a lower training loss, consistently outperforming other approaches. In detail, \projecttitle achieves an target accuracy of 0.6 after only approximately 130 rounds, while RS, QBS, DICE and DDS require 550, 350, 170 and 380 rounds, respectively. This results in a speedup of 4.23 $\times$, 2.69 $\times$, 1.31 $\times$, and 2.92 $\times$, respectively. We also observed similar conclusions in other datasets and distributions as well, where \projecttitle exhibits faster convergence compared to other baselines.
We also observed a similar speedup in latency for \projecttitle in achieving the target accuracy. This outcome is expected, as \projecttitle conducts pre-training data evaluation once before the training, thereby adding no computational overhead during the FL training process.

\myparagraph{The Data Distribution of Selected Clients}
We further analyze the results of selected clients by comparing their data distributions. Specifically, we compare \projecttitle to DICE under the selection of 10 out of 20 clients using the global distribution D6 of CIFAR-10. As shown in Table~\ref{tab:selectedclients}, both strategies select the same six clients from the client pool, namely, clients 2, 3, 7, 16, 17, and 19. 
However, in the disjoint selected clients, we observe that the clients chosen by \projecttitle (i.e., 4, 8, 10, and 12) include data samples for all classes. In contrast, the clients selected by DICE (i.e., 1, 5, 13, and 18) missing some classes, i.e., classes 8, 9, and 10.

Although DICE also takes into account the impact of both data quantity and class distributions, it suffers issues:
1) it ignores that the marginal utility of data volume is decreased; 
2) It only locally considers the class distribution instead of the global distribution.
These designs result in DICE favouring clients with larger data quantities and smaller local variances. In contrast, \projecttitle chooses clients with lower data quantities and global variance.

\begin{table}[t]
    \noindent
    \renewcommand{\arraystretch}{1.4}
    \caption{Clients selected by \projecttitle and DICE on distribution 6 of CIFAR-10 dataset}
    \vspace{-3mm}
    \resizebox{0.47\textwidth}{!}{  
    \label{tabel5}
        \begin{tabular}{c|c|cccccccccc|c}
        \toprule
        \multicolumn{2}{c|}{ \diagbox[innerwidth=2.7cm]{Client}{Class}}&C1&C2&	C3&	C4&	C5&	C6&	C7&	C8&	C9&	C10&Distribution\\
        \specialrule{0.05em}{0.1pt}{3pt}
        \multirow{6}{*}{Overlapping}&Client 2&	9&203&19&224&224&318&362&40&0&0&\begin{minipage}{0.1\textwidth}\includegraphics[width=20mm,height=5mm]{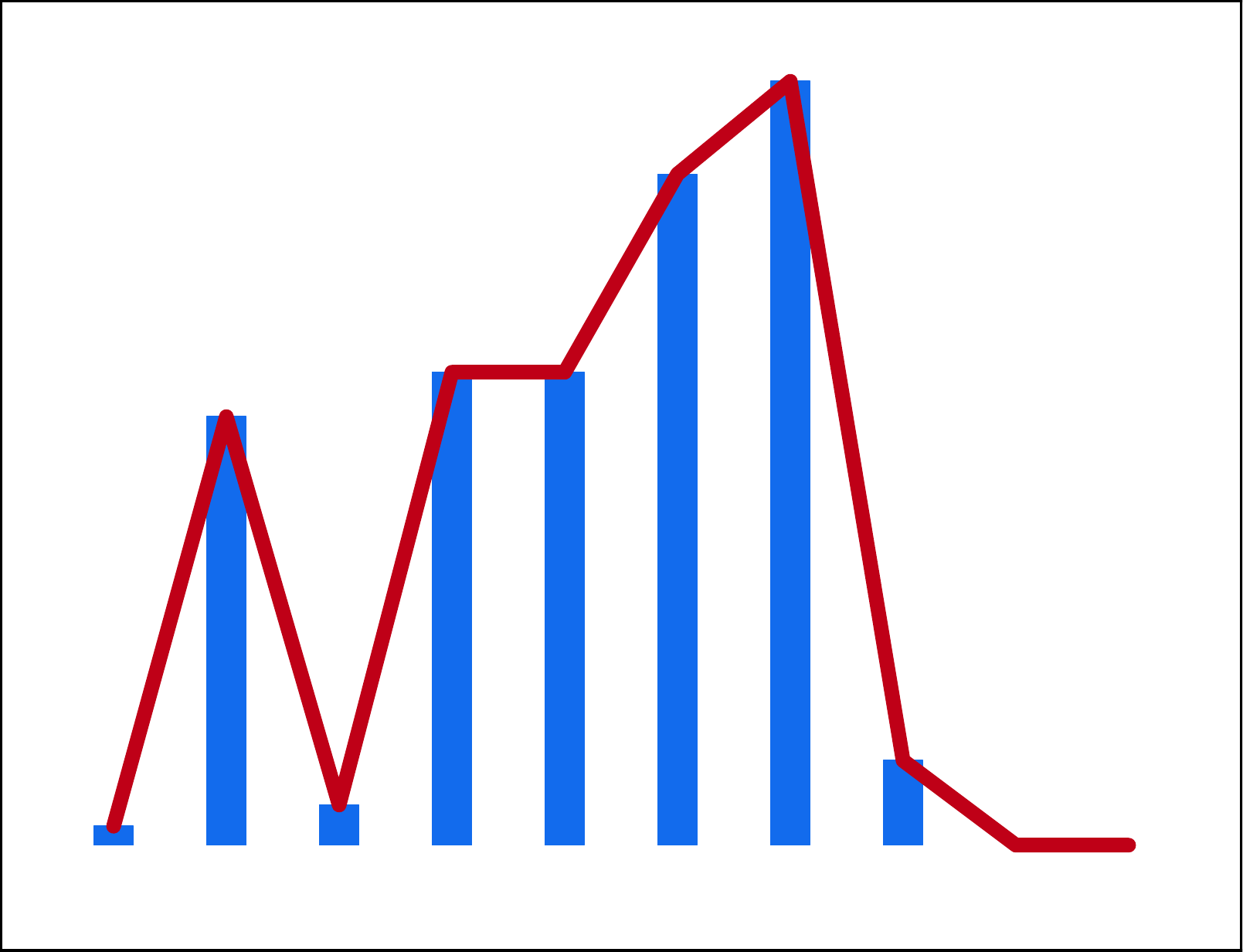}\end{minipage}\\
        &Client 3&	5&120&112&311&472&58&0&319&360&0&\begin{minipage}{0.1\textwidth}\includegraphics[width=20mm,height=5mm]{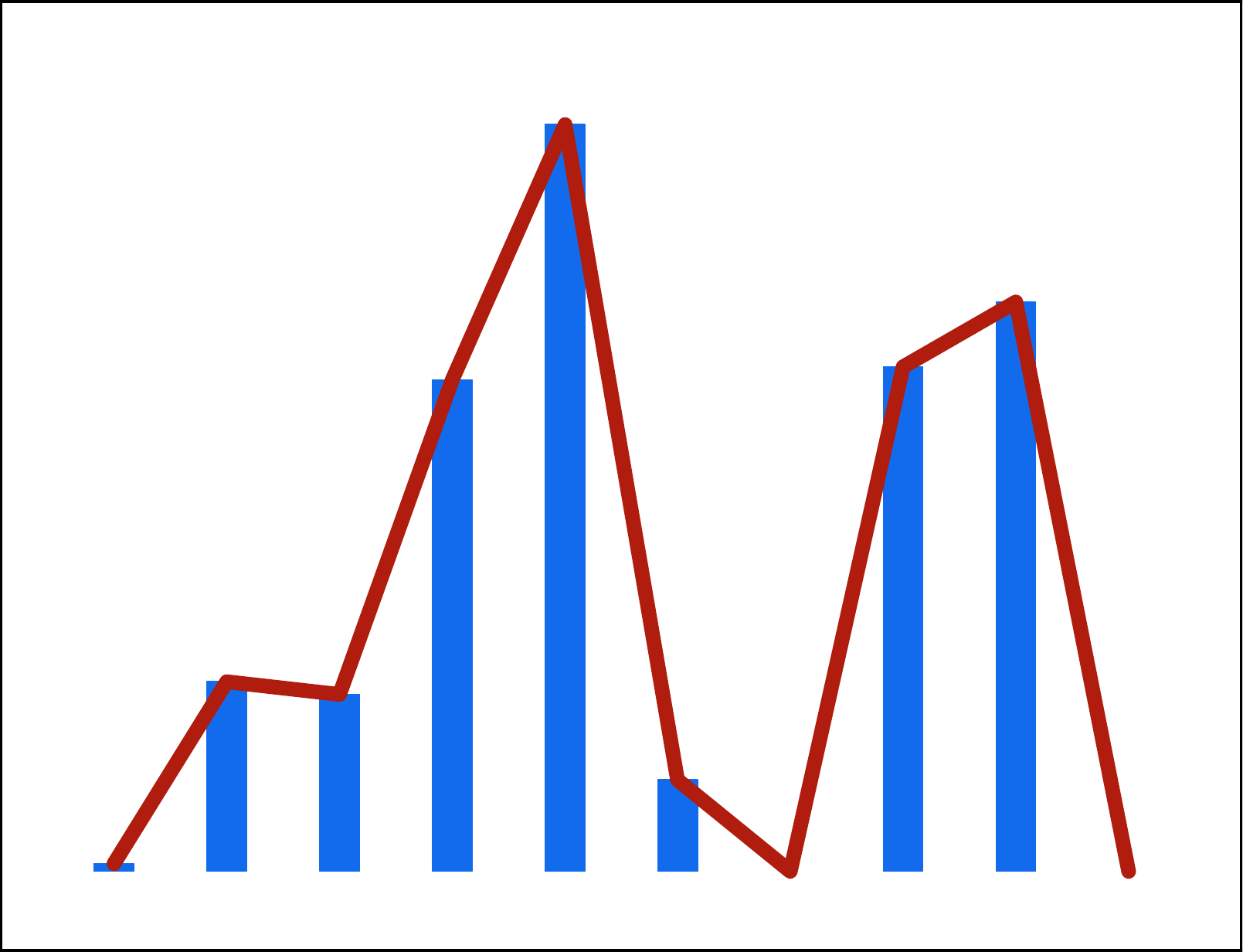}\end{minipage}\\
        &Client 7&	6&125&66&14&207&75&6&97&121&36&\begin{minipage}{0.1\textwidth}\includegraphics[width=20mm,height=5mm]{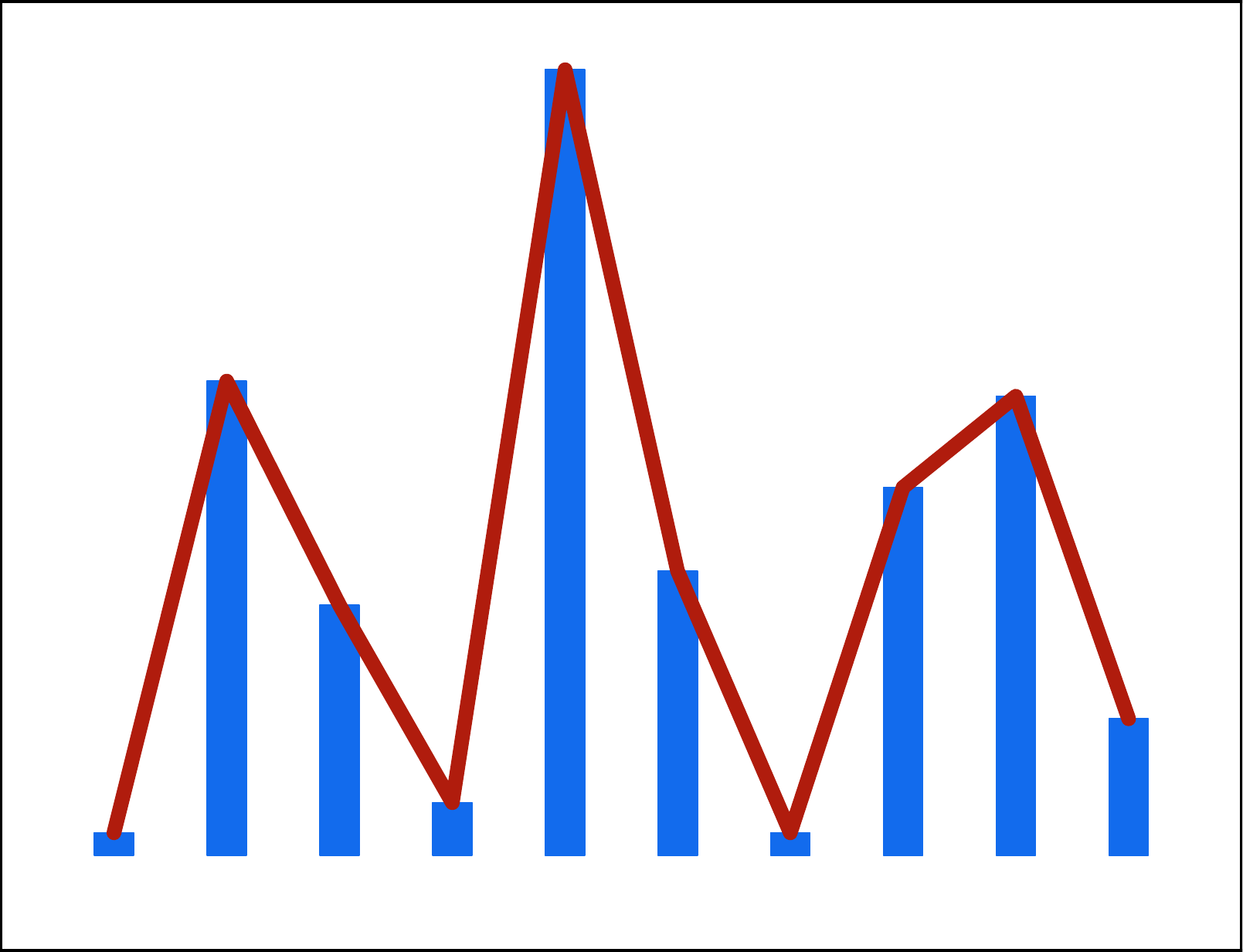}\end{minipage}\\
        &Client 16&	151&378&29&4&85&44&125&12&69&174&\begin{minipage}{0.1\textwidth}\includegraphics[width=20mm,height=5mm]{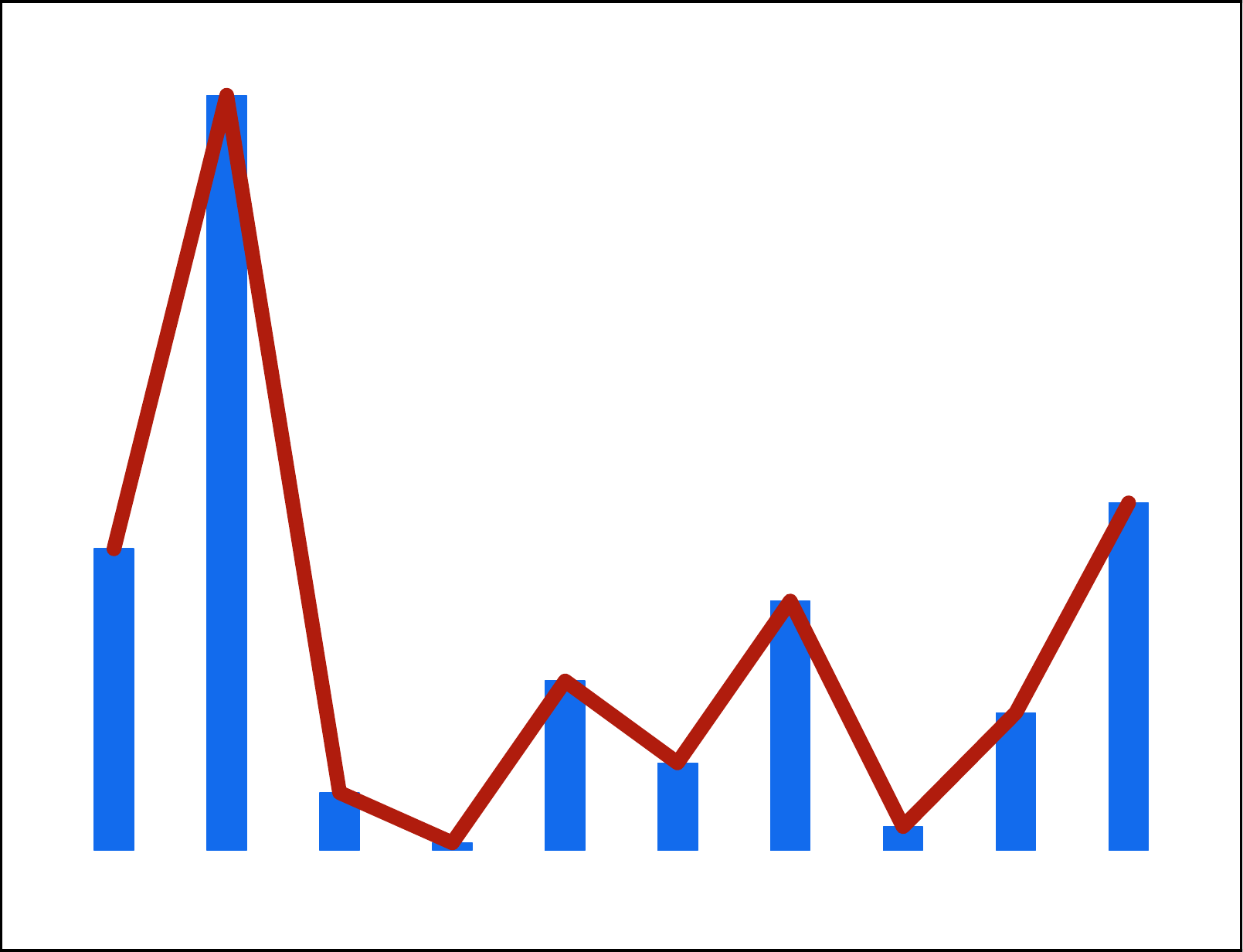}\end{minipage}\\
        &Client 17&	30&42&171&201&11&1&322&111&53&70&\begin{minipage}{0.1\textwidth}\includegraphics[width=20mm,height=5mm]{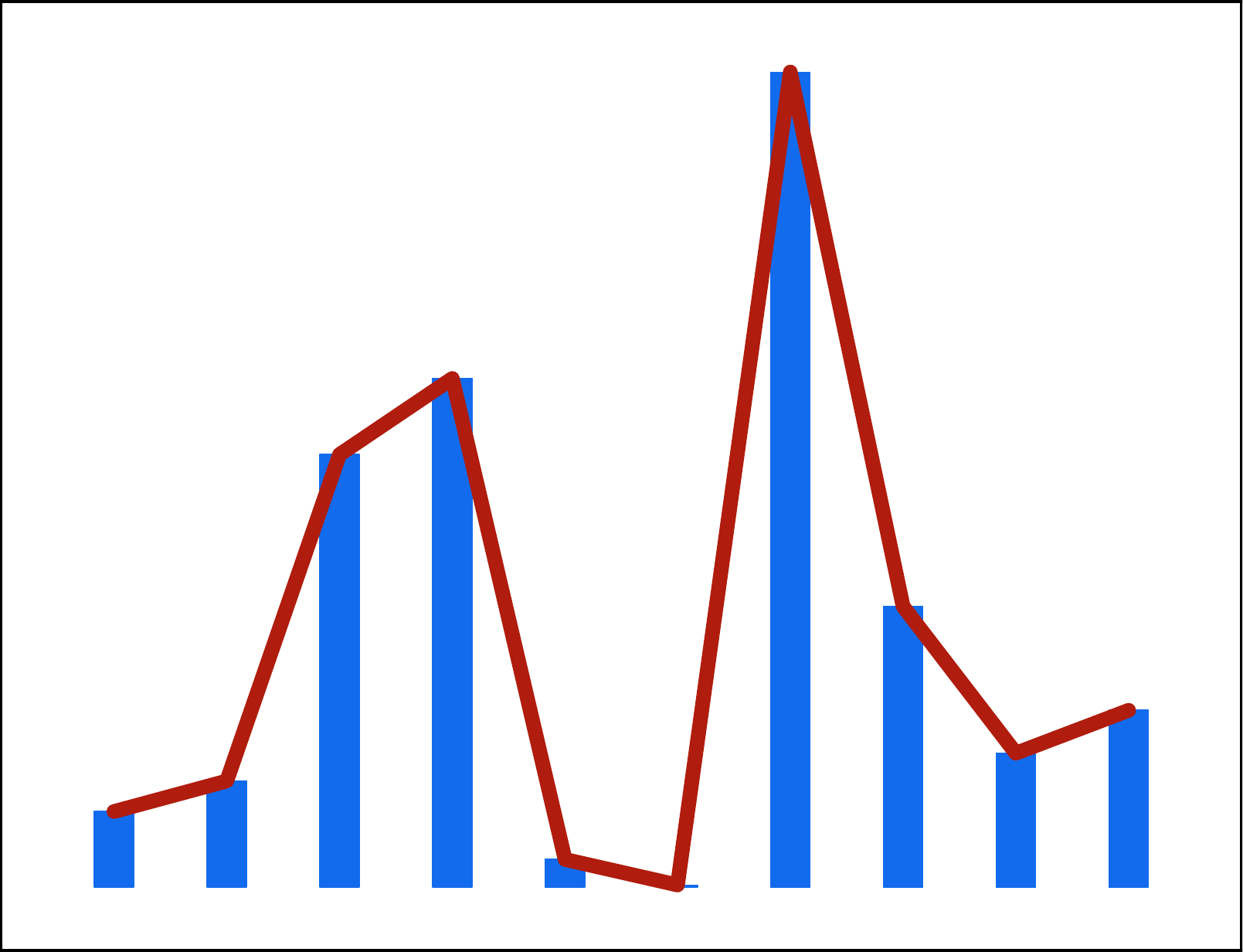}\end{minipage}\\
        &Client 19&	33&212&23&99&29&64&141&109&157&25&\begin{minipage}{0.1\textwidth}\includegraphics[width=20mm,height=5mm]{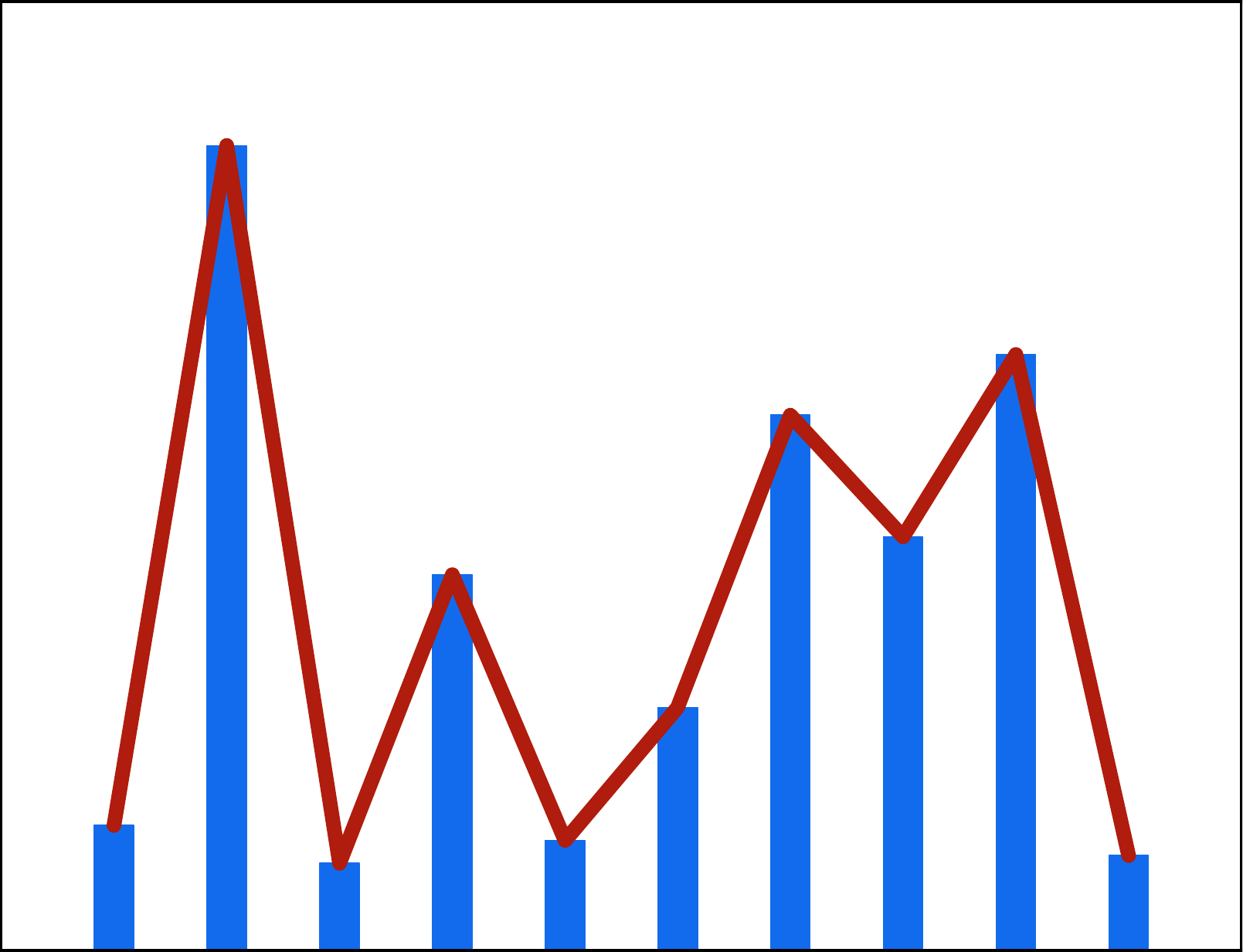}\end{minipage}\\
        \specialrule{0.05em}{3pt}{3pt}
        \multirow{5}{*}{\projecttitle-specific}&Client 4&	220&11&2&46&659&0&55&16&169&143&\begin{minipage}{0.1\textwidth}\includegraphics[width=20mm,height=5mm]{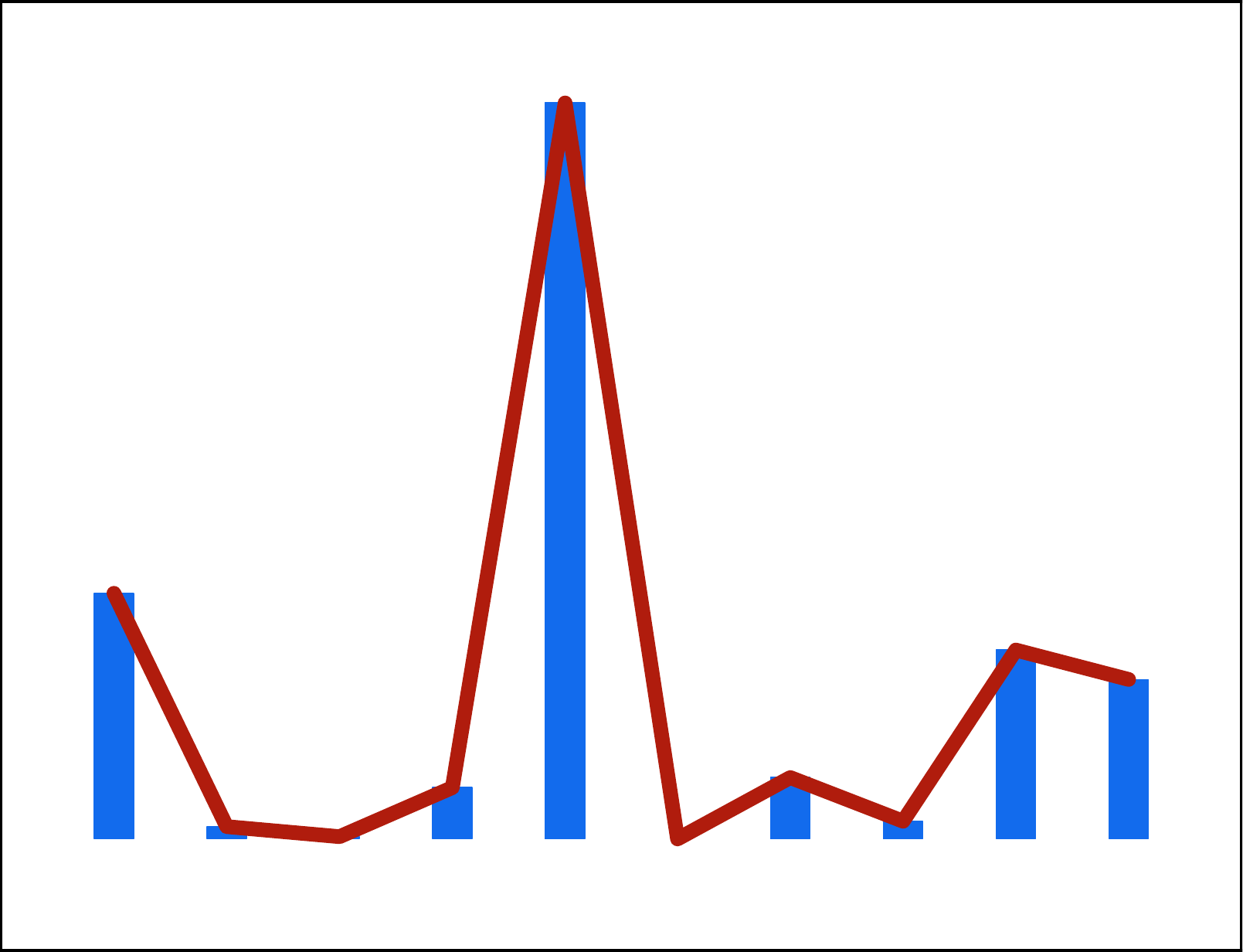}\end{minipage}\\
         &Client 8&	401&11&56&6&724&66&36&5&25&6&\begin{minipage}{0.1\textwidth}\includegraphics[width=20mm,height=5mm]{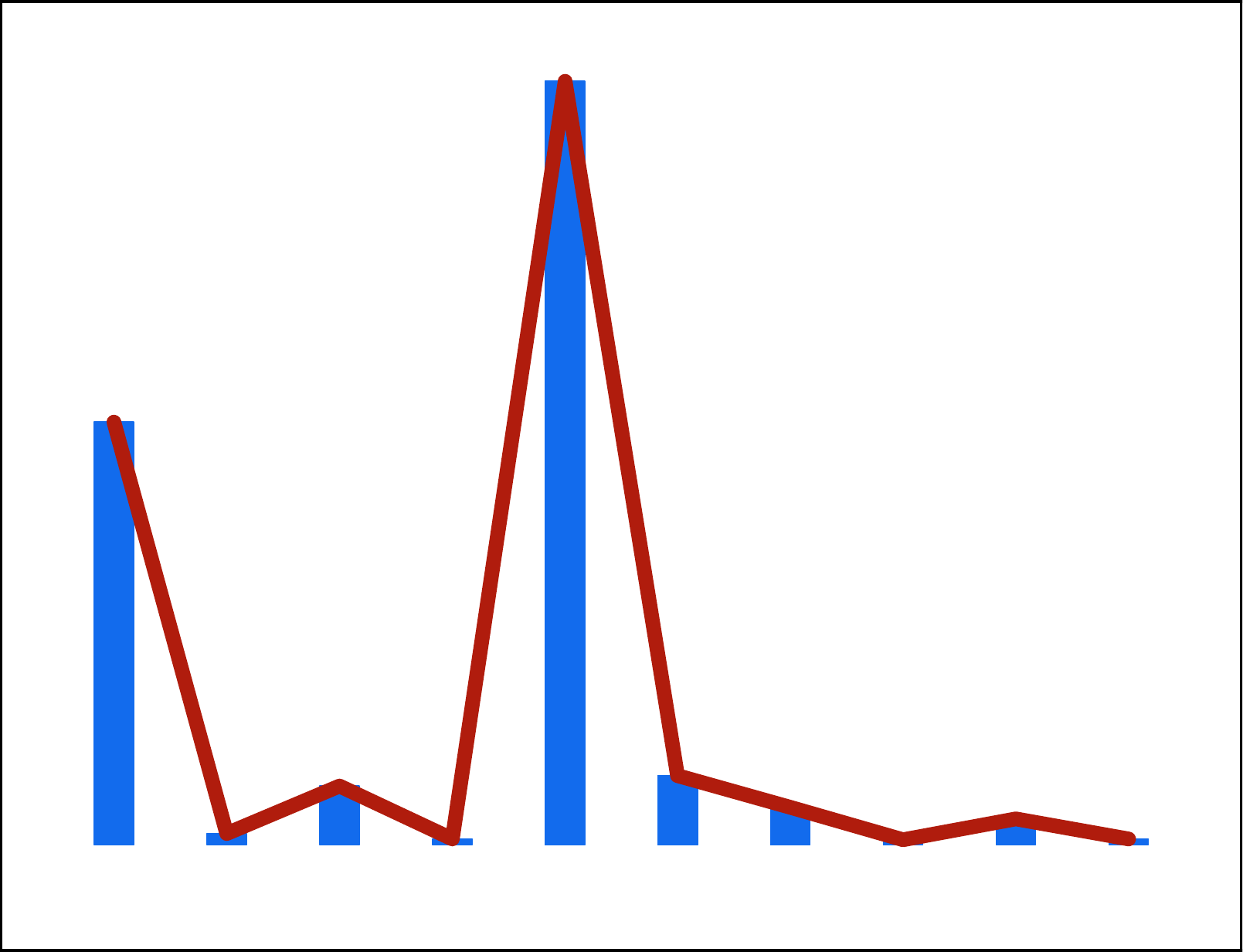}\end{minipage}\\
         &Client 10&	1&244&13&93&4&83&52&191&1&25&\begin{minipage}{0.1\textwidth}\includegraphics[width=20mm,height=5mm]{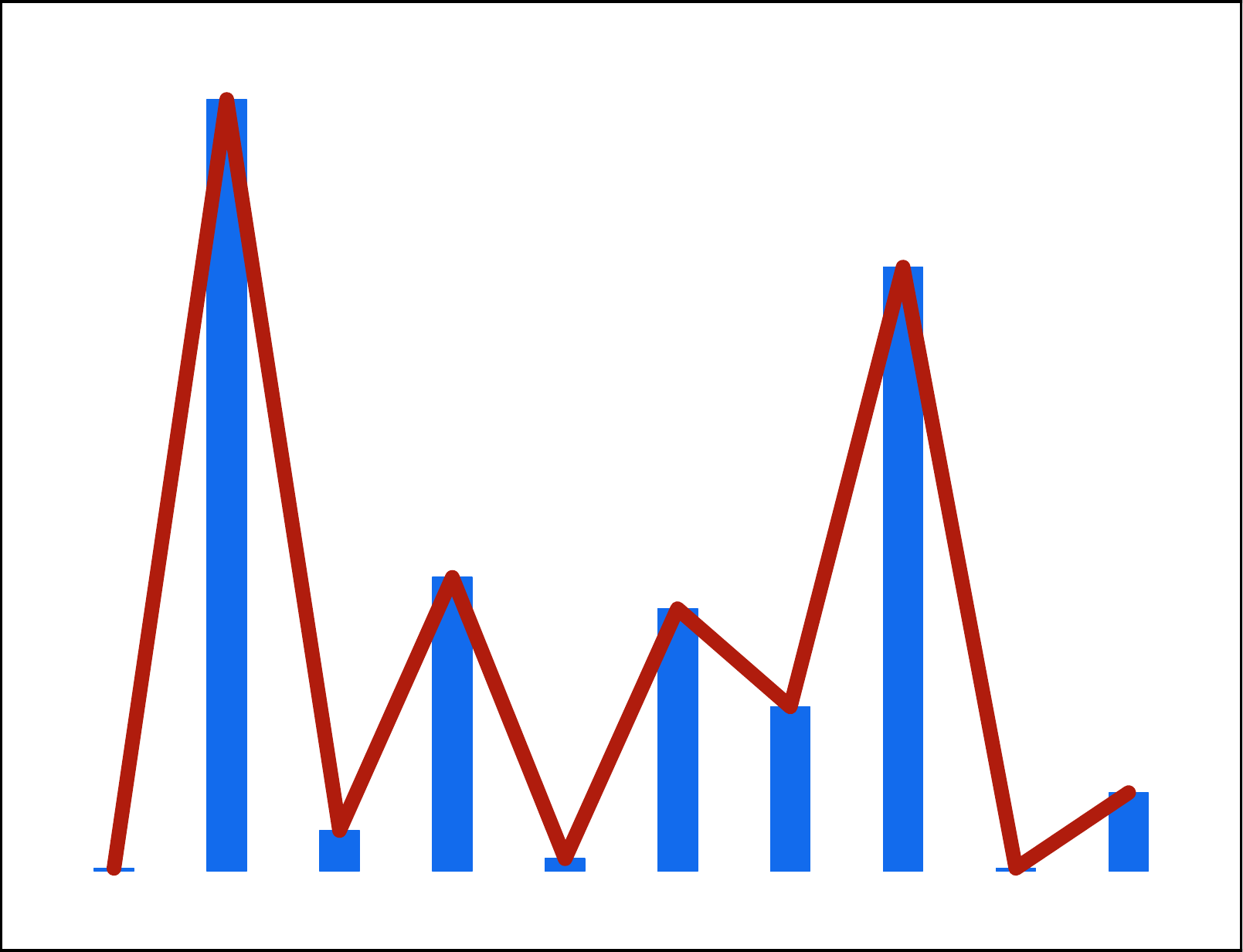}\end{minipage}\\
         &Client 12&	287&9&42&65&13&5&114&600&45&21&\begin{minipage}{0.1\textwidth}\includegraphics[width=20mm,height=5mm]{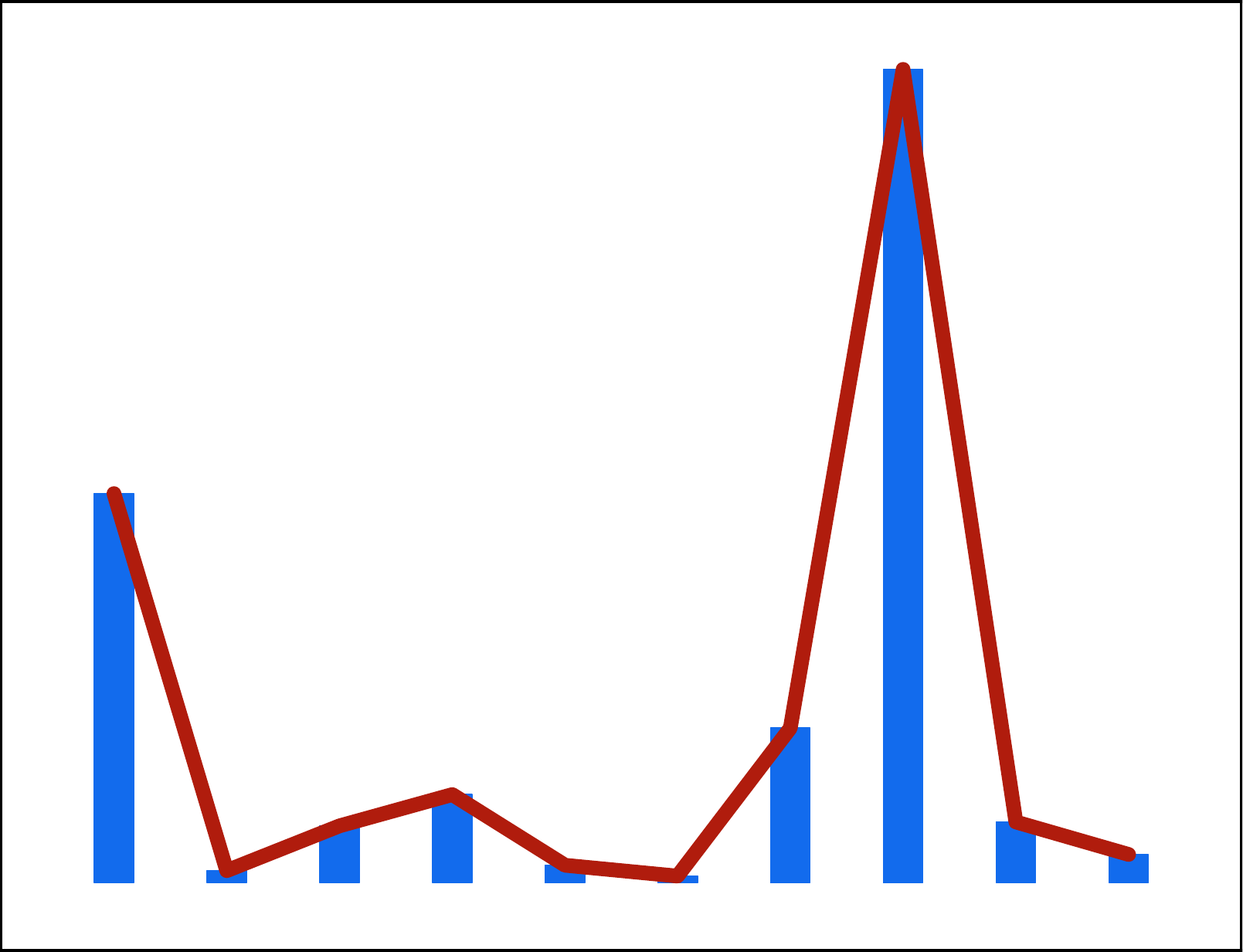}\end{minipage}\\
         \cmidrule{2-13}
         &Sum& 1143&1355&533&1063&2428&714&1413&1500&1000&500&\begin{minipage}{0.1\textwidth}\includegraphics[width=20mm,height=5mm]{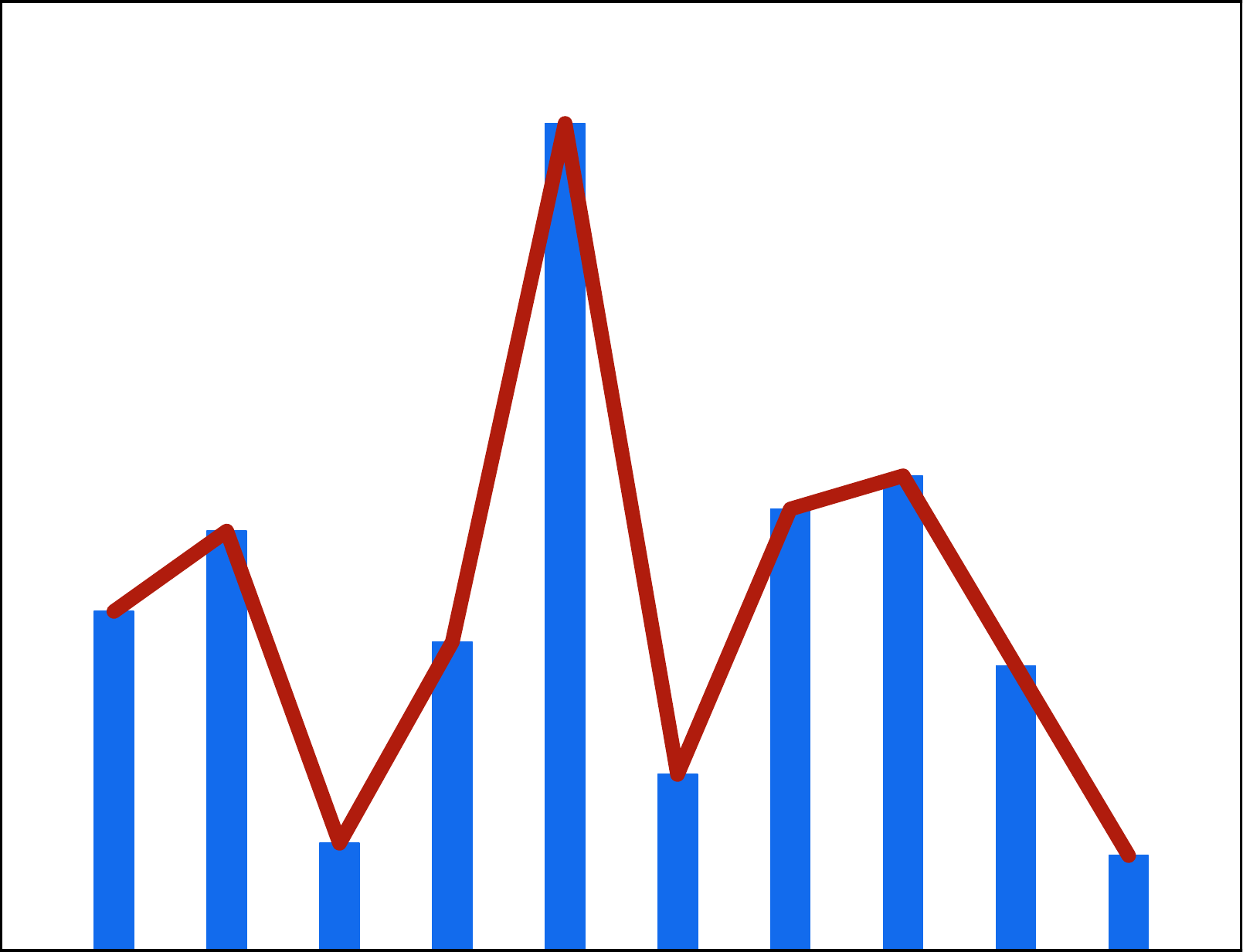}\end{minipage}\\
        \specialrule{0.05em}{3pt}{3pt}
        \multirow{5}{*}{DICE-specific}&Client 1&	530&38&59&349&33&346&209&0&0&0&\begin{minipage}{0.1\textwidth}
        \includegraphics[width=20mm, height=5mm]{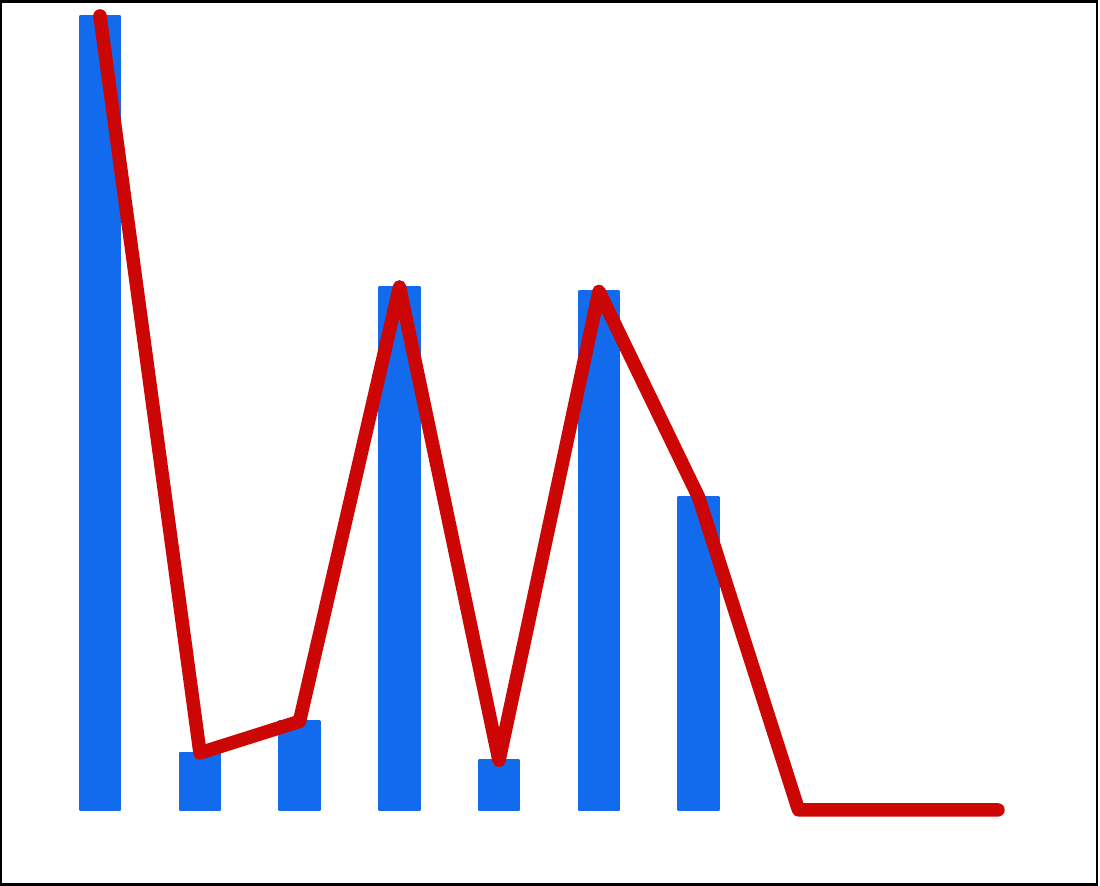}\end{minipage}\\
         &Client 5&	281&174&211&4&247&570&0&0&0&0&\begin{minipage}{0.1\textwidth}
        \includegraphics[width=20mm, height=5mm]{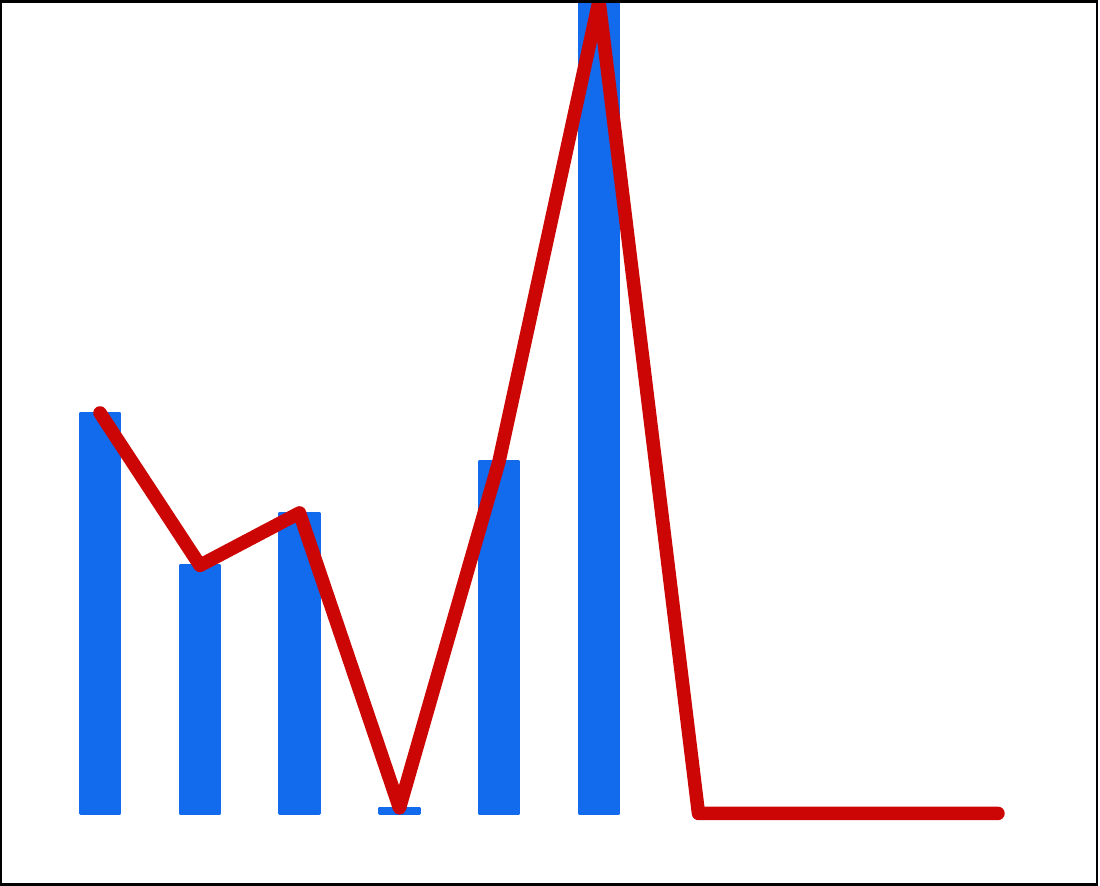}\end{minipage}\\
         &Client 13&	257&159&37&13&125&628&180&0&0&0&\begin{minipage}{0.1\textwidth}
        \includegraphics[width=20mm, height=5mm]{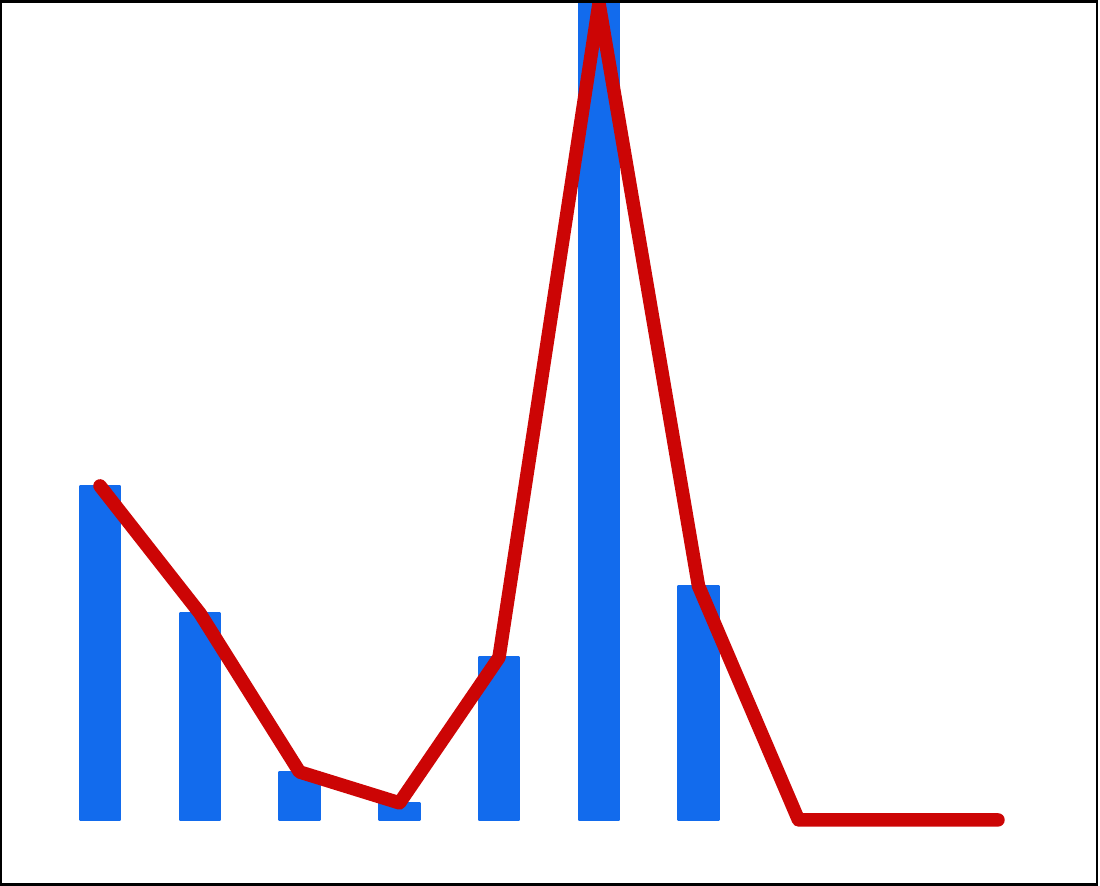}\end{minipage}\\
         &Client 18&	2&605&43&133&166&271&398&0&0&0&\begin{minipage}{0.1\textwidth}
        \includegraphics[width=20mm, height=5mm]{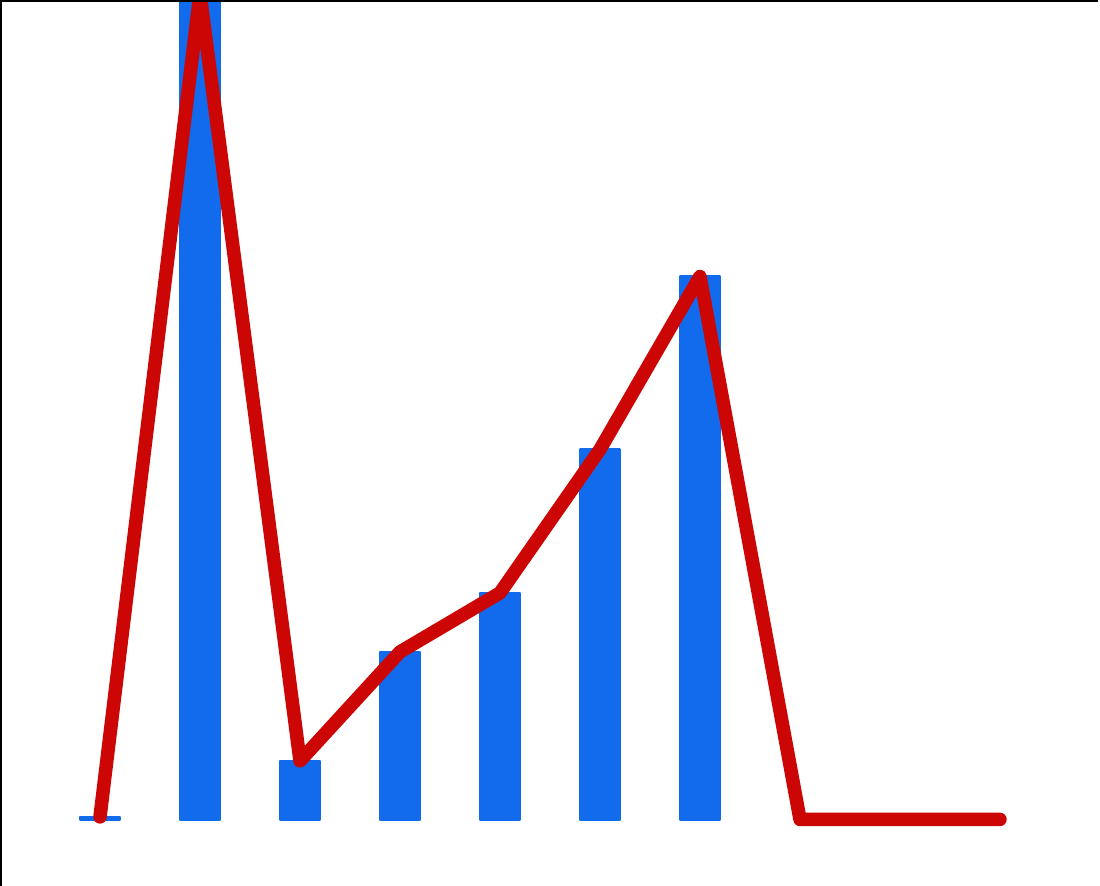}\end{minipage}\\
        \cmidrule{2-13}
         &Sum& 1304&2056&770&1342&1599&2345&1743&688&760&305&\begin{minipage}{0.1\textwidth}
        \includegraphics[width=20mm, height=5mm]{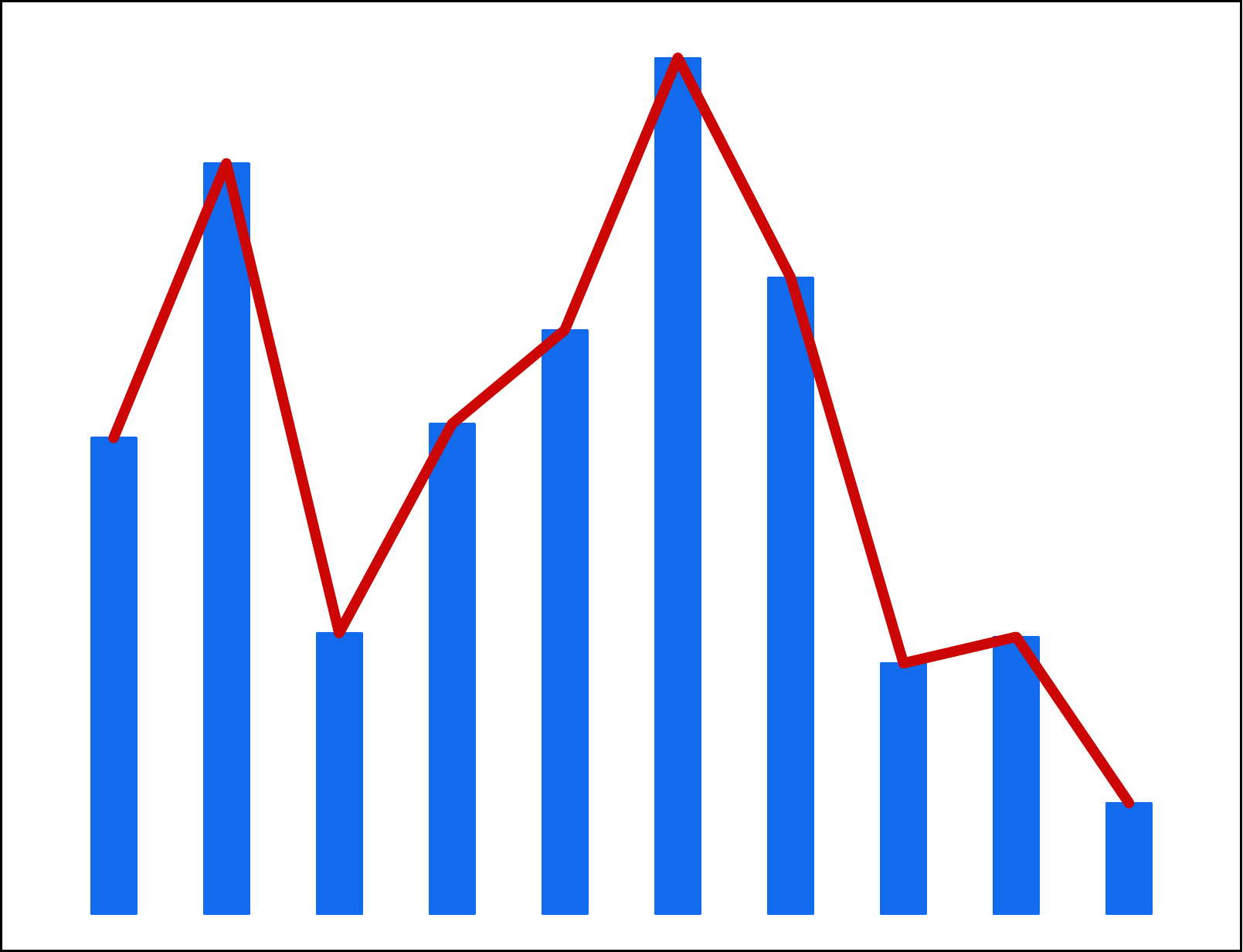}\end{minipage}\\
        \specialrule{0.05em}{3pt}{3pt}
        \multicolumn{1}{c|}{ \projecttitle}&\multicolumn{2}{c|}{\multirow{2}{*}{Data Number} }& \multicolumn{2}{c|}{ 11449 } &\multicolumn{2}{c|}{ \multirow{2}{*}{Variance}}&\multicolumn{2}{c}{ 313880.1}&\multicolumn{2}{|c|}{ \multirow{2}{*}{Accuracy}}&\multicolumn{2}{c}{77.28 $\%$ }\\
        \cmidrule{1-1}\cmidrule{4-5}\cmidrule{8-9}\cmidrule{12-13}
        \multicolumn{1}{c|}{ DICE}&\multicolumn{2}{c|}{}& \multicolumn{2}{c|}{ 12912 } &\multicolumn{2}{c|}{}&\multicolumn{2}{c}{ 431920.6}&\multicolumn{2}{|c|}{}&\multicolumn{2}{c}{59.76 $\%$ }\\
    \bottomrule
    \end{tabular}
    }
    \label{tab:selectedclients}
    \vspace{-2mm}
    \end{table}
    
\subsection{Comparison with In-Training Selection}\label{sec:in-training}
\begin{figure}[h]
\vspace{-3mm}
\centering
    \subfloat[CIFAR-10]{
    \includegraphics[width=.15\textwidth,height=0.13\textwidth]{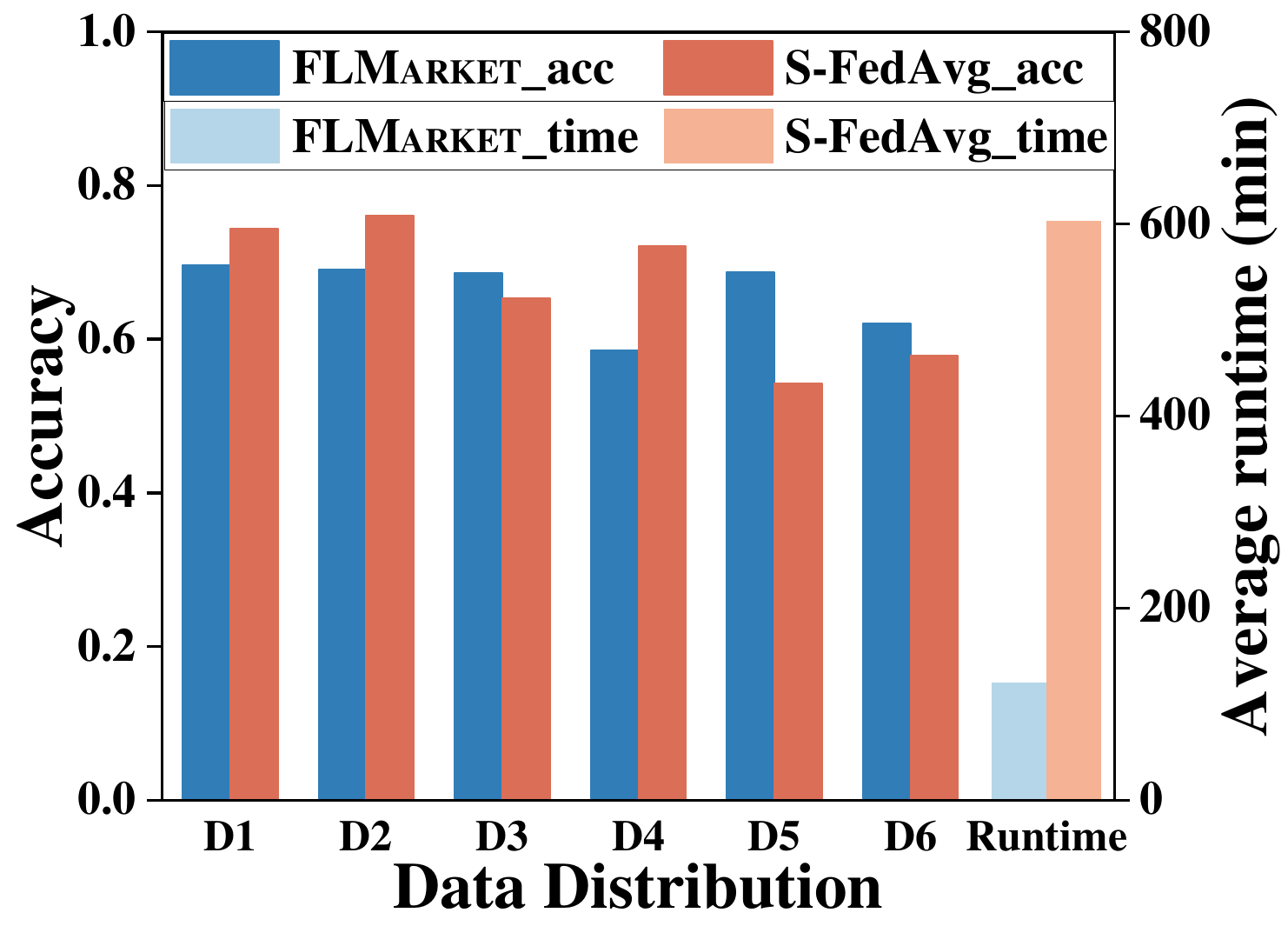}
    \label{fig:sharpley_cifar10}}
    \subfloat[CINIC-10]{
    \includegraphics[width=.15\textwidth,height=0.13\textwidth]{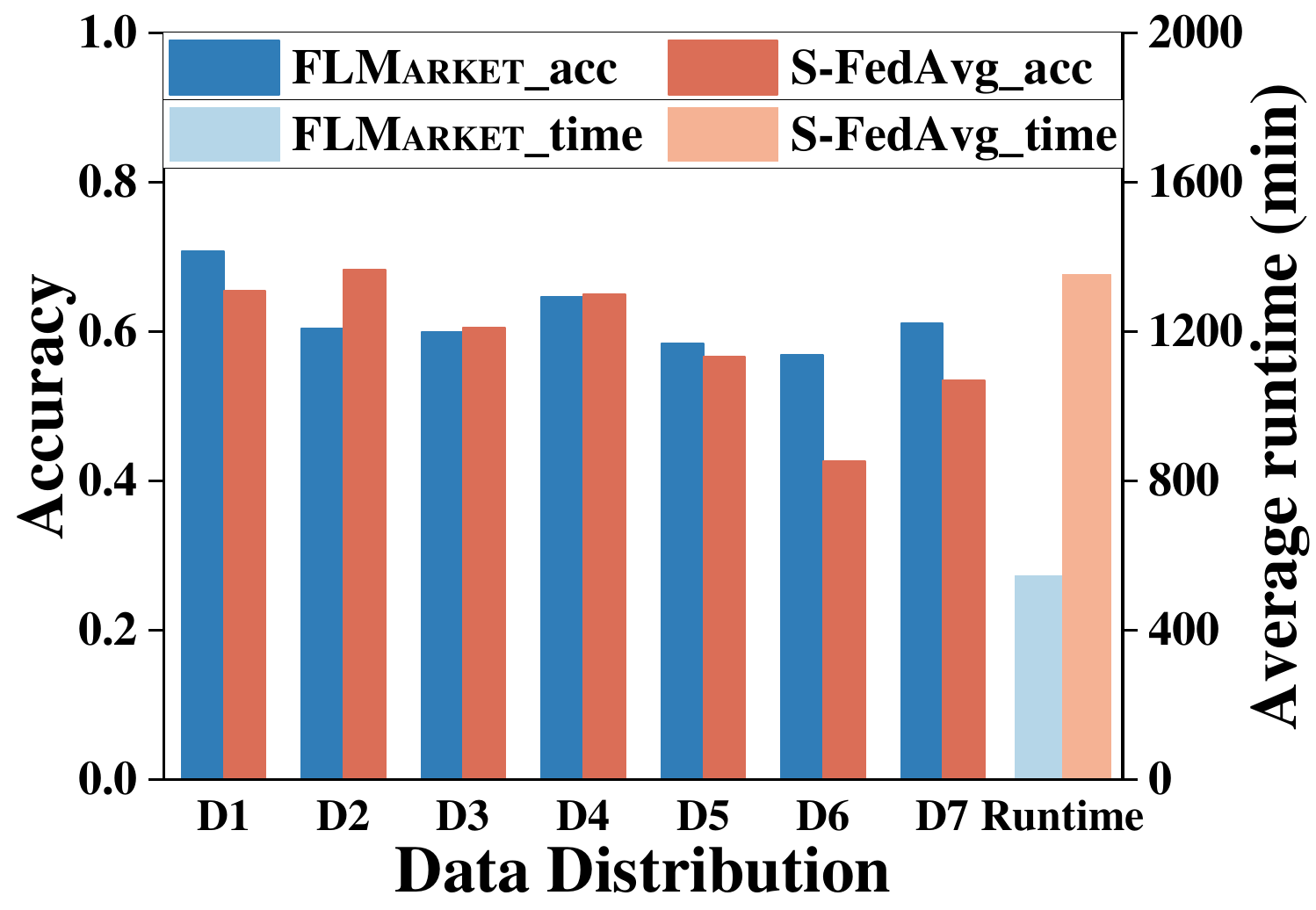}
    \label{fig:sharpley_cinic10}}
    \subfloat[DEAP]{
    \includegraphics[width=.15\textwidth,height=0.13\textwidth]{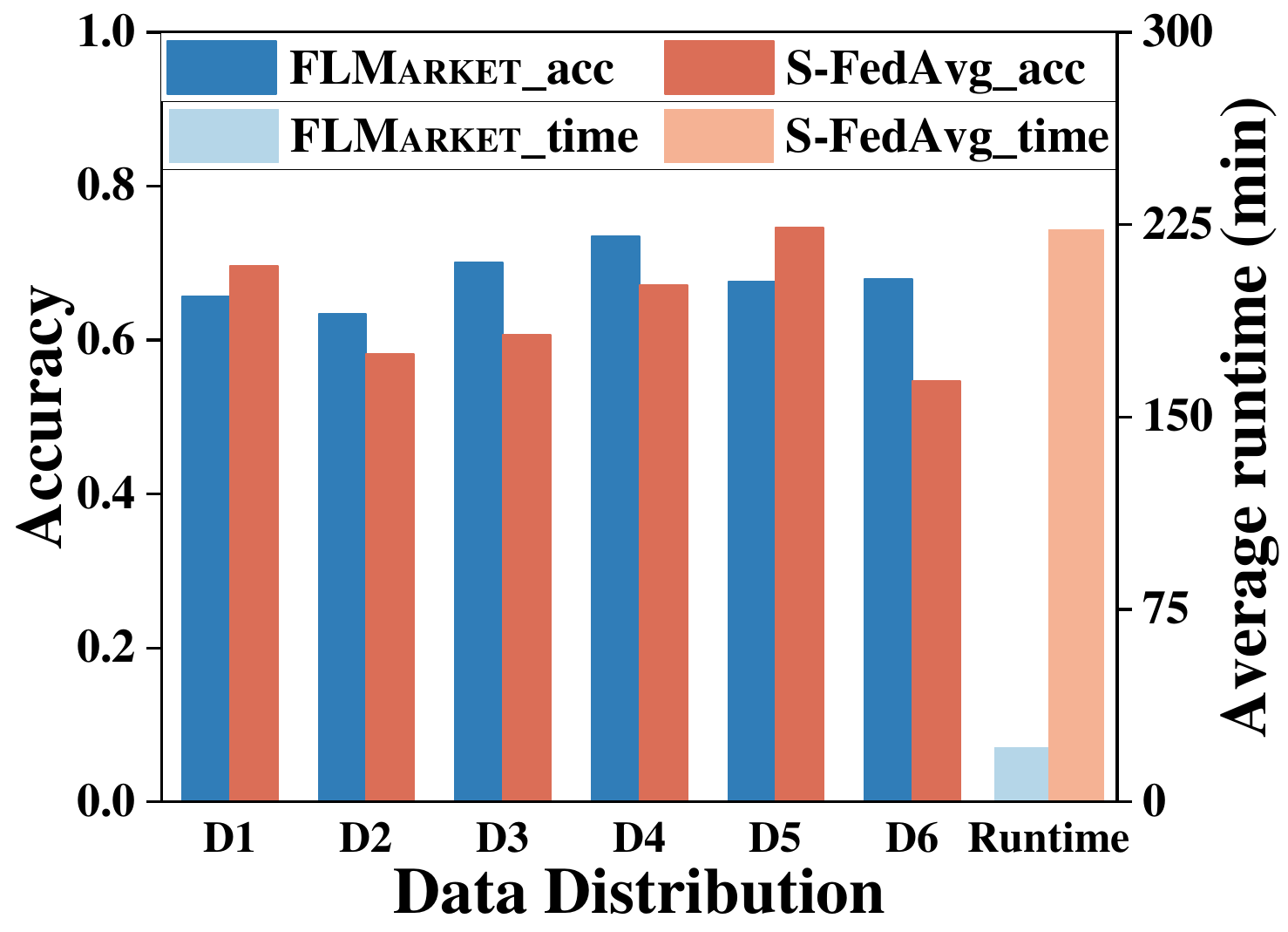}
    \label{fig:sharpley_deap}}
    \vspace{-3mm}
    \caption{Comparison of performance between \projecttitle and S-FedAvg across three datasets}
    \label{fig:sharpley_value}
    \vspace{-4mm}
\end{figure}

In-training client selection methods are impractical for pre-training pricing since they assess clients' contributions during the FL training process. 
Nonetheless, we conducted a comparison on the accuracy between \projecttitle and a popular in-training client selection method, namely S-FedAvg~\cite{nagalapatti2021game}. S-FedAvg samples a subset of clients in each training round based on their Shapley values, which are calculated using their local accuracy. Figure~\ref{fig:sharpley_value} shows the test accuracy and average runtime latency for both \projecttitle and S-FedAvg, with the selection of 5 clients out of 20 clients for all three datasets. It is worth noting that \emph{S-FedAvg sampled 5 clients in every training round, whereas \projecttitle sampled 5 clients out of the 20 only once before the start of training}.

Overall, \projecttitle achieves an average of 66.12\%, 61.75\%, and 68.04\% final accuracy over all data distributions compared to an average of 66.67\%, 58.89\%, and 64.17\% on S-FedAvg. The results surprisingly demonstrate that \projecttitle can achieve comparable or even better accuracy performance than the in-training client selection approach. \projecttitle achieves better accuracy performances on more than half of the experiments. In terms of runtime overhead, \projecttitle outperforms S-FedAvg significantly. Completing 600 rounds of training, S-FedAvg takes 603 minutes, 1354 minutes, and 223 minutes, which is 5 $\times$, 2.5 $\times$, and 11 $\times$ longer than \projecttitle for the CIFAR-10, CINIC-10, and DEAP datasets, respectively.

In summary, compared to pre-training client selection baselines, \projecttitle achieves an average improvement of  10.18\% in accuracy across different datasets, data distributions, and selection modes. When compared to in-training client selection algorithms, \projecttitle still achieves an average improvement of over 2.1\% in accuracy and an average 3.16$\times$ speedup for per-round runtime latency.
\section{Discussion}
\label{sec:discussion}
In this section, we discuss two challenges associated with deploying \projecttitle in real-world FL data markets. It is worth noting that we do not focus on these challenges in the design of our framework because simple modifications or existing solutions can effectively mitigate them. Moreover, these solutions can be seamlessly integrated into our framework, thereby minimizing the impact of these challenges on our core contributions.

\myparagraph{Dynamics of FL data market}
Practical considerations, such as the constraints of truthfulness, individual rationality, and budget, have been integrated into the theoretical design of \projecttitle. However, in real-world FL data markets, more advanced factors, such as market dynamics, may necessitate adaptive solutions for data pricing. For instance, rapid changes in client data or server budgets can influence optimal pricing outcomes. To address this, an adaptive re-launch mechanism can be incorporated into our framework, enabling the proposed bidding process to restart as needed to accommodate these variations.

\myparagraph{Malicious attack}
In designing label-sharing mechanisms, we propose the PASS protocol to prevent the leakage of label information. This protocol prevents the server from accessing sensitive client information. However, it does not guarantee protection against malicious clients who might provide false information or engage in fraudulent training to exploit rewards from the server. For instance, a common type of attack from malicious clients is the free-riding attack~\cite{lin2019free}. To address this, existing in-training free-riding attack detection techniques~\cite{Wang2023IOTJfree, Wang2022inferencedefence} can be integrated into our framework to mitigate these vulnerabilities. Based on the detection results, we can adjust the pre-training rewards to penalize malicious clients.
\vspace{-2mm}
\section{Conclusion}
\label{sec:conclusion}

In this paper, we propose \projecttitle, a privacy-preserved, pricing framework for pre-training client selection in federated learning. We design a truthful auction mechanism that is able to precisely determine the critical value and payment for the participating clients. Based on the aggregated class distribution, \projecttitle incorporates a secure data evaluation function and selects high-quality clients while meeting the budget requirements. Extensive experiments demonstrate that \projecttitle can evaluate the quality of local clients and select the best group of participants from the client pool, outperforming other baselines by a large margin.



\bibliographystyle{ACM-Reference-Format}
\bibliography{main}

\appendix
\section*{APPENDIX}
\section{Survey on the Impact of Pre-training Data Pricing on Participation Willingness in FL}\label{appendix:survey}
In this section, we provide our empirical survey on how data pricing influences users' willingness to participate in FL training. We first present our questionnaire and then provide a description of the participants. Based on the results collected, we then offer our analysis and conclusions, which highlight the significant importance of pre-training pricing in greatly enhancing the participation willingness of end users.

\subsection{Questionnaire}

\myparagraph{Title} Questionnaire on Data Pricing in Federated Learning

\myparagraph{Introduction} 
Welcome to our questionnaire! This survey aims to investigate how the pricing mechanism in federated learning (FL) influences users' participation willingness.

First, we will provide some necessary background information on FL in case you are not familiar with it. Next, participants will be asked four multiple-choice questions about their thoughts on the pricing mechanism in FL. Please select the options that are most suitable for you.

\myparagraph{FL Basics}
\begin{figure}[h]
    \centering
    \includegraphics[width=0.4\textwidth]{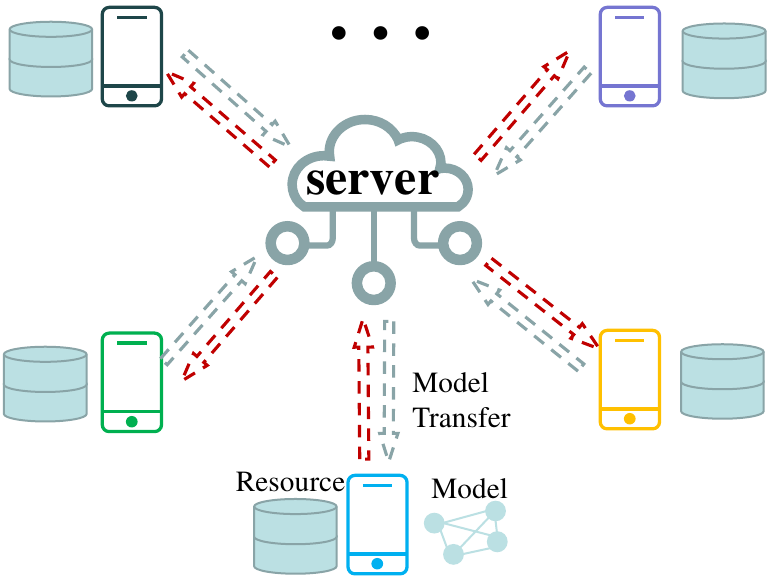}
    \caption{FL training architecture}
    \label{fig:survey}
\end{figure}

\begin{figure*}[t]
\vspace{-4mm}
\centering
    \subfloat[Familiarity with FL]{
    \includegraphics[width=.18\textwidth]{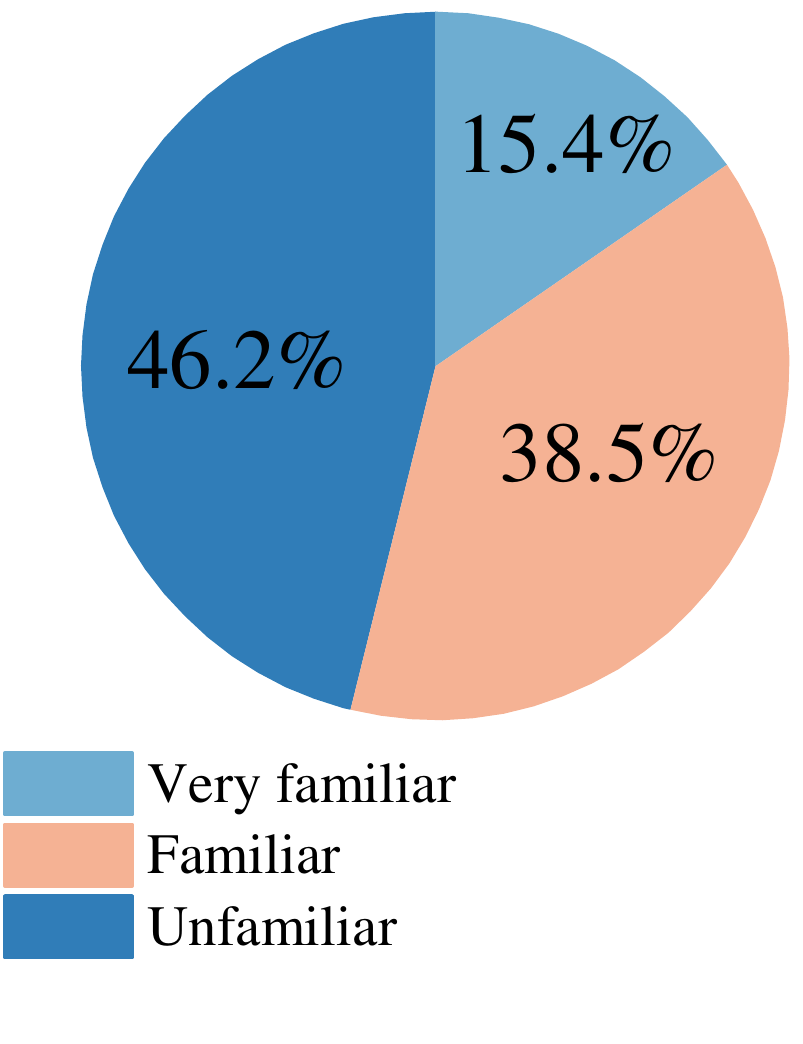}
        \label{fig:Q1}}
    \subfloat[Whether to join]{
    \includegraphics[width=.18\textwidth]{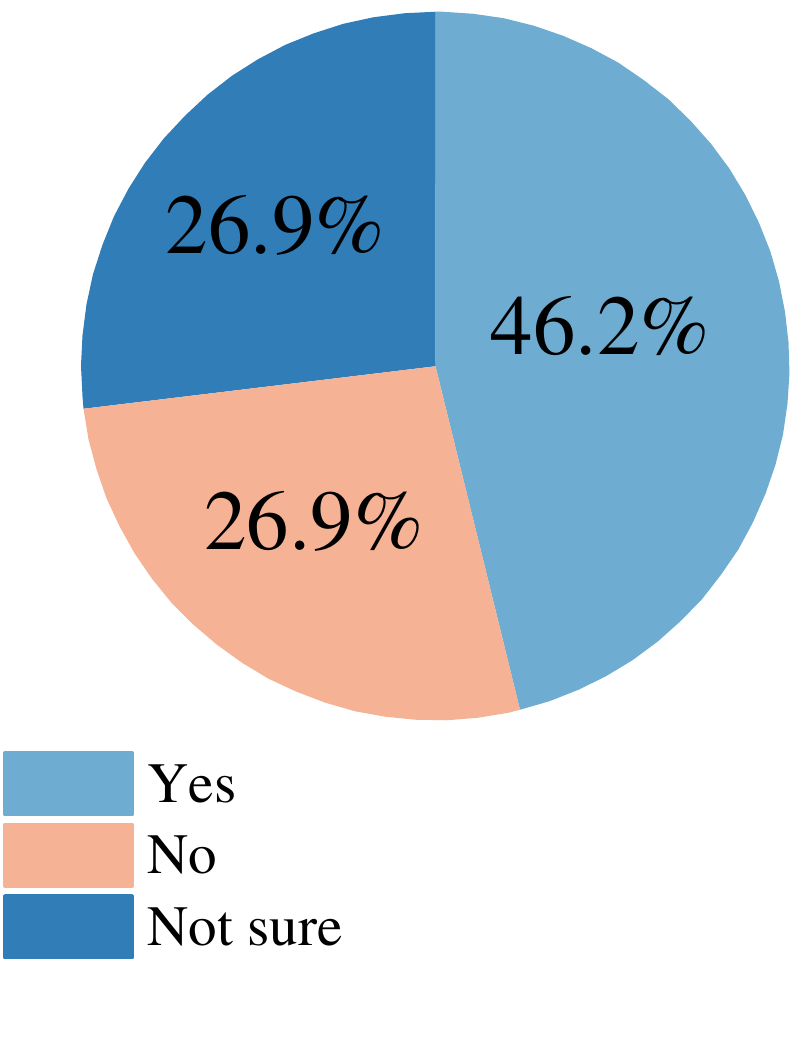}
        \label{fig:Q2}}
    \subfloat[Concerns in FL]{
    \includegraphics[width=.18\textwidth]{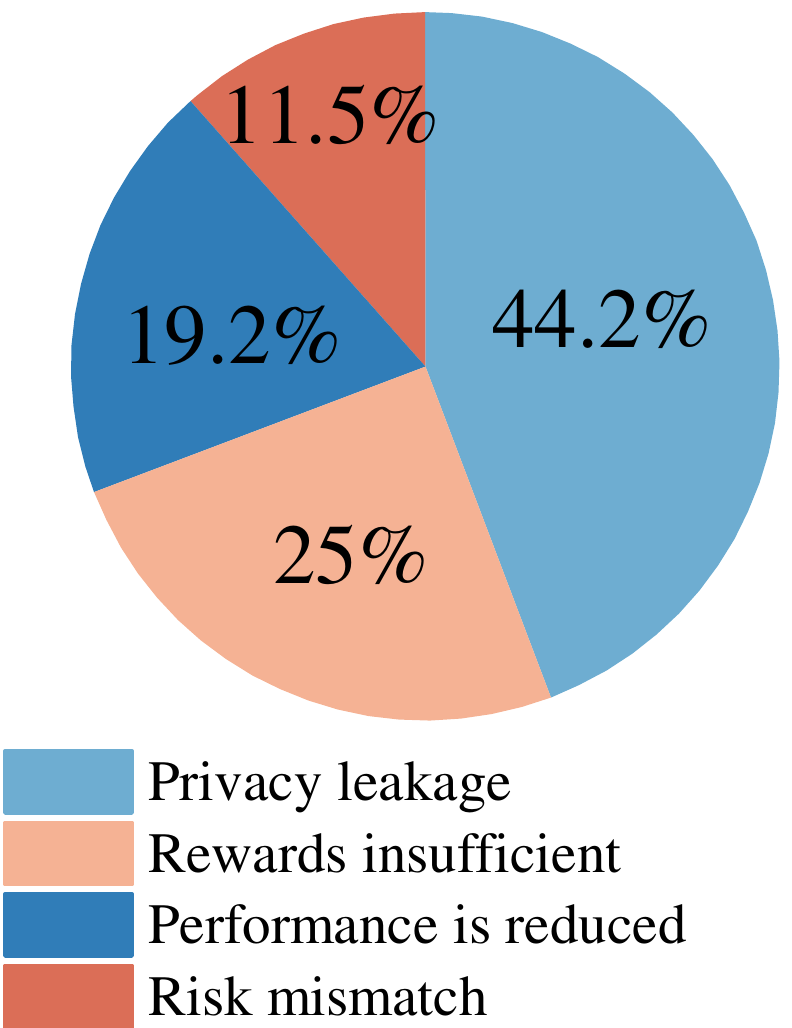}
        \label{fig:Q3}}
    \subfloat[More attractive mechanism]{
    \includegraphics[width=.18\textwidth]{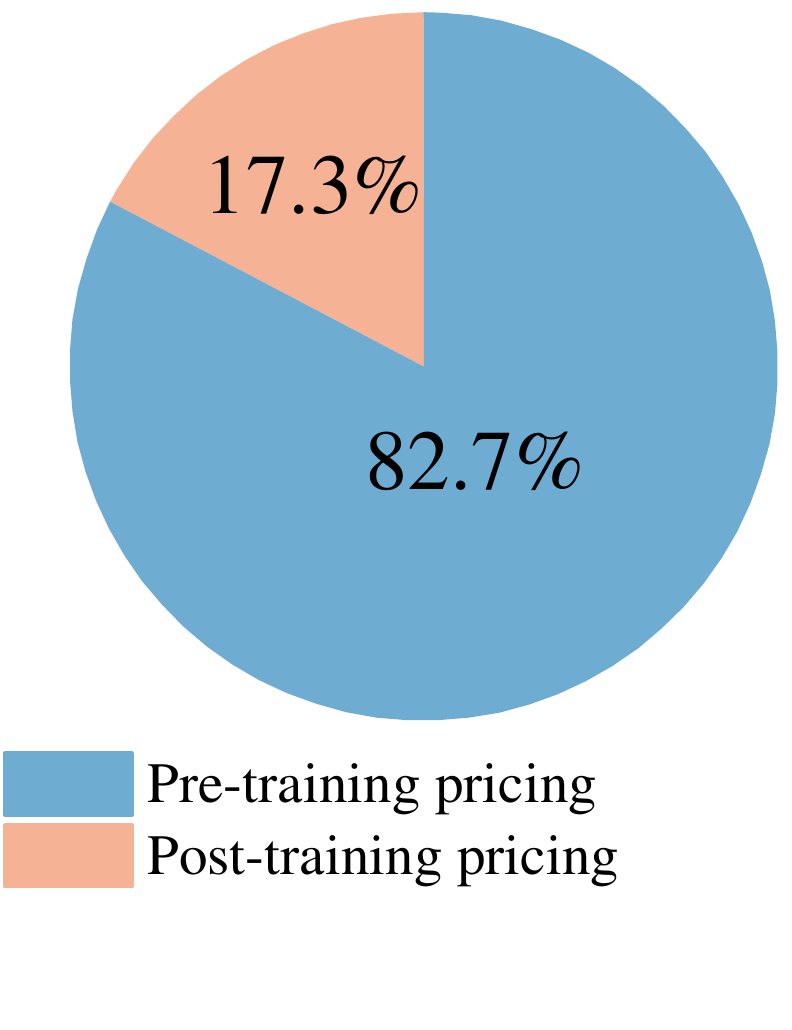}
        \label{fig:Q4}}
        \hspace{.01\textwidth}
    \subfloat[Pre-training pricing impact]{
    \includegraphics[width=.18\textwidth]{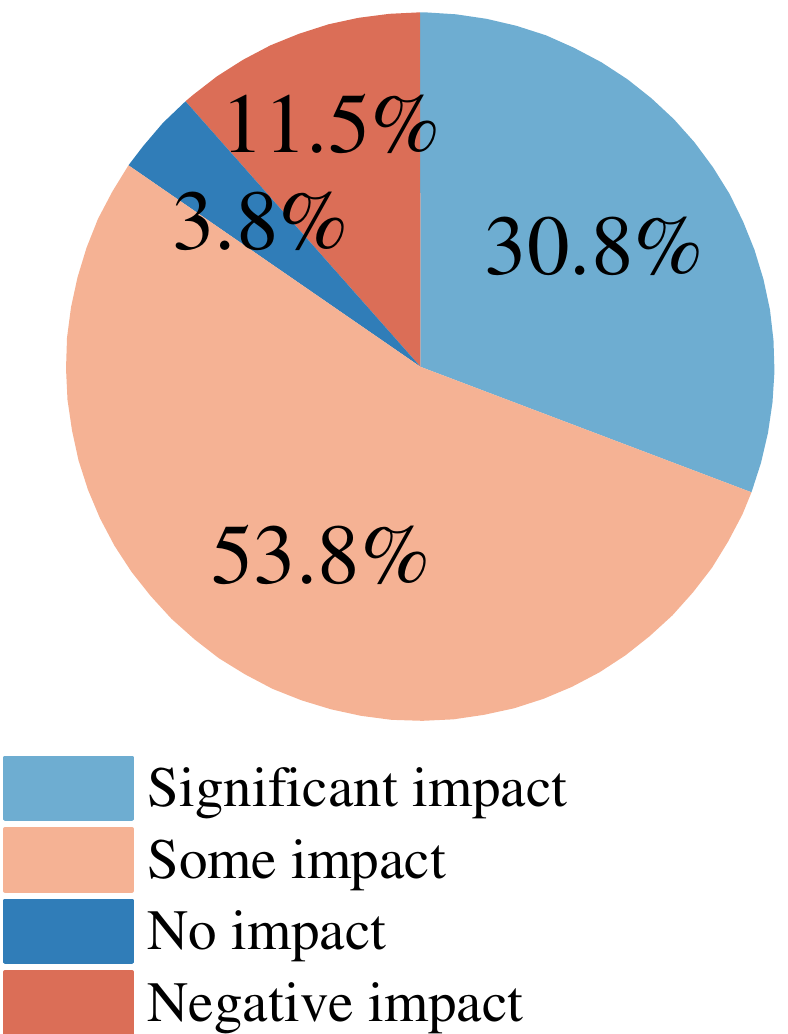}
        \label{fig:Q5}}
    \caption{Results of the questionnaire}
    \label{fig:qustionnaire}
\end{figure*}

\begin{table*}[t]
    \noindent
    \renewcommand{\arraystretch}{1.5}
    \caption{Comparison of functional indicators between \projecttitle with other frameworks}
    \vspace{-3mm}
    \resizebox{0.96\textwidth}{!}{
        \begin{tabular}{cccccccccc}
        \toprule
        \textbf{Functional indicators}&  \textbf{\projecttitle}&  \textbf{AFL}~\cite{zhou2021truthful} &  \textbf{AUCTION}~\cite{deng2021auction} & \textbf{DICE}\cite{saha2022data} & \textbf{DDS} \cite{li2021sample} &martFL\cite{li2023martfl} &DEVELOP\cite{sun2022profit} & SARDA\cite{sun2024socially} \\
        \cmidrule{1-9}
        Pre-Training Auction and Incentive
        & \ding{52} &\ding{52} &\ding{54} &\ding{54}& \ding{54} &\ding{54} &\ding{54}& \ding{54}\\
        \cmidrule{1-9}
        Data Pricing and Client Selection
       & \ding{52} &\ding{52} &\ding{54} &\ding{54} &\ding{54}&\ding{54} &\ding{54} &\ding{54} \\
        \cmidrule{1-9}
        Privacy-aware Client Evaluation
        & \ding{52}&\ding{54} &\ding{52} &\ding{52} &\ding{52}&\ding{52} &\ding{52} &\ding{52}\\
        \cmidrule{1-9}
        Task Budget Control
        & \ding{52} & \ding{54}& \ding{52}&\ding{54}&\ding{54}&\ding{54}&\ding{54}&\ding{54}\\
    \bottomrule
    \end{tabular}
    }
    \label{tab:comparsion}
    \vspace{-3mm}
\end{table*}

Figure~\ref{fig:survey} provides a typical training architecture for FL. In FL training, your role is that of a client, holding your own data on personal devices such as smartphones. The server (e.g., companies) will distribute the training tasks to your smartphones and use your local resources to train a global model. Obviously, the training process will consume your personal resources, such as on-device computation and data communication. As compensation, the server will also offer rewards to the participating clients. FL is more privacy-preserving compared to traditional centralized training since no raw data is transmitted from your device to the server.

\myparagraph{Multiple-Choice Questions}

\question{\noindent{Q1: Are you familiar with FL?}}
\begin{itemize}[leftmargin=10pt]
    \item Very familiar. I have extensive knowledge and experience in the FL. 
    \item Familiar. I have some understanding of the FL.
    \item Unfamiliar. I have little to no understanding or experience in the FL.
\end{itemize}

\question{\noindent{Q2: After having a basic understanding of FL, would you like to join an FL task by using your mobile phone for half an hour? You will receive some monetary rewards.}}
\begin{itemize}[leftmargin=10pt]
    \item Yes, I want to try.
    \item No, I don't want to try.
    \item I am not sure; I need more information to make my decision.
\end{itemize}

\question{\noindent{Q3: Which of the following concerns do you have when participating in FL tasks?}}
\begin{itemize}[leftmargin=10pt]
    \item Privacy leakage. I worry that the server might access my private data on my phone.
    \item The rewards are insufficient to cover the cost.
    \item Mobile phone performance is reduced. I worry that the training will negatively impact my phone's performance.
    \item Risk mismatch. The server does not have corresponding costs for potential breaches, such as failing to compensate the client.
    
\end{itemize}

\question{\noindent{Q4: If you already have information about the training time and resource consumption, which of the following incentive mechanism would be more attractive to you for participating in FL training?}}
\begin{itemize}[leftmargin=10pt]
    \item Pre-training pricing. A pre-training reward is evaluated and promised before training (e.g., you are promised \$10), and the final reward is adjusted after training based on the results.
    \item Post-training pricing. No pre-training rewards are provided, and the final reward is determined after training based on the results.
\end{itemize}

\question{\noindent{Q5: To what extent do you think a pre-training reward affects your willingness to participate in FL?}}
\begin{itemize}[leftmargin=10pt]
    \item A pre-training reward has a significant impact, changing my decision from not participating to participating.
    \item A pre-training reward has some impact, making participation more attractive.
    \item A pre-training reward has no impact.
    \item A pre-training reward has a negative impact, making me less likely to participate.
\end{itemize}

\myparagraph{Participant selection} 
We distributed the questionnaire both in person at the campus and through online advertisements. To ensure a diverse range of participants, we deliberately targeted various groups, including students, teachers, researchers, and engineers.

\subsection{Results Analysis}
After 7 days of the survey, we collected a total of 53 questionnaires on five questions. Figure \ref{fig:qustionnaire} displays pie charts of the options for these five questions. Despite gaining a basic understanding of FL, many participants seemed hesitant to join the FL training. As shown in Figure~\ref{fig:Q2}, about 46.2\% of the respondents expressed a positive willingness to participate, while 26.9\% explicitly refused, and another 26.9\% wanted more information before making a decision. This implies the importance of an incentive mechanism in FL to encourage more participation. Among all the concerns for joining FL training, privacy is the most important factor, with 44.2\% of respondents choosing this option. The second most significant concern is the rewards, accounting for 25\% of responses, as shown in Figure~\ref{fig:Q2}. This highlights that privacy should be the top priority to secure more participants, and reasonable rewards serve as a good incentive.

We then asked participants about their thoughts toward pre-training rewards compared to traditional post-training rewards. The results clearly show that a pre-training rewards mechanism is more effective in encouraging participation in FL. As shown in Figure \ref{fig:Q4}, 82.7\% of respondents find the pre-training pricing mechanism more attractive. This conclusion is further supported by Figure \ref{fig:Q5}, where 84.6\% of respondents believe that pre-training pricing has a positive impact (either significant or some impact) on their decision to participate in FL. 

These findings highlight that ensuring privacy and offering reasonable rewards are crucial to enhancing participation in FL training. Additionally, pre-training incentives are particularly effective in motivating participants.

\section{Comparing \projecttitle with other FL data sharing Frameworks}
\label{related_work_table}

Table~\ref{tab:comparsion} summarizes the set of functional indicators discussed in
several recent papers, and outlines whether a framework supports a functional indicator or not.

\section{Notation}
Table~\ref{tab:notation} summarizes notations used in this paper.

\begin{table}[H]
\noindent
\renewcommand\tabcolsep{5.5pt}
\caption{Notations in \projecttitle}
\resizebox{0.49\textwidth}{!}{  
\label{tab:notation}
\begin{tabular}{cl}
\toprule
    Notation  &Description \\
    \hline
    $\mathbb{T}$, $\mathbb{S}_k$, $\mathbb{E}$ & FL task, selected clients, client pool \\
    $\mathbb{N}_s$, $\mathbb{N}_{e}$ & Global class distribution vector,\\ & Local class distribution vector for client $e$ \\ 
    $S$& Server\\
    $E$ & Total number of clients\\
    $R$ & The budget for task $\mathbb{T}$ \\
    $C$, $c$ & Volume of classes in a data set, class $c$ \\
    $b_{e}$, $\mathbb{B}$  & Bid(s) from client $e$ and all clients\\
    $u_e$, $\mathbb{U}$ & Score(s) for client $e$ and all clients\\
    $\mathbb{V}$  & list of all clients sorted by score per bid\\
    $p_e$, $\mathbb{P}$  & Second-stage price(s) for client $e$ and all selected clients \\
    $N_s$, $N_e$ & Volume of data for global, client $e$ \\
    $n_s^c$, $n_{e}^{c}$ & Volume of data for class $c$ on global and class $c$ on client $e$\\ 
    $\rho(\cdot),\alpha$ &  Normalization function, threshold parameter\\
    $\theta_{c}$ & Weighted coefficient of category\\
    $\phi(\cdot)$ & Data quantity function for each class $c$ on client $e$ \\ 
    $SK$, $PK$ & Private key, public key\\
    $s_{e,v}$ & Random seed \\
    $\epsilon_{e,v}$ & Positive and negative sign\\
    $r$,$R$ & Security parameter, public parameter\\
    $\mathbb{RV}$ & Signed random vector\\
    $\mathbb{Y}_e$ & Pseudo data distribution for client $e$\\
    \bottomrule
\end{tabular}
}
\end{table}

\section{Choosing $f(\cdot)$ and $\theta_{c}$ in Evaluation Function}\label{sec:sen_analysis}

We use CIFAR-10 dataset to study the effectiveness of the score function ${u}_{e}$ as shown in Equation~\ref{eq:evaluation_function} (\S 2.2). We configure six different distributions from this dataset for our evaluation.
\begin{figure}[h]
\vspace{-3mm}
\centering
    \subfloat[Test accuracy with or without varying $\theta_{c}$]{
    \includegraphics[width=0.23\textwidth]{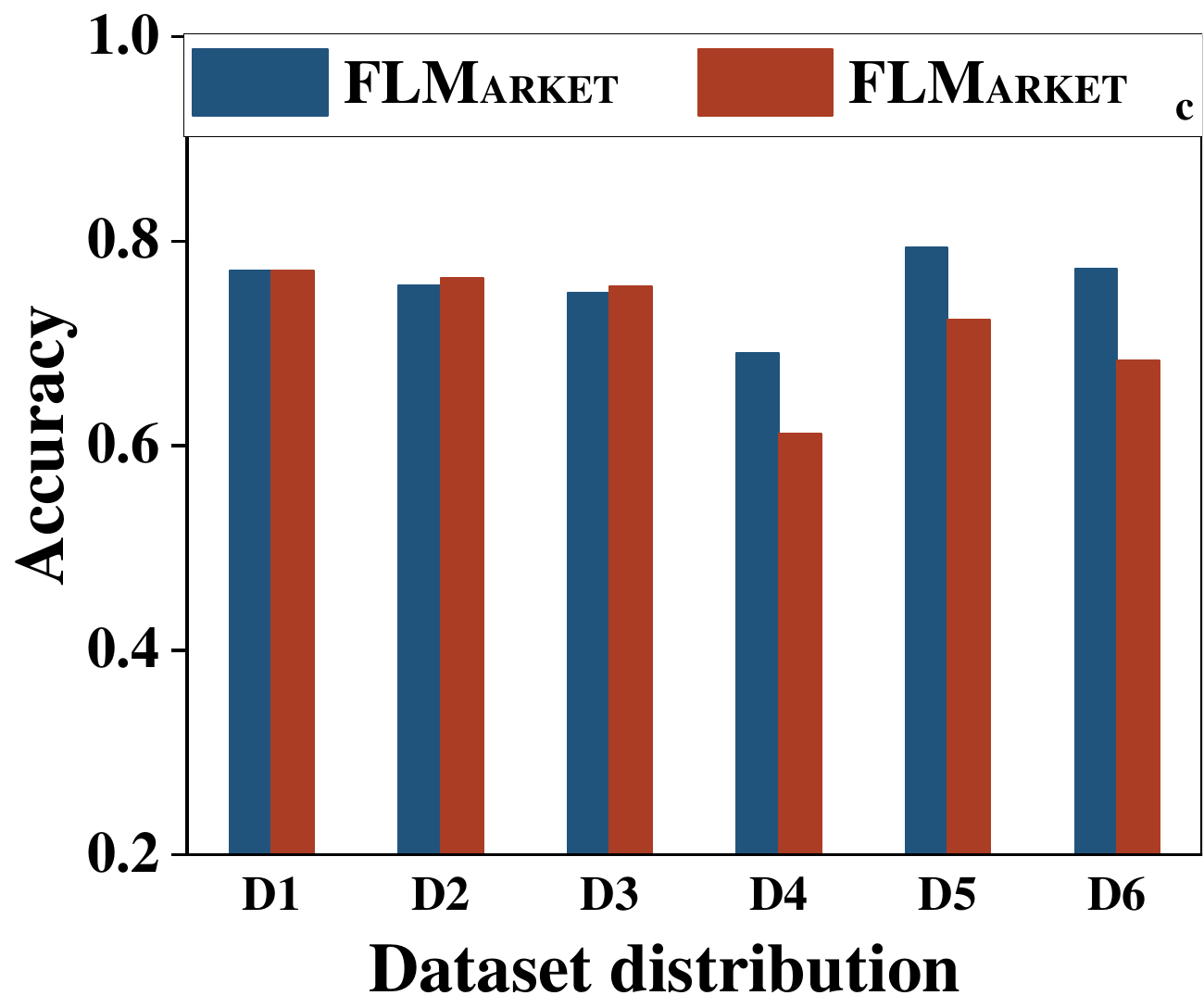}
    \label{fig:test_theta}
    }
    \subfloat[Test accuracy with different $f(\cdot)$ functions]{
     \includegraphics[width=0.23\textwidth]{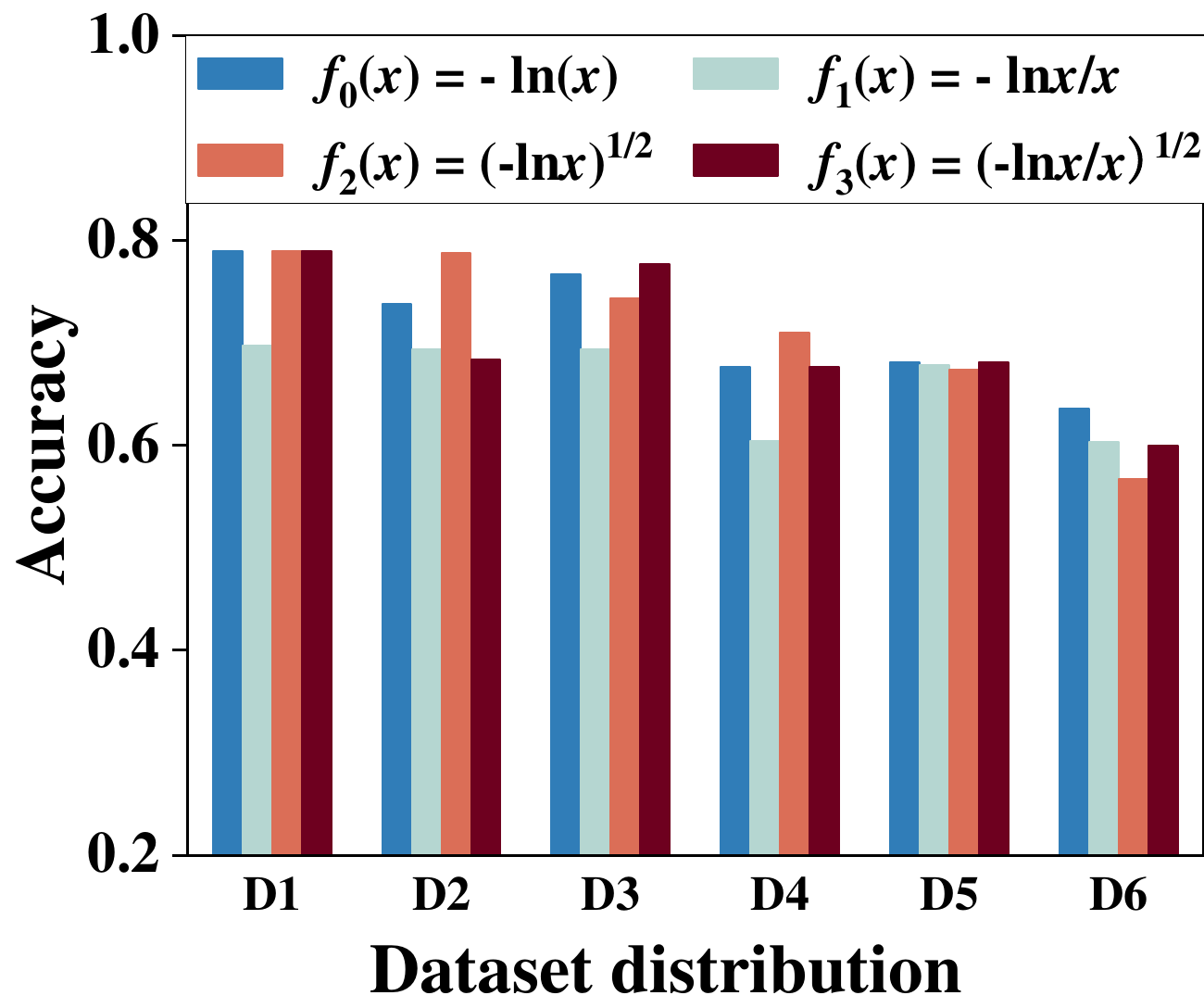}
    \label{fig:test_phi}
    }
    \vspace{-3mm}
    \caption{The impact of different category coefficients $\theta_{c}$ and $f(\cdot)$}
    \label{fig:data_evaluation_func}
    \vspace{-3mm}
\end{figure}

\myparagraph{The choice of $f(\cdot)$} 
$f(\cdot)$ represents the relationship between data volume and model accuracy. This relationship was previously formalized by Ben-David. et. al.~\cite{ben2010theory} that the loss function scales at an $O(d \log(2m)/m )$ rate with respect to the sample size m as shown in lemma~\ref{lemma:function}. 
In addition, the scale laws in large language models also imply a similar logarithmic relation with diminishing returns of more training data~\cite{kaplan2020scaling}.
Therefore, we propose that the function $f(\cdot)$ should approximately follow this law. 
\begin{lemma}[\cite{ben2010theory}]
\label{lemma:function}
    Let $\mathcal{H}$ be a hypothesis space on $\mathcal{X}$ with VC dimension $d$. If $\mathcal{U}$ and $\mathcal{U}'$ are samples of size $m$ from $\mathcal{D}$ and $\mathcal{D}'$ respectively and $\hat{d}_\mathcal{H} (\mathcal{U},\mathcal{U}')$ is the empirical $\mathcal{H}$-divergence between samples, then for any $\delta \in (0,1)$, with probability at least $1-\delta$,
    \begin{equation}
        d_\mathcal{H}(\mathcal{D},\mathcal{D'}) \leq \hat{d}_\mathcal{H}(\mathcal{U},\mathcal{U}')+4\sqrt{\frac{d\log (2m)+\log(\frac{2}{\delta})}{m}}
    \end{equation}
\end{lemma}
We also empirically evaluated four functions, including
\begin{equation}\nonumber
\begin{aligned}
    f_{0}(x)=-\ln(x), &\qquad f_{1}(x)=-\ln(x)/x\\
    f_{2}(x)=\sqrt{-\ln(x)}, &\qquad f_{3}(x)=\sqrt{-\ln(x)/x}
\end{aligned}
\end{equation}
for scoring each clients. Figure~\ref{fig:test_phi} shows that the average accuracy of $f_{0}(x)=-\ln(x)$ across six distributions is 1.71\% higher than the average accuracy of the other three functions, and the variance across the six distributions is 0.56\% lower than the average variance of the other three functions. These results are consistent with the theoretical work and recent observations.

\myparagraph{The Impact of Different Category Coefficients $\theta_{c}$} To illustrate the impact of different $\theta_{c}$ coefficients, we conducted comparative experiments with varying data category $\theta_{c}$ coefficients and identical $\theta_{c}$ coefficients. Figure~\ref{fig:test_theta} demonstrates that variable $\theta_{c}$ values achieve an average 3.76\% higher accuracy than constant $\theta_{c}$ values. Notably, within the context of the unbalanced distributions of D4 to D6, the enhancements in accuracy due to varying $\theta_{c}$ are more obvious, resulting in improvements of 7.89\%, 7.04\%, and 8.92\% compared to constant $\theta_{c}$ values. This illustrates the advantage of assigning different weight coefficients to different data categories.

\section{Proof}
\subsection{Proof of the Lemma~\ref{lemma:monotone}}\label{appendix:proof_of_monotone}
\begin{proof}
    To prove the winner selection algorithm is monotone, we have to show that any winner $e \in \mathbb{S}_k$ will still be selected as it decreases its bid, i.e., $b_e^{\prime} < b_e$.
    
    When $b_e^{\prime} < b_e$, its score per bid $u_e/b_e^{\prime}$ increases. Thus, in the sorted list $\mathbb{V}$, the new position index $e^{\prime}\leq e$. 
    According to budget constraint,
    \begin{equation}
        b_e^{\prime}<b_e\leq \frac{R}{2}\cdot \frac{u_e}{U(\mathbb{S}_k\cup \{e\})}.
    \end{equation}
    Therefore the new bid $b_e^{\prime}$ is consistent with budget constraints and $e$ will still be selected at a lower bid.
\end{proof}

\subsection{Proof of the Lemma~\ref{lemma:critical_price}}\label{appendix:proof_of_critical_price}
\begin{proof}
    We need to prove when $e$ claims a bid $b_e^{\prime}\leq p_e$ will lose the auction and $b_e^{\prime}>p_e$ will win the auction. 
    As we mentioned above that $\hat{k}$ is the smallest index satisfies the budget constraint condition in $\mathbb{V}^{-e}$. Therefore, let's set $r\in [1,\hat{k}+1]$ indicate the index of maximum $p_{e(j)}^{\prime}$ in $\mathbb{V}^{-e}$. i.e., the payment of $e$ is $p_e=p_{e(r)}^{\prime}$. 
    
    When $b_e^{\prime}\leq p_e$, according to the definition of $p_e$, we know $b_e^{\prime}\leq p_e=p^{\prime}_{e(r)}=\min\{\lambda_{e(r)},\beta_{e(r)}\}$. i.e, $b_e^{\prime}\leq \lambda_{e(r)}$ and $b_e^{\prime}\leq \beta_{e(r)}$. 
    We obtain the following inequality:
    \begin{equation}
        b_e^{\prime}\leq \lambda_{e(r)}=\frac{u_e\cdot b_r}{u_r}\Rightarrow \frac{u_e}{b_e^{\prime}}\geq\frac{u_r}{b_r}.
    \end{equation}
    Therefore, $e$ will take the place of $r$ in $\mathbb{V}$ and win the auction.

    As for $b_e^{\prime}>p_e$, we consider the following two scenarios.

    \begin{itemize}[leftmargin=10pt]
        \item $\lambda_{e(r)}\leq \beta_{e(r)}$. The payment $p_e=p_{e(r)}^{\prime}=\min\{\lambda_{e(r)},\beta_{e(r)}\}=\lambda_{e(r)}$, so $b_e^{\prime}>\lambda_{e(r)}$ since $b_e^{\prime}>p_e$.
        Therefore, we can deduce that $u_e/b_e^{\prime}<u_r/b_r$ and $e$ is behind $r$ in $\mathbb{V}$.
        Next, we consider the list $[r+1,\hat{k}+1]$ that ranks behind $r$ in $\mathbb{V}^{-e}$. Let $j\in[r+1,\hat{k}+1]$, if $\lambda_{e(r)}\geq\lambda_{e(j)}$ then $e$ will not take part of $j$ in $\mathbb{V}$ since $b_e^{\prime}>\lambda_{e(r)}\geq \lambda_{e(j)}$. So $e$ will lose the auction. If $\lambda_{e(r)}<\lambda_{e(j)}$ then we get the inequality:
        \begin{equation}\label{eq:lambda_price}
            \lambda_{e(j)}>\lambda_{e(r)}=p_{e(r)}^{\prime}>p_{e(j)}^{\prime}.
        \end{equation} 
        If $p_{e(j)}^{\prime}=\lambda_{e(j)}$ then $p_{e(j)}^{\prime}=\lambda_{e(j)}>\lambda_{e(r)}=p_{e(r)}$ which has contradiction with Inequality (\ref{eq:lambda_price}). Therefore, $\lambda_{e(j)}> p_{e(j)}^{\prime}=\beta_{e(j)}$ and $e$ will lose the auction because $b_e^{\prime}>\lambda_{e(r)}>\beta_{e(j)}$ which violates the budget constraint in location $j$.
        
        
        \item $\lambda_{e(r)}> \beta_{e(r)}$. The payment $p_e=\lambda_{e(r)}$, so $b_e^{\prime}>\beta_{e(r)}$.
        Considering that in $\mathbb{V}^{-e}$, $j\in[1,\hat{k}+1]$. If $\beta_{e(r)}>\beta_{e(j)}$ then $b_e^{\prime}>\beta_{e(r)}>\beta_{e(j)}$. Therefore $e$ will not win the auction in $\mathbb{V}$ because of the budget constraint. If $\beta_{e(r)}\leq\beta_{e(j)}$, we can get 
        \begin{equation}
          \beta_{e(j)}\geq p_{e(r)}^{\prime}=\beta_{e(r)}>p_{e(j)}^{\prime}  
        \end{equation}
        
        We know that $p_{e(r)}^{\prime}=\beta_{e(r)}$ and $\beta_{e(r)}\leq \beta_{e(j)}$, if $p_{e(j)}^{\prime}=\beta_{e(j)}$ then $p_{e(j)}^{\prime}\geq p^{\prime}_{e(r)}$ which contradicts with definition of $p_{e(r)}^{\prime}$. Therefore, $p_{e(j)}^{\prime}=\lambda_{e(j)}$ and then $b_e^{\prime}>\beta_{e(r)}>\lambda_{e(j)}$. Thus, $e$ will be behind $j$ in $\mathbb{V}$ and lose the auction.
    \end{itemize}
    
    Therefore, in any case, when $e$ claims a bid bigger than $p_e$, it will lose the auction.
\end{proof}

\subsection{Proof of the Lemma~\ref{lemma:extra_lemma}}\label{appendix:proof_of_lemma}
\begin{proof}
     Let's prove the Lemma \ref{lemma:extra_lemma}. To simplify the symbolic representation, we use $\mathbb{S}_{\{2\setminus1\}\setminus\hat{e}}$ for $\{\mathbb{S}_2\setminus\mathbb{S}_1\}\setminus\{\hat{e}\}$. Based on the knowledge presented earlier, we obtain the following equation.
       \begin{equation}
           \frac{U(\mathbb{S}_2)-U(\mathbb{S}_1)}{\sum_{i\in\mathbb{S}_2}b_i-\sum_{j\in\mathbb{S}_1}b_j}=\frac{\sum_{e\in\mathbb{S}_2\setminus\mathbb{S}_1}u_e}{\sum_{e\in\mathbb{S}_2\setminus\mathbb{S}_1}b_e}
       =\frac{\sum_{e\in\mathbb{S}_{\{2\setminus1\}\setminus\hat{e}}}u_e+u_{\hat{e}}}{\sum_{e\in\mathbb{S}_{\{2\setminus1\}\setminus\hat{e}}}b_e+b_{\hat{e}}}
       \end{equation}
       
       We know that for $e\in\mathbb{S}_2\setminus\mathbb{S}_1$, $u_{\hat{e}}/b_{\hat{e}}\geq u_e/b_e$. Then we can get: 
       \begin{equation}\label{eq:subset_leqargmax}
            \begin{aligned}
            &\frac{u_{\hat{e}}}{b_{\hat{e}}}-\frac{\sum_{e\in\mathbb{S}_{\{2\setminus1\}\setminus\hat{e}}}u_e+u_{\hat{e}}}{\sum_{e\in\mathbb{S}_{\{2\setminus1\}\setminus\hat{e}}}b_e+b_{\hat{e}}}\\
            =&
            \frac{u_{\hat{e}}\cdot(\sum_{e\in\mathbb{S}_{\{2\setminus1\}\setminus\hat{e}}}b_e+b_{\hat{e}})
            -b_{\hat{e}}\cdot(\sum_{e\in\mathbb{S}_{\{2\setminus1\}\setminus\hat{e}}}u_e+u_{\hat{e}})}
            {b_{\hat{e}}\cdot(\sum_{e\in\mathbb{S}_{\{2\setminus1\}\setminus\hat{e}}}b_e+b_{\hat{e}})}\\
            =&\frac{u_{\hat{e}}\cdot\sum_{e\in\mathbb{S}_{\{2\setminus1\}\setminus\hat{e}}}b_e
            -b_{\hat{e}}\cdot\sum_{e\in\mathbb{S}_{\{2\setminus1\}\setminus\hat{e}}}u_e}
            {b_{\hat{e}}\cdot(\sum_{e\in\mathbb{S}_{\{2\setminus1\}\setminus\hat{e}}}b_e+b_{\hat{e}})}\\
            =&\frac{\sum_{e\in\mathbb{S}_{\{2\setminus1\}\setminus\hat{e}}}(u_{\hat{e}}\cdot b_e-b_{\hat{e}}\cdot u_e)}
            {b_{\hat{e}}\cdot(\sum_{e\in\mathbb{S}_{\{2\setminus1\}\setminus\hat{e}}}b_e+b_{\hat{e}})}
            \end{aligned}
       \end{equation}
       
      Since $u_{\hat{e}}/b_{\hat{e}}>u_e/b_e$ for $e\in\mathbb{S}_{\{2\setminus1\}\setminus\hat{e}}$. Thus $u_{\hat{e}}\cdot b_e-b_{\hat{e}}\cdot u_e>0$, then the Equation (\ref{eq:subset_leqargmax}) is positive and we get that:
      \begin{equation}
          \frac{u_{\hat{e}}}{b_{\hat{e}}}>\frac{\sum_{e\in\mathbb{S}_{\{2\setminus1\}\setminus\hat{e}}}u_e+u_{\hat{e}}}{\sum_{e\in\mathbb{S}_{\{2\setminus1\}\setminus\hat{e}}}b_e+b_{\hat{e}}}=\frac{U(\mathbb{S}_2)-U(\mathbb{S}_1)}{\sum_{i\in\mathbb{S}_2}b_i-\sum_{j\in\mathbb{S}_1}b_j}
      \end{equation}
      
      Therefore, 
      \begin{equation}
            \frac{U(\mathbb{S}_2)-U(\mathbb{S}_1)}{\sum_{i\in\mathbb{S}_2}b_i-\sum_{j\in\mathbb{S}_1}b_j}< \frac{u_{\hat{e}}}{b_{\hat{e}}} 
       \end{equation}
\end{proof}

\section{The PASS Protocol}\label{pass_protocol}
\begin{figure}[t]
\centering
\includegraphics[width=0.45\textwidth]{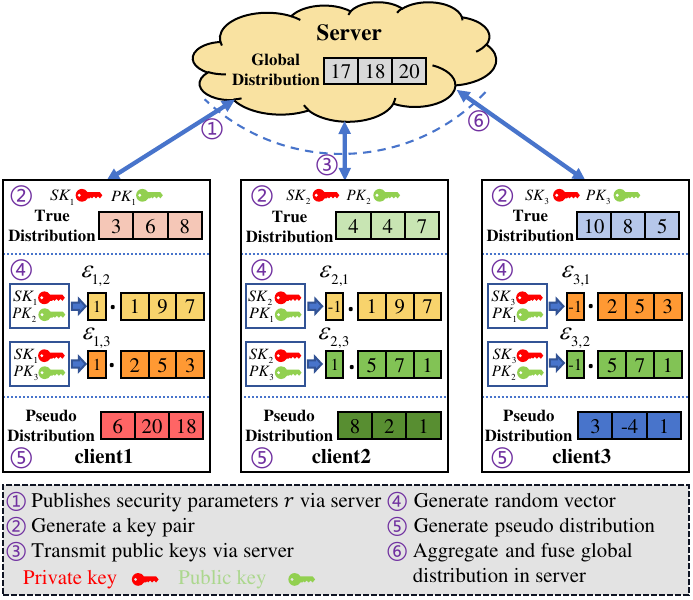}
    \caption{The example of PASS}
    \label{fig:ASS-GAD-EP}
\end{figure}

Table \ref{tab:protocol} shows the protocol of the $PASS$ to aggregate data distribution from clients and Figure~\ref{fig:ASS-GAD-EP} indicates a running example of  $PASS$.
\begin{table}[h]
    \noindent
    \caption{The PASS protocol}
    \vspace{-3mm} 
    \resizebox{0.45\textwidth}{!}{
    \label{tabel5}
        \begin{tabular}{p{8cm}}
        \toprule
        \textbf{Set up:} \\
        -- All clients are given the security parameter $r$ by the server $S$.\\
        \textbf{Step 1:} \\
        \emph{Client $e$}:\\
        -- Honestly generate $R\leftarrow KA.param(r)$ and generate key pairs $(PK_e,SK_e)\leftarrow KA.gen(R)$.\\
        -- Send $PK_e$ to the server $S$.\\
        \emph{Server $S$}:\\
        -- Receive public keys $PK_{e,e\in \mathbb{E}}$ from clients and broadcast ${(PK_e,e)}_{e\in \mathbb{E}}$ to every client.\\
        \textbf{Step 2}:\\
        \emph{Client $e$}:\\
        -- Received the list $\{(PK_v,v)\}_{v\in \mathbb{E}}$ broadcast from the server $S$.\\
        -- For each client $v\in \mathbb{E}\setminus \{e\}$, generated random seed $s_{e,v}\leftarrow KA.agree(SK_e,PK_v)$.\\ 
        -- Based on $s_{e,v}$ generate $\mathbb{RV}_{e,v}\leftarrow \epsilon_{e,v}\cdot PRG(s_{e.v})$ using PRG, where $\epsilon_{e,v}=1$ if $e<v$ and $\epsilon_{e,v}=-1$ if $e>v$.\\
        -- Adding $\mathbb{N}_e$ to all $\mathbb{RV}_{e,v}$ yields the pseudo local data distribution $\mathbb{Y}_e\leftarrow \mathbb{N}_e+\sum_{v\in \mathbb{E}\setminus\{e\}}\mathbb{RV}_{e,v}$ and send it to the server.\\
        \emph{Server $S$}:\\
        -- Receive pseudo local data distribution $Y_{e,e\in \mathbb{E}}$ from clients .\\
        -- Summing all $\mathbb{Y}_e$ to get the global distribution $\mathbb{N}_s\leftarrow\sum_{e\in \mathbb{E}}\mathbb{Y}_e $ and broadcast it to all clients.\\
        \bottomrule
    \end{tabular}
    }
    \label{tab:protocol}
    \end{table}

    \myparagraph{Example} The example of $PASS$ is in Figure~\ref{fig:ASS-GAD-EP}. Assume there are 3 clients willing to participate in the FL training. Their true data distribution is $[3,6,8], [4,4,7]$, and $[10,8,5]$ respectively, where the number represents the data volume of the class. For instance, the data volume of class 1, class 2, and class 3 in Client1 is 3, 6, and 8 respectively. Client1 uses private key $SK_1$ and public key $PK_2,PK_3$ generates vector $[1,9,7]$ and $[2,5,3]$ respectively. Similarly, Client2 generates vector $[1,9,7]$ and $[5,7,1]$, Client3 generates vector $[2,5,3]$ and $[5,7,1]$. 
$\epsilon_{1,2}=1$ for $1<2$ and $\epsilon_{1,3}=1$ for $1<3$, so Client1 adds $1\cdot[1,9,7]$ and $1\cdot[2,5,3]$ generates pseudo data distribution $[6,20,18]=[3,6,8]+[1,9,7]+[2,5,3]$. 
Similarly, Client2 adds $-1\cdot[1,9,7]$ and $1\cdot[2,5,3]$ to generates pseudo data distribution $[8,2,1]=[4,4,7]-[1,9,7]+[5,7,1]$. 
Client3 adds $-1\cdot[2,5,3]$ and $-1\cdot[5,7,1]$ to generates pseudo data distribution $[3,-4,1]=[10,8,5]-[2,5,3]-[5,7,1]$. Thereafter, this information is uploaded to the central server, the global data distribution is computed accordingly $[17,18,20]=[6,20,18]+[8,2,1]+[3,-4,1]$.

\section{Security Analysis of PASS}
\label{sec:analyse of privacy}
PASS reveals little additional information with others beyond the pseudo-distribution and public key, which reduces the risk of privacy breaches. Next, we analyse the privacy protection ability of $PASS$ in the context of limited sharing of information.

\begin{theorem}
Any combination based on client-generated pseudo data distributions is computationally indistinguishable from a uniformly sampled element $\mathbb{Z}$ of the same space by the server, except for the aggregation combining all clients.
\end{theorem}

\begin{proof}

    For any $\{\mathbb{N}_e\}_{e\in \mathbb{E}}$, where $\forall e \in \mathbb{E}$, $\mathbb{N}_e\in \mathbb{R}^C$. If $e$ has the $\mathbb{RV}_{e,v}\in \mathbb{R}^C,$ and $v$ has the corresponding $\mathbb{RV}_{v,e}=-\mathbb{RV}_{e,v}$.
    Then each client $e$ adds all $\mathbb{RV}_{e,v}$ with $\mathbb{N}_e$ gets:
    \begin{gather}
        \mathbb{Y}=\{\mathbb{N}_e+\sum_{v\in \mathbb{E}\setminus \{e\}} \mathbb{RV}_{e,v}\}_{e\in \mathbb{E}}
    \end{gather}
    Considering any combination from $\mathbb{Y}$, it can be defined as:
    \begin{gather}
        \sum_{e\in \mathbb{I}\subseteq \mathbb{E}}\{\mathbb{N}_e+\sum_{v\in \mathbb{E}\setminus \{e\}} \mathbb{RV}_{e,v}\}
    \end{gather}
    
    Then we compare it with a random generated element $\mathbb{Z}$ of the same space $ \mathbb{Z} \in \mathbb{R}^C$.
Then we can consider: 
    \begin{equation}
        \sum_{e\in \mathbb{I}\subseteq \mathbb{E}}\{\mathbb{N}_e+\sum_{v\in \mathbb{E}\setminus \{e\}} \mathbb{RV}_{e,v}\} \equiv \mathbb{Z} \qquad s.t.\quad \mathbb{I} \neq \mathbb{E}
    \end{equation}
When $\mathbb{I} \neq \mathbb{E}$, the sum of the client-generated pseudo distributions looks random. In other words, only when all the client-generated pseudo data distributions are aggregated, the server can obtain a non-random distribution, which is the global distribution we desire.   
\end{proof}

\begin{figure*}[h]
\centering
    \subfloat[Distribution List of CIFAR-10]{
    \includegraphics[width=0.4\textwidth]{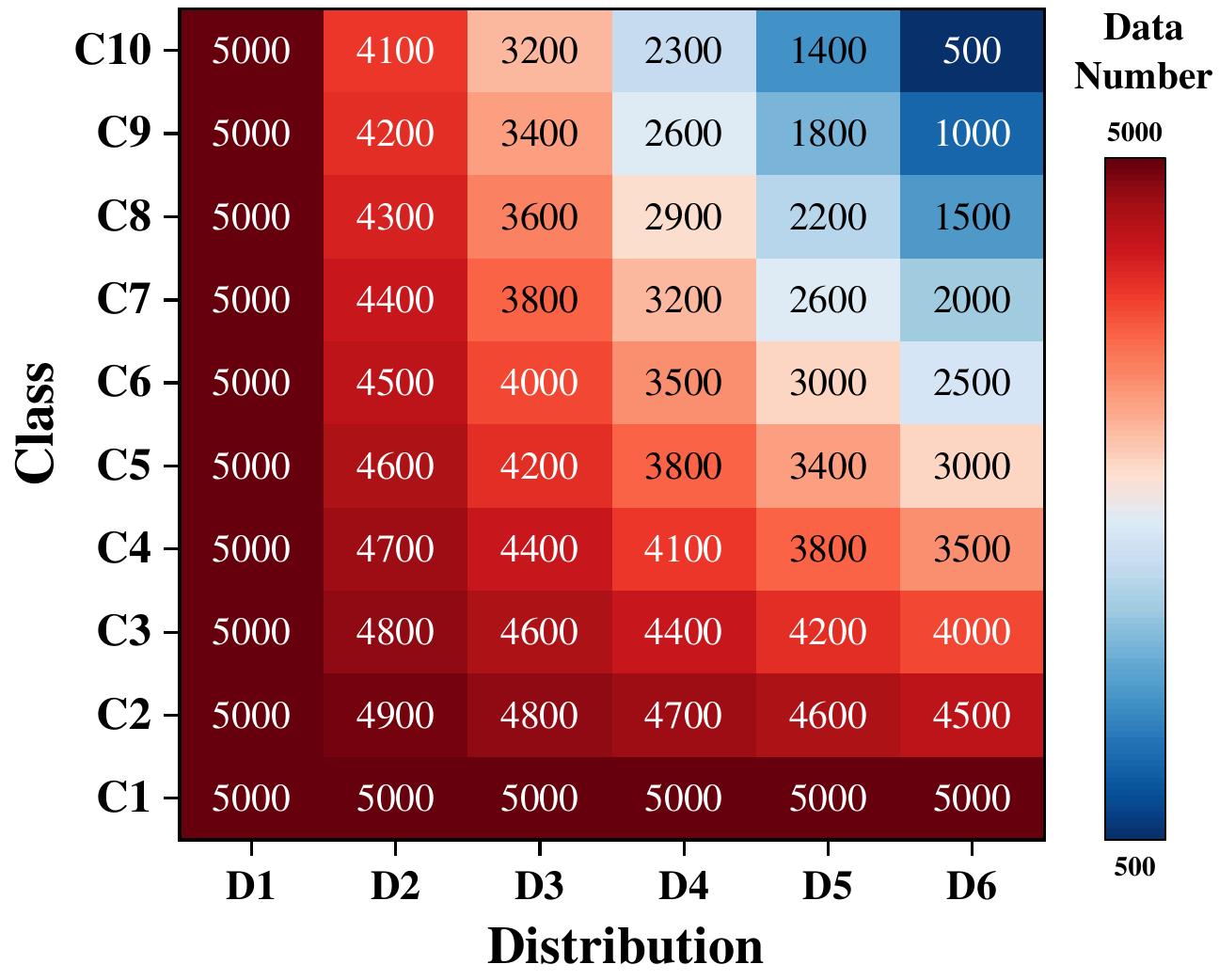}
    \label{fig:Distribution_list}
    }
    \subfloat[Allocating CIFAR-10 D1 into Clients]{
     \includegraphics[width=0.4\textwidth]{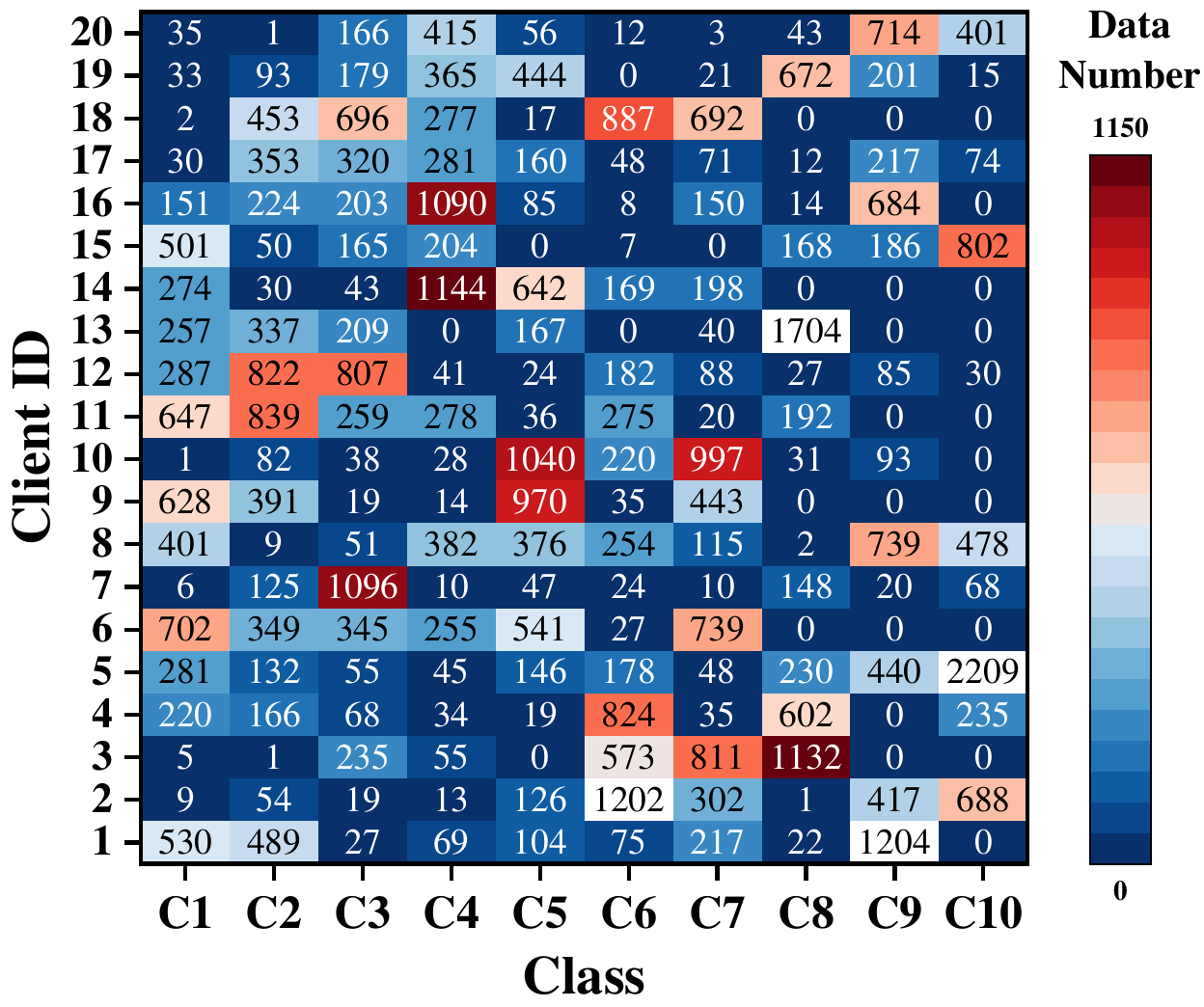}
    \label{fig:dis_example}
    }
    \caption{The generated data distributions of CIFAR-10 and an example of its allocation}
    \label{fig:data_distribution}
\end{figure*}

\begin{theorem}
    The proposed mechanism can protect clients' privacy in the semi-honest client environment.
\end{theorem}
\begin{proof}
    Each client $e$ generates a key pair $(SK_e, PK_e)$ and every client has its personalized key pair. 
    Clients only transmit their public key $PK_e$ to other clients and the private key $SK_e$ don't transmit in any form. Thus, $SK_e$ will not be eavesdropped by others.

    The adversary client $a$ gets the public key of client $e$ and generates random seed $s_{a,e}$ which is the same as client $e$ generated. 
    Then client $a$ generates $\mathbb{RV}_{a,e}\leftarrow \epsilon_{a,e}\cdot PRG(s_{a,e})$. According to the protocol client $a$ can infer $\mathbb{RV}_{e,a}=-\mathbb{RV}_{a,e}$ because $\epsilon_{a,e}=-\epsilon_{e,a}$. However, $\mathbb{RV}_{e,a}$ and $PK_e$ are all the information that client $a$ can get about the client $e$. 

    $\mathbb{N}_e$ can be inferred from Equation~(\ref{eq:proof_semi_client}). It means only getting all the $\mathbb{RV}_{e,v}$ and $\mathbb{Y}_e$, the privacy of $e$ can be breached. The $\mathbb{RV}_{e,v}$ are generated locally on the clients and are not transmitted, so stealing them is very difficult.
    
    \begin{equation}
    \label{eq:proof_semi_client}
    \small
        \mathbb{N}_e=\mathbb{Y}_e-\sum_{v\in \mathbb{E}\setminus{\{e\}}}\mathbb{RV}_{e,v}
    \end{equation}
    
    As analysed above, client $a$ can only infer one $\mathbb{RV}_{e,v}$ about client $e$. Therefore, even if $a$ manages to obtain $\mathbb{Y}_e$ through certain means, it is still unable to compromise the privacy of $e$ due to lack of other $\mathbb{RV}_{e,v}$. 

    Therefore, our mechanism can protect the clients' privacy in the semi-honest client environment.
\end{proof}

\begin{theorem}
\label{server theorem}
    The proposed mechanism can protect clients’ privacy in the semi-honest server environment.
\end{theorem}
\begin{proof}
    The server has access to the public key $PK_e$ but not to any of the private key $SK_e$ of each client $e$. The only value sent by each client $e$ is the pseudo distribution $\mathbb{Y}_e$. 
    The server is unable to access the random seed $s_{e,v}$ since it is not shared by clients. In addition, the server can not generate it due to the lack of the necessary private keys, $SK_e$ or $SK_v$.
    Thus, $\mathbb{RV}_{e,v}\leftarrow \epsilon_{e,v}\cdot PRG(s_{e,v})$ as the key part to infer the privacy of client $e$ based on Equation (\ref{eq:proof_semi_client}) is difficult to obtain for the server.
    
    As a result, only two messages the server can get. One is the pseudo distribution $\mathbb{Y}_e$ of clients and the other is the global distribution $\mathbb{N}_s$ combined with all pseudo distribution $\mathbb{Y}_e$.
\end{proof}

\section{The example of generating training dataset}\label{sec_dataset}
Figure~\ref{fig:Distribution_list} demonstrates the process of transforming the originally balanced global data distribution (a total of 50K data, with 5K in each class) of the CIFAR-10 dataset into an unbalanced distribution. We incrementally remove data from each class to produce unbalanced global distributions, labeled as D2 to D6. 
In the second step, based on the different global data distributions, we allocate data for each class to each client following a Dirichlet distribution ($\alpha=0.5$) for simulating Non-i.i.d. scenarios~\cite{hsu2019measuring,li2022federated,lin2020ensemble}. Figure~\ref{fig:dis_example} shows the results of distributing the D1 global distribution to 20 clients.

\section{Description of baselines}
\label{baseline}
A brief description of the baselines used in our experiments is provided below:
\begin{itemize}[leftmargin=10pt]
    \item \emph{Random Selection (RS)}: a simple method that randomly selects a fixed number of clients.
    \item \emph{Quantity Based Selection (QBS)}~\cite{zeng2020fmore}: QBS selects a fixed number of clients only based on their data volumes in descending order.
    \item \emph{DICE}~\cite{saha2022data}: DICE selects clients with the highest data quality scores, defined in terms of the data volume ratio and the standard deviation of data volume in different categories.
    \item \emph{Diversity-driven Selection (DDS)}~\cite{li2021sample}: DDS selects a subset of clients based on two criteria, (i) Statistical homogeneity, which assesses the similarity between a client's distribution and a uniform distribution; 
    ii) Content diversity, which measures the distance between clients on embedding vectors of their dataset. The selection process favours clients with both high statistical homogeneity and content diversity.
    \item \emph{S-FedAvg}~\cite{nagalapatti2021game}: an in-training client selection method where the Sharply value of local model accuracy is introduced after each round of training. In addition, a Shapley value-based Federated Averaging (S-FedAvg) algorithm is presented to select clients with high contributions to the FL task.
\end{itemize}

\section{communication analysis}
\label{communication_analyze}
\myparagraph{Clients Side Analysis}
On the client side, the communication cost consists of two parts: In the first part, each client uploads its public key to the server, which distributes it to other clients, then receives $(E-1)$ public keys of other clients from the server. The communication cost of the first part is $(1+(E-1)) L_k=E L_k$, where $L_k$ is the number of bits in the public key exchange. In the second part, each client uploads its pseudo data distribution $\mathbb{Y}_e$ to the server, the size of $\mathbb{Y}_e$ is $C\lceil \log_2 y_e^c \rceil$ ($y_e^c$ is the element of $\mathbb{Y}_e$ and $\lceil \log_2 y_e^c \rceil$ is the minimum number of bits required for $y_e^c$). The communication cost is $C\lceil \log_2 y_e^c \rceil$. Then the total communication cost of each client is $E L_k+C \lceil \log_2 y_e^c \rceil$, 

\myparagraph{Server Side Analysis}
On the server side, the two parts of communication cost are public key exchange and pseudo data distribution reception respectively. For the public key exchange part, the communication cost of public key reception is $E L_k$ and the dispatch communication cost is $(E(E-1))  L_k$. The total communication cost of public key exchange is $E^2 L_k$.
For pseudo data distribution receptiotn, he communication cost is $E C \lceil \log_2 y_e^c \rceil$.

\myparagraph{Communication Cost of FL Training}
In each round of FL training, clients upload the local model parameters to the server and receive the global model parameters from the server. The communication cost of each client is $2ML_m$, $M$ is the total number of model parameters and $L_m$ is the number of bits in each parameter. The communication cost of the server is $2ML_m E$ due to the server transmitting the model parameters for each client. This communication cost will happen in each round of FL training.

The number of parameters in the model is far greater than the class number and client number. Therefore the communication cost of FL training is far greater than PASS.

\section{Evaluation of \projecttitle on the total number of 100 clients} 
\label{sec:large_clients}
\begin{figure}[ht]
\vspace{-3mm}
\centering
    \subfloat[25 clients selected]{
    \includegraphics[width=.15\textwidth,height=0.122\textwidth]{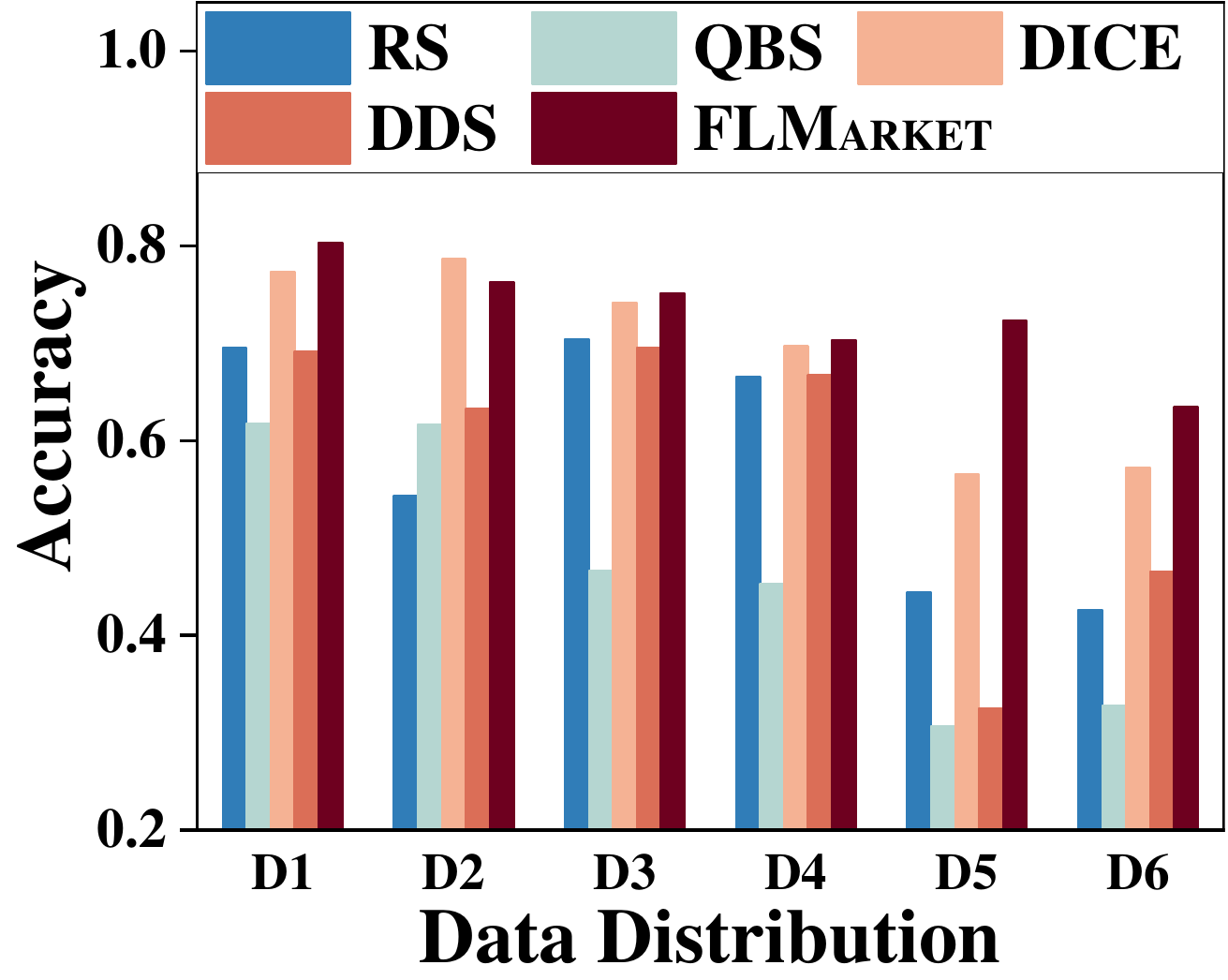}
    \label{fig:cf100x25}}
    \subfloat[50 clients selected]{
    \includegraphics[width=.15\textwidth,height=0.122\textwidth]{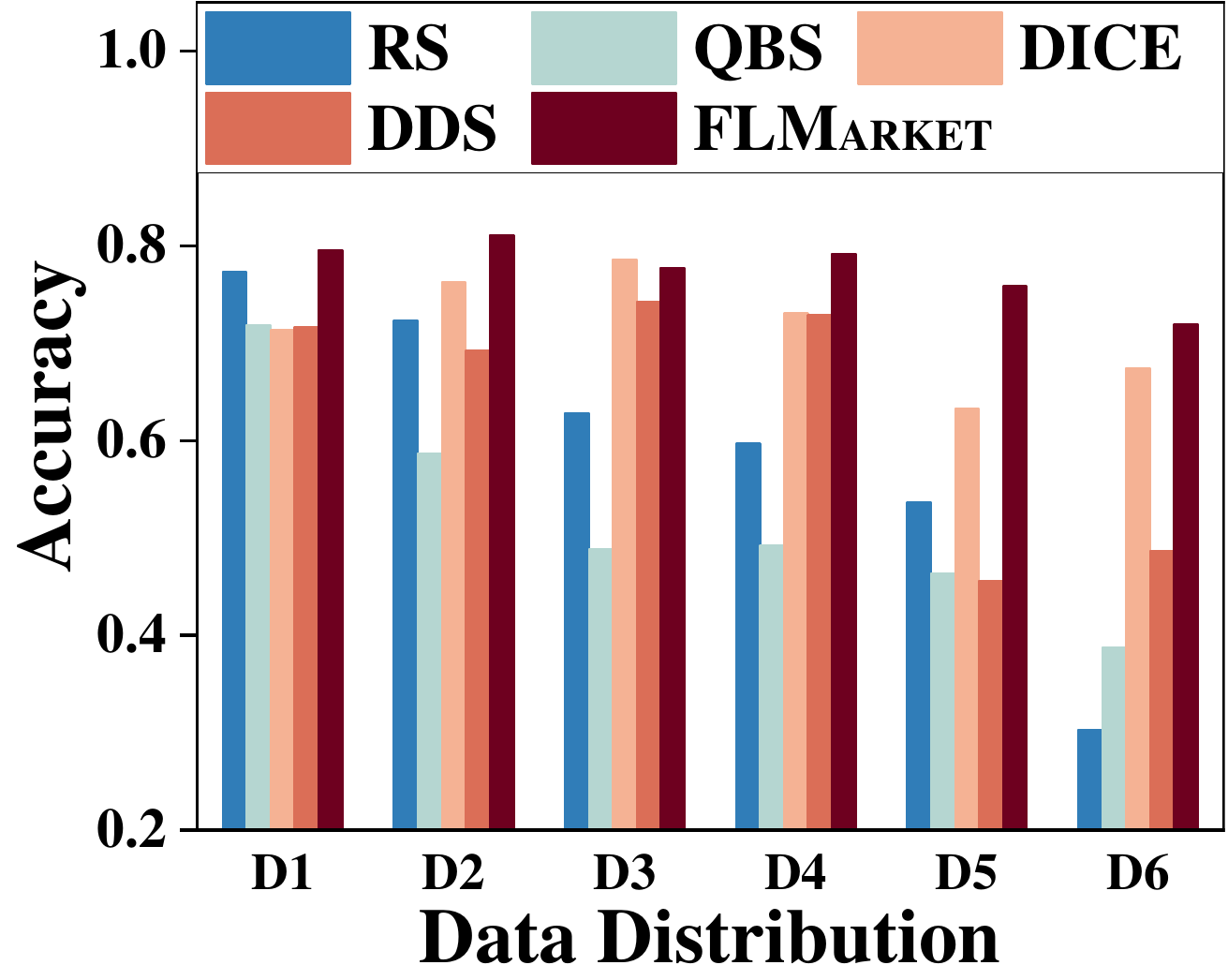}
    \label{fig:cf100x50}}
    \subfloat[75 clients selected]{
    \includegraphics[width=.15\textwidth,height=0.122\textwidth]{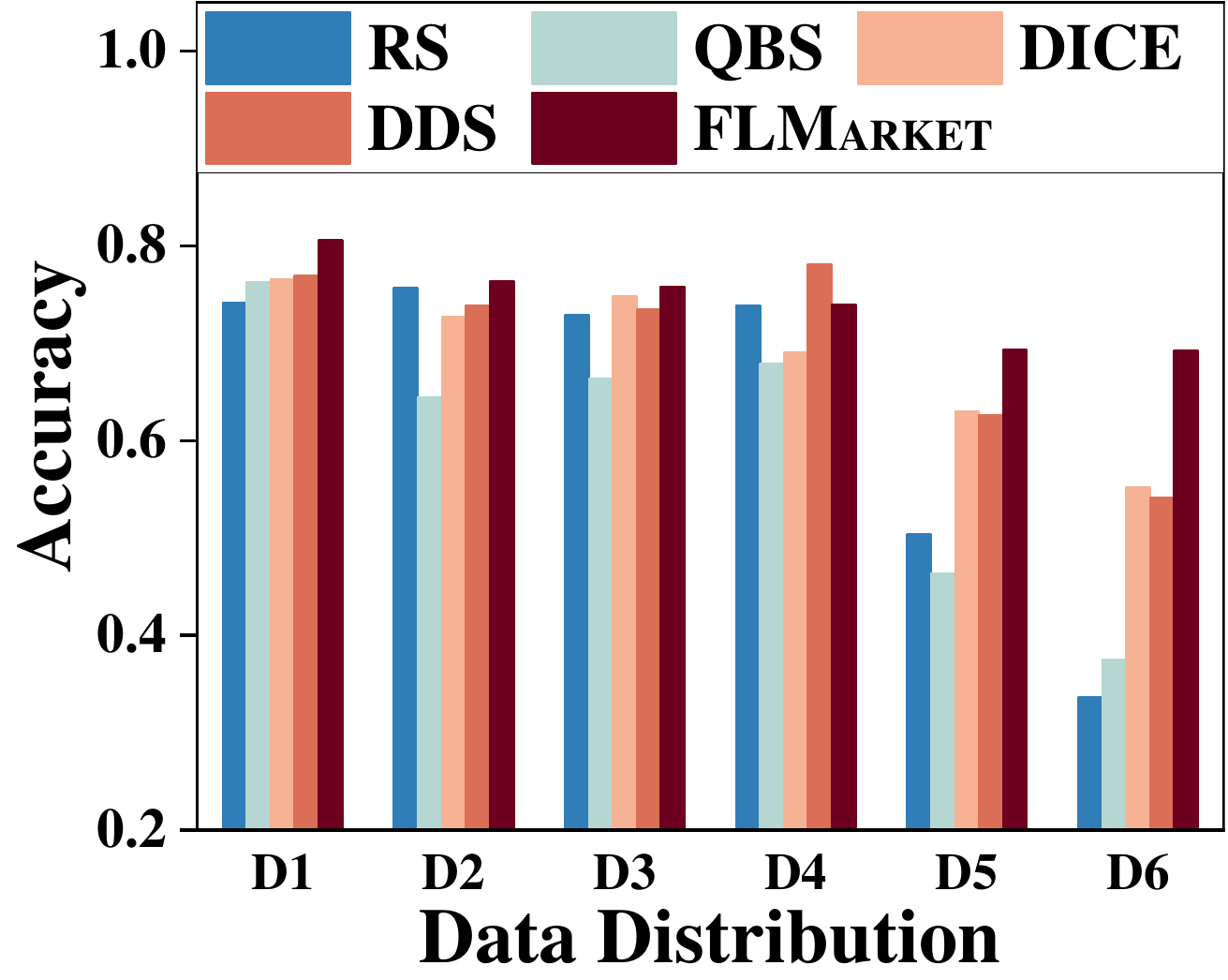}
    \label{fig:cf100x75}}
    \vspace{-3mm}
    \caption{CIFAR-10: $n$ clients selected from 100 clients}
    \label{fig:cifar10x100}
    \vspace{-3mm}
\end{figure}
To evaluate the performance of \projecttitle with a large number of clients, we present the results obtained from experiments in which 25, 50, and 75 clients were selected from a pool of 100 clients. The global data distributions remain consistent with those depicted in Figure~\ref{fig:cifar}, with the only difference being the number of clients.
Figure~\ref{fig:cifar10x100} illustrates that \projecttitle continues to achieve significant average accuracy improvements of 14.66 $\%$, 22.06 $\%$, 5.19 $\%$ and 11.08 $\%$ when compared to other baseline methods, with a large number of clients.
However, comparing Figure~\ref{fig:cifar10x100} and \ref{fig:cifar}, when the client pool is enlarged and more clients are selected, the accuracy improvement of \projecttitle drops by 1-5 $\%$. This is because selecting more clients increases the probability of baseline methods to select high-quality clients.

\end{document}
\endinput